\documentclass[10pt,twocolumn,letterpaper]{article}

\usepackage{cvpr}              %

\usepackage{graphicx}
\usepackage{amsmath}
\usepackage{amssymb}
\usepackage{booktabs}
\usepackage{pifont}%

\newcommand{\cmark}{\ding{51}}%
\newcommand{\xmark}{\ding{55}}%

\usepackage{algorithm}
\usepackage{algpseudocode}

\usepackage{times}
\usepackage{epsfig}
\usepackage{graphicx}
\usepackage{amssymb}

\usepackage{float}
\usepackage{booktabs}
\usepackage{caption}
\usepackage{subcaption}
\usepackage{multirow}
\usepackage[dvipsnames]{xcolor}

\mathchardef\mhyphen="2D
\usepackage[utf8]{inputenc}
\usepackage{comment}
\usepackage{tabularx,colortbl}
\usepackage{xparse,mathtools}
\usepackage{diagbox}
\usepackage{adjustbox,lipsum}

\usepackage{breqn}

\usepackage{tabulary,overpic}

\newlength\savewidth

\usepackage{siunitx}
\usepackage[pagebackref=true,breaklinks=true,colorlinks,bookmarks=false]{hyperref}

\newcommand{\bB}{\mathbf{B}}

\newcommand{\bg}{\mathbf{g}}
\newcommand{\bH}{\mathbf{H}}
\newcommand{\bI}{\mathbf{I}}

\newcommand{\bK}{\mathbf{K}}

\newcommand{\bp}{\mathbf{p}}\newcommand{\bP}{\mathbf{P}}
\newcommand{\bq}{\mathbf{q}}\newcommand{\bQ}{\mathbf{Q}}
\newcommand{\bR}{\mathbf{R}}

\newcommand{\bt}{\mathbf{t}}

\newcommand{\bV}{\mathbf{V}}

\newcommand{\bx}{\mathbf{x}}
\newcommand{\by}{\mathbf{y}}

\newcommand{\btheta}{\boldsymbol{\theta}}

\newcommand{\bpsi}{\boldsymbol{\psi}}

\newcommand{\cD}{\mathcal{D}}

\newcommand{\cF}{\mathcal{F}}

\newcommand{\cL}{\mathcal{L}}

\newcommand{\cX}{\mathcal{X}}
\newcommand{\cY}{\mathcal{Y}}

\DeclareMathOperator*{\argmin}{argmin~}

\makeatletter
\DeclareRobustCommand\onedot{\futurelet\@let@token\@onedot}
\def\@onedot{\ifx\@let@token.\else.\null\fi\xspace}
\def\eg{e.g\onedot} 
\def\ie{i.e\onedot}

\def\etal{et~al\onedot}

\makeatother

\newcommand{\boldparagraph}[1]{\vspace{0.2cm}\noindent{\bf #1:}}

\definecolor{darkgreen}{rgb}{0,0.7,0}

\DeclareMathOperator{\E}{\mathbb{E}}

\usepackage[capitalize]{cleveref}
\crefname{section}{Sec.}{Secs.}
\Crefname{section}{Section}{Sections}
\Crefname{table}{Table}{Tables}
\crefname{table}{Tab.}{Tabs.}

\def\cvprPaperID{2335} %
\def\confName{CVPR}
\def\confYear{2023}

\begin{document}

\title{Coaching a Teachable Student}

\author{Jimuyang Zhang \quad Zanming Huang \quad Eshed Ohn-Bar\\
Boston University\\
{\tt\small \{zhangjim, huangtom, eohnbar\}@bu.edu}
}
\maketitle

\begin{abstract}
We propose a novel knowledge distillation framework for effectively teaching a sensorimotor student agent to drive from the supervision of a privileged teacher agent. Current distillation for sensorimotor agents methods tend to result in suboptimal learned driving behavior by the student, which we hypothesize is due to inherent differences between the input, modeling capacity, and optimization processes of the two agents. We develop a novel distillation scheme that can address these limitations and close the gap between the sensorimotor agent and its privileged teacher. Our key insight is to design a student which learns to align their input features with the teacher's privileged Bird's Eye View (BEV) space. The student then can benefit from direct supervision by the teacher over the internal representation learning. To scaffold the difficult sensorimotor learning task, the student model is optimized via a student-paced coaching mechanism with various auxiliary supervision. We further propose a high-capacity imitation learned privileged agent that surpasses prior privileged agents in CARLA and ensures the student learns safe driving behavior. Our proposed sensorimotor agent results in a robust image-based behavior cloning agent in CARLA, improving over current models by over $20.6\%$ in driving score without requiring LiDAR, historical observations, ensemble of models, on-policy data aggregation or reinforcement learning.

\end{abstract}

\begin{figure}[!t]
    \centering
    \includegraphics[width=3.26in,trim=0.3cm 3cm 2cm 0cm, clip]{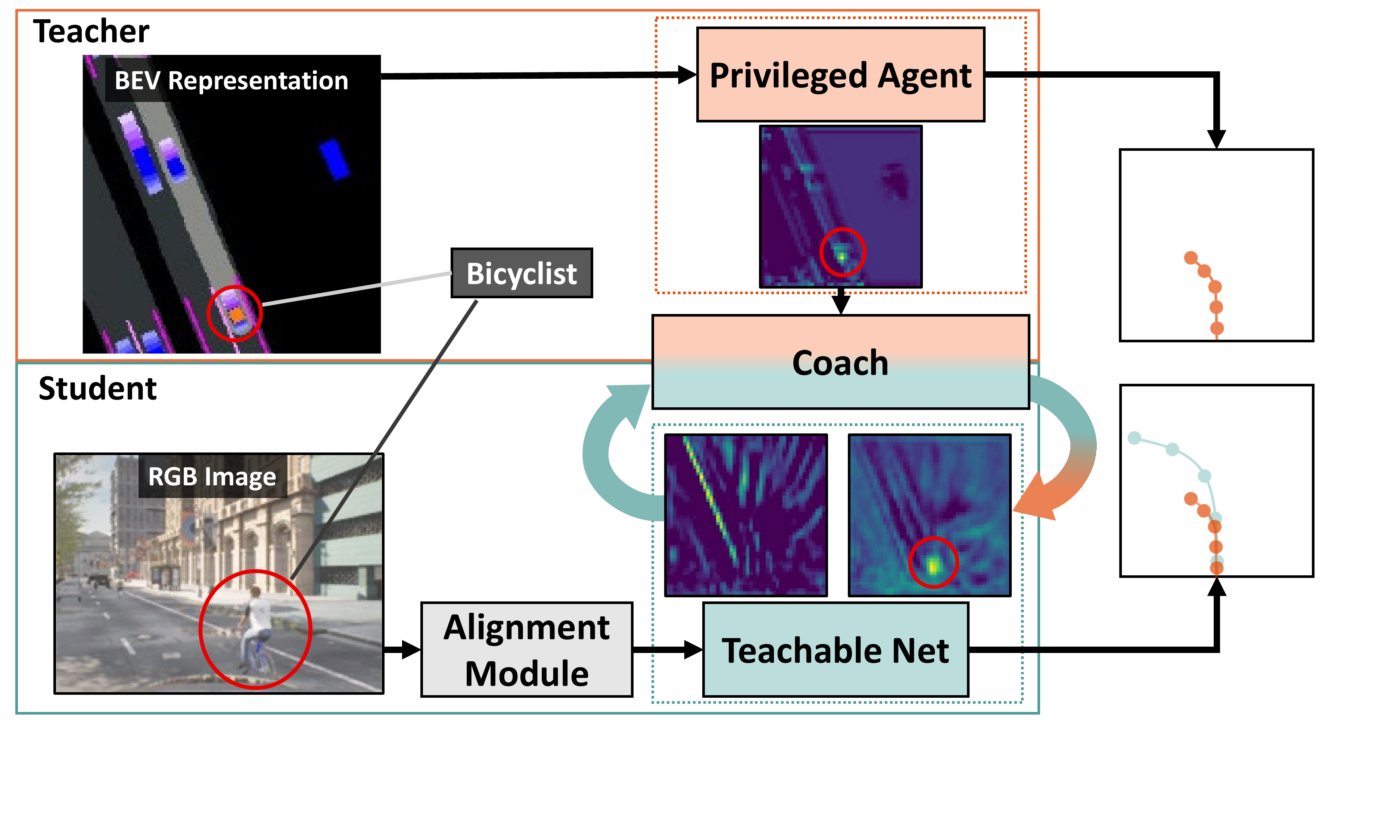}
    \vspace{-0.25in}
    \caption{\textbf{Effective Knowledge Distillation for Sensorimotor Agents.} Our proposed CaT (Coaching a Teachable student) framework enables highly effective knowledge transfer between a privileged teacher and a sensorimotor (\ie, image-based) student. An alignment module learns to transform image-based features to the teacher's BEV feature space, where the student can then leverage extensive and direct supervision on its learned intermediate representations. The student model is optimized via a coaching mechanism with extensive auxiliary supervision in order to further scaffold the difficult sensorimotor learning task. 
    }
    \label{fig:fig1}
    \vspace{-0.2in}
\end{figure}

\vspace{-0.15in}
\section{Introduction}
\label{intro}

Learning internal representations for making intricate driving decisions from images involves a complex optimization task~\cite{pomerleau1989alvinn,levine2016end,bansal2018chauffeurnet}. The inherent challenge for end-to-end training of driving agents lies in the immense complexity of learning to map high-dimensional visual observations into general and safe navigational decisions~\cite{lbc,wen2020fighting,de2019causal}. Even given millions of training examples~\cite{bansal2018chauffeurnet}, today's agents still fail to reliably learn an internal representation that can be used for robust processing of complex visual scenarios (\eg, dense urban settings with intricate layouts and dynamic obstacles) in a safe and task-driven manner~\cite{lbc,wen2020fighting,de2019causal}.

To ease the challenging sensorimotor training task, recent approaches decompose the task into stages, \eg, by first training a high-capacity privileged network with complete knowledge of the world and distilling its knowledge into a less capable vision-based student network~\cite{lbc,lav,zhang2021end,gou2021knowledge,miki2022learning}. However, due to the inherent differences between the inputs and architectures of the two agents, current methods rely on limited supervisory mechanisms from the teacher, \ie, exclusively through the teacher's output~\cite{lbc,lav} or knowledge distillation of a single final fully-connected layer~\cite{zhang2021end,zhao2019sam,wu2022trajectory}. Moreover, the privileged teacher’s demonstration targets may be noisy or difficult for the student to imitate, given the limited perspective~\cite{he2012imitation}. In this work, we sought to develop a more effective knowledge distillation paradigm for training a sensorimotor agent to drive. Our key insight is to enable more extensive supervision from the teacher by reducing the gap between internal modeling and learning capabilities between the two agents. 

Our proposed approach for holistic knowledge distillation is informed by human instruction, which often involves structured supervision in addition to high-level demonstrations, \eg, providing various hints to scaffold information in a way that the student can better understand~\cite{hiemstra1990individualizing}. When teaching others new and challenging skills, \ie, where a student may not be able to replicate the demonstration such as riding a bicycle or driving a vehicle, teachers may provide additional supervision regarding the underlying task structure and their own internal reasoning~\cite{van2002scaffolding}. Analogously to our settings, the privileged teaching agent can potentially provide richer supervision  when teaching a limited capacity sensorimotor student, \ie, through more careful and direct guidance of the underlying representation learning. 

 In our work, we introduce CaT, a novel method for teaching a sensorimotor student to drive using supervision from a privileged teacher. Our key insights are threefold: \textbf{1) Effective Teacher:} We propose to incorporate explicit safety-aware cues into the BEV space that facilitate a surprisingly effective teacher agent design. While prior privileged agents struggle to learn to drive in complex urban driving scenes, we demonstrate our learned agent to match expert-level decision-making. \textbf{2) Teachable Student via Alignment:} An IPM-based transformer alignment module can facilitate direct distillation of most of the teacher’s features and better guide the student learning process. \textbf{3) Student-paced Coaching:} A coaching mechanism for managing difficult samples can scaffold knowledge and lead to improved model optimization by better considering the ability of the student. By holistically tackling the complex knowledge distillation task with extensive teacher and auxiliary supervision, we are able to train a state-of-the-art image-based agent in CARLA~\cite{Dosovitskiy2017CORL}. Through ablation studies, input design, and interpretable feature analysis, we also provide critical insights into current limitations in learning robust and generalized representations for driving agents.

\section{Related Work}
\vspace{-0.15cm}
\boldparagraph{Knowledge Distillation} We study knowledge distillation methods originally developed for model compression and acceleration~\cite{hinton2015distilling,romero2014fitnet,yim2017gift,chung2020feature,yang2020knowledge,huang2017like,passalis2018learning,ji2021show,heo2019comprehensive,chen2021cross} in the context of training sensorimotor. While approaches for feature distillation have been minimally explored in this context, we are motivated by their success in other domains, including image classification \cite{kim2018paraphrasing,srinivas2018knowledge,koratana2019lit,aflalo2020knapsack,kushawaha2021distilling}, object detection \cite{wang2019distilling}, semantic segmentation \cite{zhang2021robust,liu2021source}, and natural language processing \cite{xu2020bert,aguilar2020knowledge,wang2020minilmv2}. Yet, applying such techniques is not trivial given the drastically differing inputs between the privileged agent and student and the overwhelming sensorimotor task. Consequently, driving policy distillation methods only provide supervision either from the teacher’s output~\cite{lav, lbc} or the features of a single fully-connected
layer~\cite{zhang2021end,zhao2019sam,wu2022trajectory}, which (we hypothesize) does not provide sufficient hints to guide the student training. In this work, we provide novel mechanisms to close this gap and enable extensive supervision through deep feature distillation.

\boldparagraph{Imitation Learning to Drive} Recent approaches in imitation learning (IL) to drive can be traced to Pomerleau~\cite{pomerleau1989alvinn}. 
Recently, more elaborate IL-based approaches for driving have emerged~\cite{zhang2021learning,Zhoueaaw6661,osa2018algorithmic,li2018oil,zhao2019lates,zhang2022selfd,Chen2015ICCVa,gupta2017cognitive,chen2021learning}. Specifically, decomposing the imitation learning task into two stages, \ie, by first learning a privileged agent through behavior cloning~\cite{lbc,lav} or reinforcement learning~\cite{zhang2021end,wu2022trajectory,toromanoff2020end,chekroun2021gri} and then training the sensorimotor agent to mimic the output of the privileged agent. Our study is motivated by such approaches, yet we explore the benefits of increased supervision by the privileged agent. Moreover, prior privileged agents produce noisy and sub-optimal demonstration supervision in complex urban scenarios (even when optimized with RL, as demonstrated by our analysis). While most current studies employ the RL-based teacher of~\cite{zhang2021end}, this can be problematic and inefficient in safety-critical conditions in the real-world. In our work, we introduce a novel privileged teacher which can greatly surpass prior agents while only relying on offline behavior cloning.

\boldparagraph{Intermediate Representation for Driving} 
Learning an effective 3D scene representation is crucial for safe autonomous driving. Researchers may obtain such representations by lifting the image to 3D using estimated depth and projecting the frustums into a BEV grid~\cite{philion2020lift,hu2021fiery}. Alternatively, transformer-based architectures can also enable mapping a camera image to the BEV space~ \cite{persformer,zhou2022cross}, \ie, by attending to image-based information when populating a BEV-sampled grid. Related to our work is the study of Chen~\etal~\cite{lav}, which learns a BEV representation from RGB image and LiDAR input. A motion planner is then used to generate future waypoints from the BEV. As this introduces significant challenges, we propose to leverage an IPM-based alignment module which can better structure to image-to-BEV projection task.

\boldparagraph{Curriculum and Self-paced Learning} 
Approaches for curriculum learning methods often structure learning from various difficulty samples~\cite{bengio2009curriculum,li2016weakly,jiang2018mentornet,guo2018curriculumnet,tudor2016hard,shi2016weakly,tang2018attention,kocmi2017curriculum,wang2019dynamically,zhou2020uncertainty}. To better consider the challenging sensorimotor task and limited capacity of the student, we adopt a curriculum (\ie, an imitation coach), with progressively more challenging samples. He~\etal~\cite{he2012imitation} also proposes training a coach across iterations of on-policy data collection~\cite{ross2011reduction}. In contrast, we implement and demonstrate the benefit of a coach over training iterations without a data aggregation part. This can facilitate more efficient model training. The approach is motivated by self-paced learning~\cite{kumar2010self,xiang2020learning}, where the model selects easy samples dynamically at each iteration based on a defined loss. In contrast to such studies, we only smooth the targets of the hard samples (based on the student), instead of discarding them. We empirically find this to result in improved coaching, potentially due to improved scaffolding of the difficult samples.

\section{Method}
\label{sec:method}
\begin{figure*}[!t]
    \centering
    \includegraphics[width=\textwidth,trim=0cm 0.4cm 0.7cm 0cm, clip]{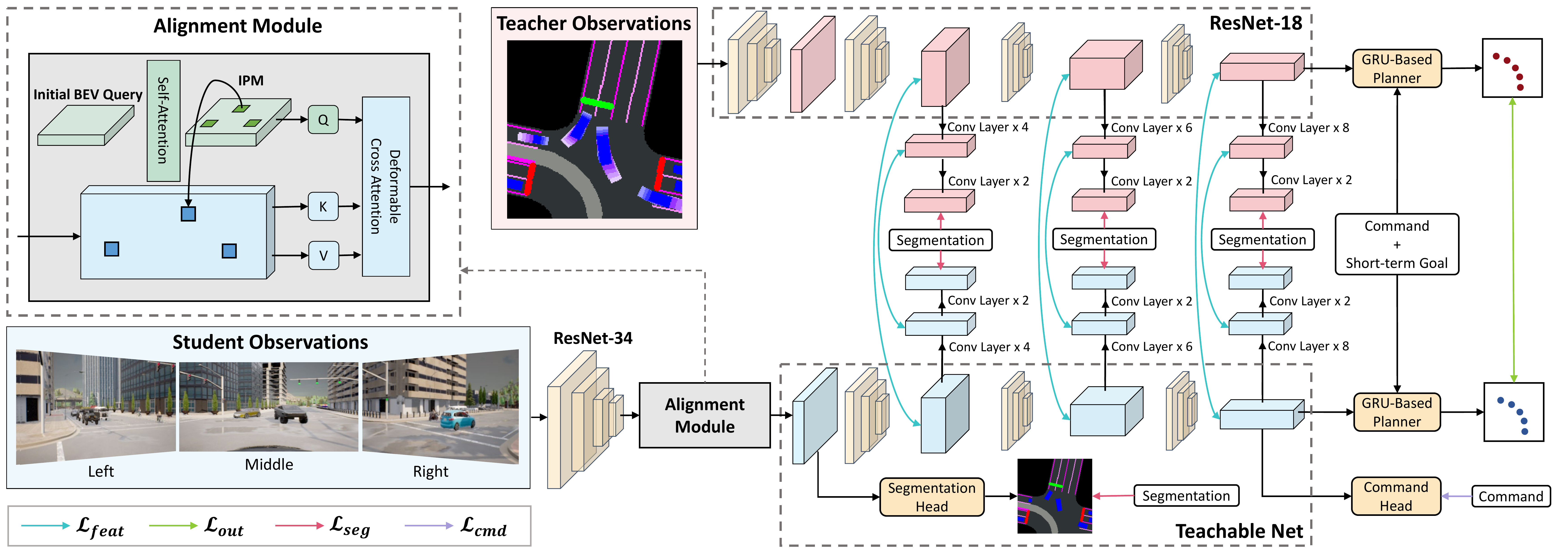}
    \caption{\textbf{Overview of Architecture and Distillation Process of CaT.} We first train a high-capacity privileged teacher agent (colored in red) which takes an augmented BEV representation as input, and produces a set of future waypoints conditioned on command and short-term goal. Our teachable student agent (colored in blue) maps the image features to the BEV space using an alignment module, which is trained by distillation of intermediate features (arrows colored in blue) and final output (arrow colored in green). Additionally, we leverage supervision from auxiliary segmentation (arrow colored in red) and command prediction tasks (arrow colored in purple) interlaced throughout the network to facilitate the learning of task-relevant features.
    }
    \label{fig:overview}
\end{figure*}

Our goal is to decompose the challenging sensorimotor learning task and ease model optimization through effective supervision from a teacher agent. We first formulate the problem of privileged agent distillation in Sec.~\ref{subsec:form}. Next, we address current limitations in privileged agent design to train a robust imitation learned teacher in Sec.~\ref{subsec:teacher}. We then propose a transformer-based student architecture to resolve image-to-BEV feature alignment (Sec.~\ref{subsec:student}). Finally, we train the sensorimotor student via extensive supervision and knowledge distillation (Sec.~\ref{subsec:distillation}) via a progressive (\ie, coaching) mechanism (Sec.~\ref{subsec:coaching}).

\subsection{Formulation}
\label{subsec:form}
Driving can be formulated as a sequential decision-making problem over a set of sensory observations $\cX^s$, a set of actions (motor commands) $\cY$, and a policy function $f^s_{\btheta}\colon\mathcal{X}^s \to \mathcal{Y}$ for mapping observations to actions at each time step~\cite{osa2018algorithmic,ross2011reduction,he2012imitation}. We consider the task of learning a sensorimotor student agent via knowledge distillation~\cite{lbc,gou2021knowledge} from a teacher agent $f_{{\bpsi}}^t\colon\mathcal{X}^t \to \mathcal{Y}$ with privileged access to environmental information, \eg, ground-truth lanes, obstacles, and traffic light states. We parameterize the two agents as neural networks using weights $[\btheta, \bpsi] \in \mathbb{R}^d$, and denote $\cF^{s}(\cdot; \btheta)$ and $\cF^{t}(\cdot; \bpsi)$ to be the feature maps for the sensorimotor student network and teacher network, respectively. Given a dataset $\cD$ comprising sensory and privileged observations and a loss function $\cL$, the student can be optimized from the teacher using
 \begin{equation}
 \begin{aligned}
     \argmin_\btheta
     &\E_{(\bx^s,\bx^t) \sim \cD} [\cL(\cF^{s}(\bx^s; \btheta), \cF^{t}(\bx^t; \bpsi) )] 
\end{aligned}  
\label{eqn:form}
\end{equation}
where the loss may be defined over a final layer, \ie, to match the output of the teacher~\cite{lbc,lav}, or over internal layers as well, \ie, feature distillation~\cite{zhang2021end,wu2022trajectory,zhao2019sam}. Given the challenging end-to-end sensorimotor learning task, the role of the teacher is to provide effective supervision to the student, \ie, informative targets in Eqn.~\ref{eqn:form}. However, as the two agents operate from \textit{drastically differing inputs}, the knowledge transfer from the teacher to the student can become ineffective. Specifically, the task of the student is significantly more challenging than that of the teacher, \ie, due to inherent noise and uncertainty. Indeed, there is currently a substantial gap between the performance of current sensorimotor learning agents and their teaching agents, even assuming access to expensive and high-quality 3D observations such as LiDAR~\cite{chitta2022transfuser,lbc,lav} (not used in our work). The teaching process should also ideally consider any limited capacity of the student to facilitate effective optimization and learning~\cite{he2012imitation}. Moreover, learning an optimal and robust teacher in itself can be a challenging task~\cite{lbc}, even assuming the simplified learning task such as solved perception, as we demonstrate in Sec.~\ref{sec:analysis} and addressed in Sec.~\ref{subsec:teacher}.

\boldparagraph{Problem Setup and Agent Observations}
We develop our CaT framework in the context of the CARLA simulator~\cite{Dosovitskiy2017CORL}. Our objective is to train a goal-conditional sensorimotor agent ~\cite{Codevilla2018ICRA,chitta2022transfuser} for mapping image and goal observations to vehicle throttle and steering control~\cite{muller2018driving,lbc}. We follow standard conditional imitation learning~\cite{lav,chitta2022transfuser} and assume access to three (non-overlapping) RGB camera views $\bI=[\bI_0, \bI_1, \bI_2]\in\mathbb{R}^{W\times H\times 3}$, a categorical navigational command $c\in\{1, \dots, 6\}$ (\ie, turn left, turn right, follow, forward, left lane changing and right lane changing) and an intermediate noisy goal $\bg\in\mathbb{R}^2$ sampled from a GNSS (Global Navigation Satellite System)~\cite{lav}. We note that both the command and GNSS observations can be easily obtained by today's vehicles, \ie, through generation of an A$^{*}$ plan~\cite{hart1968formal} from a static map and coarse positioning system. The GNSS goals are sampled every 50-100 meters and reflect real-world measurement errors~\cite{lav}. We compute a BEV $\bB\in\{0,1\}^{W_B\times H_B \times C_B}$ by rendering privileged (\ie, ground-truth) information from the underlying simulation, including 3D location and state of lanes, pedestrians, vehicles, and traffic lights~\cite{zhang2021end}. While prior work has generally leveraged a standardized BEV representation~\cite{lbc,zhang2021learning,hu2021fiery,saha2022translating,huang2022assister}, we find its design to be crucial to training an optimal teacher policy as will be discussed in Sec~\ref{subsec:teacher}. We let $\bx^s = (\bI, \bg, c) \in \mathcal{X}^s$ and $\bx^t = (\bB, \bg, c) \in \mathcal{X}^t$ be the student and teacher observations, respectively. Given these observations at each time step, the agent learns to predict 10 future 2D waypoints in top-down vehicle coordinates for the next 2.5 seconds of driving. These are then given to a lateral and longitudinal PID controllers~\cite{lbc,muller2018driving,visioli2006practical} to generate the final action. 

Given this formulation, we next discuss our teacher agent design and training process. In particular, we aim to learn a privileged agent which can not only effectively solve the goal-oriented navigation task, but also \textit{facilitate effective distillation and coaching} of a sensorimotor student policy, as will be discussed in Sec.~\ref{subsec:distillation}.

\subsection{Learning an Effective Teacher}
\label{subsec:teacher}

In our formulation, learning a student agent begins with training an effective privileged agent $f_{{\bpsi}}^t$. However, we find this itself to be a complex task. Even with complete knowledge of perfect BEV perception, current privileged agents are suboptimal, significantly under-performing CARLA's built-in autopilot~\cite{lbc,zhang2021end}. In this section, we uncover the underlying reason for this under-performance through the BEV design. Informed by this key finding, we train a highly effective imitation-learned teacher that surpasses prior privileged agents and matches expert-level performance in CARLA without requiring extensive data collection~\cite{ross2011reduction} or reinforcement learning~\cite{zhang2021end}. Our proposed privileged agent can then enable effective supervision of the student in Sec.~\ref{subsec:student}. 

\boldparagraph{Teacher Training via Direct Expert Imitation}
 In imitation learning, an oracle (\ie, expert driver) $f^*\nolinebreak\colon\mathcal{X}^t\nolinebreak\to \nolinebreak \mathcal{Y}$ is defined for demonstrating optimal actions $\by^*$. In CARLA~\cite{Dosovitskiy2017CORL}, the expert is defined using longitudinal and lateral PID controllers which are carefully tuned and augmented by a collection of manually specified rules for handling diverse scenarios (\eg, sudden braking) over the underlying simulation state. The designed expert can then be used to generate trajectories as supervised data for learning the privileged policy function $f_{{\bpsi}}^t$ with behavior cloning~\cite{Codevilla2019ICCV,Codevilla2018ICRA,muller2018driving}. Our privileged agent is a ResNet-18 model~\cite{He2016CVPR}, as shown in Fig.~\ref{fig:overview} (further increasing modeling capacity was not found to be beneficial). The ResNet model is followed by a GRU-based conditional waypoint predictor~\cite{lav,chitta2022transfuser}. Due to the complexity of our dense urban navigation task, \eg, with multi-lane roads, intersections, and merging, we incrementally refine the waypoints using two GRUs within each conditional branch. The first GRU regresses the preliminary set of waypoints via sequential waypoints directly from the embedded features. Subsequently, the second GRU takes the predicted waypoints together with the short-term goal and embedded features, to produce a refined set of waypoint targets.

 Even assuming perfect perception, we find it difficult to train privileged agents that exhibit robust planning behavior in complex urban driving scenarios. Specifically, when using the most challenging CARLA benchmark (Longest6~\cite{chitta2022transfuser}, also detailed in Sec~\ref{sec:analysis}) using the standard BEV representation of Chen~\etal~\cite{lbc} leads to 26\% driving score compared to 72\% by the rule-based expert. The reinforcement learning agent of Zhang~\etal~\cite{zhang2021end} achieves $60\%$ on this task. We sought to explore the limits of behavior cloning for this task, as a noisy and sub-optimal teacher can hinder the training of the student model. In particular, we hypothesize that the reason for the poor performance lies in the increased complexity of the BEV state representation in dense scenarios. In such scenarios, learning to extract task-relevant details can become more challenging. To address this issue, we propose to introduce additional channels (akin to affordances~\cite{gibson2014ecological,Chen2015ICCVa,Sauer2018CORL}) that can more easily translate into safety-critical decision-making, simplify the learning task, and provide an expert-level teaching agent. 
 
\begin{figure*}[t]
    \centering
    \setlength{\tabcolsep}{2pt}
    \begin{tabular}{cccccc}
  \multirow{2}{*}[1cm]{\includegraphics[width=3.2cm]{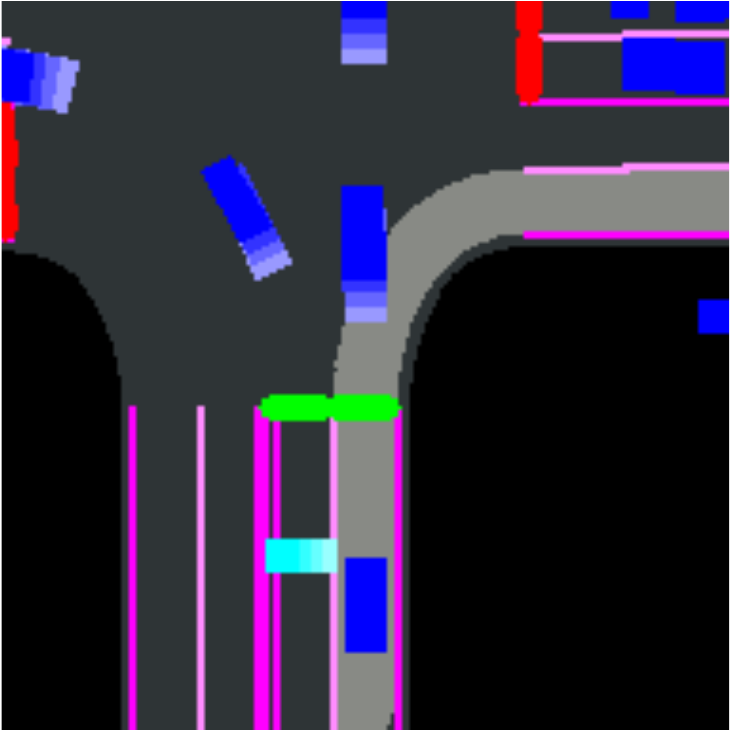}} &
 \multirow{2}{*}[1cm]{\includegraphics[width=3.2cm]{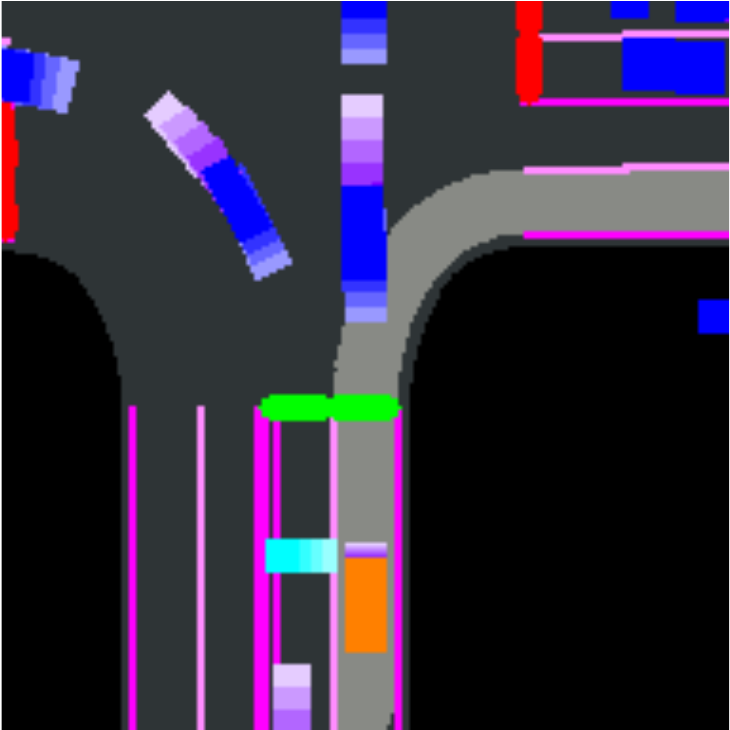}} &
  \includegraphics[width=1.8cm]{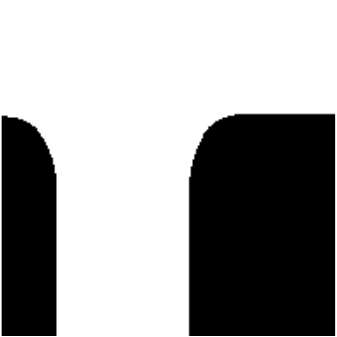} & 
  \includegraphics[width=1.8cm]{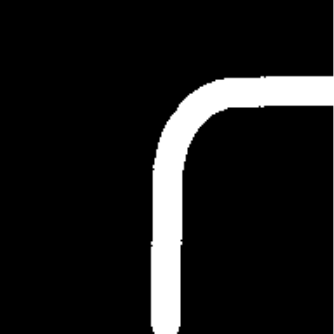} & 
  \includegraphics[width=1.8cm]{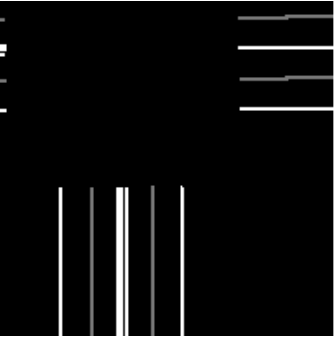} & 
  \includegraphics[width=1.8cm]{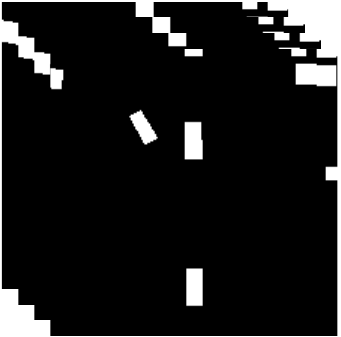} 
    \\
     && (c) Road  & (d) Desired Route & (e) Lane Marks & (f) Vehicles 
    \\
    &&
  \includegraphics[width=1.8cm]{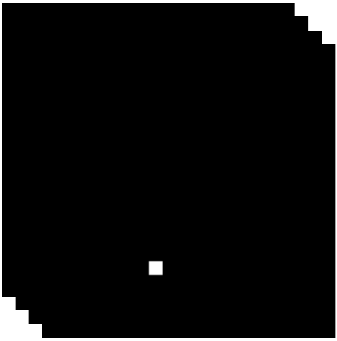} & 
  \includegraphics[width=1.8cm]{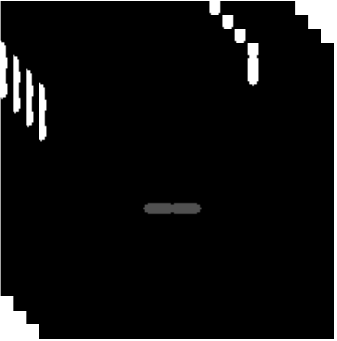} & 
  \includegraphics[width=1.8cm]{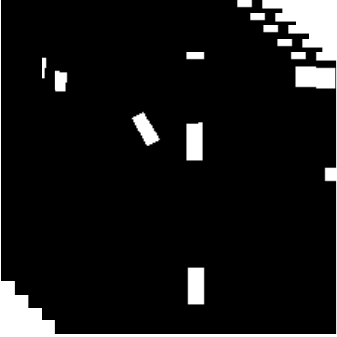} & 
  \includegraphics[width=1.8cm]{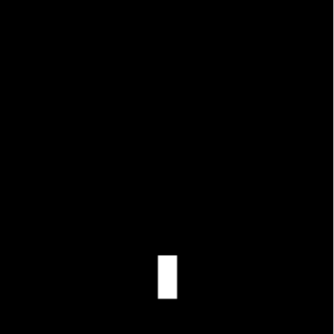} 
    \\
     \vspace{-0.15cm}
   (a) Baseline \cite{zhang2021end} & (b) With Safety Hints & (g) Pedestrians  & (h) Traffic Lights & (i) \colorbox{Lavender}{Agent Forecast} & (j) 
   \colorbox{BurntOrange}{Entity Attention}
    \end{tabular}
    \vspace{-0.05in}
    \caption{\textbf{Visualization of Our Proposed BEV with Safety Hints.} (a) and (b) visualize the baseline and our proposed BEV by encoding different channels into colors. Roads are shown in dark grey, Desired Route is shown in light grey, Lane Marks are shown in magenta,  Vehicles are shown in blue with past trajectories shown in decreasing saturation with time into the past, Pedestrians are shown in light blue, Traffic Lights are shown in red or green according to its state, Agent Forecasts are shown in purple with decreasing saturation with time into the future, Entity Attention is shown in orange. (c) - (j) shows the different channels separately.}
    \label{fig:bev}
    \vspace{-0.15in}
\end{figure*}

 \boldparagraph{BEV with Safety Hints} Our BEV encodes state information into distinct channels, including drivable road regions, the desired route based on the command and layout, lane markings, and dynamic obstacles (see Fig.~\ref{fig:bev}). Given the low performance of prior privileged agents, in particular around dynamic objects, the agent could benefit from more explicit safety-driven cues in the BEV.  We propose to add two types of channels of (1) predicted agents' future and (2) entity attention. First, we utilize a kinematics bicycle model~\cite{chitta2022transfuser} in order to efficiently predict future trajectories of dynamic objects. This enables us to iteratively predict and represent short-term future position, orientation, and speed of agents (our supplementary contains additional information regarding this process). Secondly, we encode an explicit attention channel for highlighting \textit{potential future infractions}, \ie, based on the kinematics bicycle model applied to the ego-vehicle. While this assumes that the current vehicle throttle and steering remain unchanged, the entity attention channel helps the agent better reason over safety while correlating states with actions. Surprisingly, the additions of the safety hints channels provide a strong teacher with offline behavior cloning alone, even improving over the rule-based expert (Table~\ref{tab:leaderboard}, 73\% vs. 72\%). We note that many of the remaining failures are due to simulation time-out in crowded intersections. Our boosted teacher agent will be critical for effective distillation in Sec.~\ref{subsec:distillation}.

 \subsection{Learning a Teachable Student}
 \label{subsec:student}
 Due to differing inputs and modeling capacities, it can be difficult to align the image-based student features and output with the BEV-based privileged teacher. We propose to alleviate misalignment by designing a teachable student network, \ie, a model that matches as much of the teacher architecture as possible. Specifically, we propose to improve teacher-student model alignment by leveraging an internal image-to-BEV alignment module. As shown in Fig.~\ref{fig:overview}, the three residual blocks following the alignment module can consequently facilitate knowledge transfer via \textit{direct distillation of most of the teacher's features} in Sec.~\ref{subsec:distillation}.

 \boldparagraph{IPM-based Transformer Alignment Module} Explicitly mapping images to BEV space for downstream tasks can be challenging. While this can be achieved via depth estimation and lifting~\cite{philion2020lift,hu2021fiery}, we sought to incorporate an efficient and differentiable module which can effectively attend and reason over various features while mapping from front view to BEV. A transformer-based architecture\cite{vaswani2017attention,persformer} provides a natural choice as it can be optimized end-to-end with respect to the driving task while also carefully aligning features among \textit{spaces with arbitrary dimensionalities}. As shown in Fig.~\ref{fig:overview}, we sample queries $\bQ_{init}\in~\mathbb{R}^{L_H\times L_W\times L_C}$ using a spatial parameterization of the BEV space and process them using a self-attention module~\cite{vaswani2017attention,persformer}. The output of self-attention module $Q$ can be formulated as 
 \begin{equation}
\bQ=\text{softmax} \left( \frac{\bQ_{init}\bK^T_{init}}{\sqrt{d}} \right)\bV_{init}
\end{equation}
 where $\bK_{init}$, $\bV_{init}\in~\mathbb{R}^{L_H\times L_W\times L_C}$ provide the key-value pairs and $d$ is the dimension of the query. Subsequently, each query point $\bq\in~\bQ$ is mapped to its corresponding reference point  $\bp\in~\mathbb{R}^3$ (homogeneous coordinates) in the image features $\cF_{{RGB}}\in~\mathbb{R}^{Z_{H}\times Z_{W}\times Z_{C}}$ using an Inverse Perspective Mapping (IPM)
\vspace{-0.1cm}
 \begin{equation}
 \bp = s \bP_k \bR_k (\bq - \bt_k) \label{eq:perspective}
 \vspace{-0.1cm}
\end{equation}
 where $s$ is a scale factor, $\bP_k\in~\mathbb{R}^{3\times3}$, $\bR_k\in~\mathbb{R}^{3\times3}$, and $\bt_k\in~\mathbb{R}^3$ are the $k^{th}$ camera intrinsics, rotation and translation relative to the center of the ego-vehicle, respectively~\cite{bertozzi1998extension}. Note that we employ a different IPM mapping for each camera view to perform the projection. 

To populate the student's BEV features $\cF_{BEV}\in~\mathbb{R}^{H_{B}\times W_{B}\times C_{B}}$ (\ie, the output of the alignment module in Fig.~\ref{fig:overview}), we leverage a deformable cross-attention mechanism based on deformable DETR~\cite{zhu2020deformable}. This enables the network to attend to multiple regions around the reference point in the image features when transforming to BEV feature space. The reference point $\bp$ and the learned deformable offsets can then be used as keys $\bK$, with their corresponding features from the extracted images features the values $\bV$, for a cross-attention module
\begin{equation}
\cF_{{BEV}} = \text{DeformAttn} \left( \bQ, \cF_{RGB}, \bH \right)
\end{equation}
where $\bH = \{\bP_k, \bR_k, t_k\}_{k=1}^3$ contains the combined multi-view IPM parameters from Eqn.~\ref{eq:perspective}. In addition to the proposed alignment and teachable modules, we also incorporate prediction modules for auxiliary tasks (\ie, BEV prediction~\cite{chitta2022transfuser}), discussed in Sec.~\ref{subsec:distillation}. 
 
  \vspace{-0.05cm}
\boldparagraph{GRU-based Waypoint Predictor} The transformed image-to-BEV features are inputted to three residual blocks and a GRU-based conditional branch based on the command~\cite{lav}.

\subsection{Student Loss} 
\label{subsec:distillation}

Even with our carefully designed architecture, the student's image-based learning task remains significantly more challenging than the teacher's. Thus, both optimization and alignment with the teacher can benefit from extensive supervision and gradual student instruction, \ie, to scaffold supervision and better match the student's ability. Our distillation process provides ample beneficial supervision for the student through three holistic mechanisms. First, we incorporate deep distillation losses, \ie, both over output (\ie, waypoint regression task) and feature maps, as shown in Fig.~\ref{fig:overview}, that can more directly supervise internal layers. Second, we incorporate additional auxiliary losses that regularize the optimization process. This includes ground-truth supervised losses, \eg, defined over segmentation and control command, and a geometric loss defined over the post-alignment features that supervises for correct BEV structure. Finally, we leverage a student-paced coaching mechanism for gradually increasing the difficulty of the waypoint prediction task throughout the iterations in Sec.~\ref{subsec:coaching}. 

 \boldparagraph{Loss Functions} Our optimization objective for guiding the distillation process is a weighted sum over both distillation and auxiliary tasks: 
\begin{equation}
    \mathcal{L}_{{CaT}}= \mathcal{L}_{{out}} + \mathcal{L}_{feat} + \mathcal{L}_{seg} + \mathcal{L}_{{cmd}}
    \label{eqn:loss}
\end{equation}
The various loss terms enable guiding holistic aspects of student learning.

\boldparagraph{Output Distillation} We leverage an ${L}_1$ loss computed over \textit{all of the conditional branches} for each instance
\begin{equation}
\mathcal{L}_{out}= \sum_{c=1}^C  \| f^s_{{\btheta}}(\bx^{s}, c)^{} - f^{t}_{{\bpsi}}(\bx^{t}, c) \|_1 
\end{equation}
This can be done by sampling from the teacher with different commands $c$ (following Chen~\etal~\cite{lbc}).

\boldparagraph{Feature Distillation Loss} Our proposed student architecture provides BEV-space feature alignment, \ie, for direct internal feature matching between the student and the teacher (the first term in our loss in Eqn.~\ref{eq:featloss}). This is in contrast to current methods which distill the output or a final fully-connected layer~\cite{lbc, lav,zhang2021end,wu2022trajectory}. However, while this can provide rich supervision for the student, the two models should not match entirely, \ie, due to differences in information processing for a perception-based planning task. Moreover, the BEV-based features can be highly sparse and structured, with slight feature offsets being less meaningful than obtaining a task and scene-relevant representation. To flexibly account for the student-based task and effectively represent structural information in BEV space, our feature loss is computed as
\vspace{-0.1in}
\begin{equation}
    \begin{aligned}
    \cL_{feat} = &\sum_{i=1}^3 \Bigl[ \| \cF_{i}^{s}(\bx^s) - \cF_{i}^{t}(\bx^t) \|_2  \,+ \\
    &\|T^s_i(\cF_{i}^{s}(\bx^s)) - T^t_i(\cF_{i}^{t}(\bx^t)) \|_2 
    \,+ \\
    & \lambda_{CD}\| \cF_{i}^{s}(\bx^s) - {\cF_{i}^{t}(\bx^t))} \|_{CD} \Bigr] 
    \end{aligned}
    \label{eq:featloss}
\end{equation}
where we use $\cF_{i}$ to denote layers within the networks (as shown in Fig.~\ref{fig:overview}, we directly distill three layers), $T_{i}$ indicates the features following several convolutional layers (trained jointly with the student network model), and $CD$ stands for a Chamfer Distance~\cite{paschalidou2021neural} (we set $\lambda_{CD}=0.1$). The final term is computed over thresholded activations and a spatial soft-argmax~\cite{lbc} from each feature map.

\boldparagraph{Task-oriented Auxiliary Tasks} To ensure essential task-relevant information is preserved throughout the distillation process, we leverage regularizing supervision in the form of segmentation and command-based auxiliary tasks interlaced throughout the network~\cite{chitta2022transfuser,ohn2020learning}. Our segmentation loss $\cL_{seg}$ is a cross-entropy loss, computed over the ground-truth BEV and averaged across the segmentation prediction heads. Our command prediction loss $\cL_{cmd}$ is binary cross-entropy supervised by the ground-truth command $c$, which facilitates learning task-relevant features~\cite{lav,Codevilla2019ICCV}.

\subsection{Student-paced Coaching} 
\label{subsec:coaching}
While Sec.~\ref{subsec:distillation} provides ample supervision on the student's internal representations, the teacher's targets may still be difficult to imitate for the student. To better consider the learning ability of the student, we propose to leverage a student-paced training mechanism which can gradually coach the student, \ie, through increasingly difficult samples. When computing the loss in Eqn.~\ref{eqn:loss}, we define a coach,  
\vspace{-0.1in}
\begin{equation}
\mathcal{F}^{t} \leftarrow \lambda_i \mathcal{F}^{s} + (1-\lambda_i) \mathcal{F}^{t}, \text{ if } \cL_{CaT} > \tau_i 
\label{eqn:coach}
\end{equation}
which interpolates the teaching targets with the student's predictions and features. $\lambda_i$ is a hyperparameter which is linearly decreased to 0 over training iteration $i$. The modified targets are only computed over the hard samples at each batch. Within each batch, $\tau_i$ is a threshold defining the lowest 50\% of the scored samples for the loss with respect to the privileged teacher. While smoothing the targets in this manner may seem counterintuitive, \ie, compared to techniques which mine hard examples, the coaching mechanism aims to stabilize training by reducing the difficulty when the student is unable to perform the optimal action. Eqn.~\ref{eqn:coach} enables adjusting the learning rate in a sample-selective manner (initially suppressing difficult samples).

\begin{table*}[!t]
  \caption{\textbf{Quantitative Evaluation on the Longest6 Benchmark.} Comparison of CaT with prior methods in terms of Driving Score (DS), Route Completion (RC), and Infraction Score (IS). 
  Additional infraction metrics (Pedestrian Collisions (Ped), Vehicle Collisions (Veh), Layout Collisions (LC), Red Light Violations (Red), Off-road Infraction (OR), Route deviation (Dev), Route Timeouts (TO), Agent Blocked (Blk)) are shown. FD refers to Feature Distillation, SH refers to Safety Hints. *-re-trained by us using the publicly available code. Mean and standard deviation are computed over three runs. CaT outperforms the state-of-the-art by a large margin in terms of DS. 
  }
  \label{tab:leaderboard}
  \vspace{-0.2cm}
  \centering
  \normalsize
  \resizebox{\textwidth}{!}{\begin{tabular}{p{5.5cm} p{1pt} |cc|ccc|cccccccc}%
    \toprule
    \textbf{Method} & \multicolumn{1}{p{0.1cm}|}{}
    & \multicolumn{1}{p{1cm}}{\centering \textbf{RGB}}
    & \multicolumn{1}{p{1cm}|}{\centering \textbf{LiDAR}}
    & \multicolumn{1}{p{1.8cm}}{\centering \textbf{DS} $\uparrow$}
    & \multicolumn{1}{p{1.8cm}}{\centering \textbf{RC} $\uparrow$}
    & \multicolumn{1}{p{1.8cm}|}{\centering \textbf{IS} $\uparrow$}
    & \multicolumn{1}{p{1cm}}{\centering \textbf{Ped} $\downarrow$}
    & \multicolumn{1}{p{1cm}}{\centering \textbf{Veh} $\downarrow$}
    & \multicolumn{1}{p{1cm}}{\centering \textbf{LC} $\downarrow$}
    & \multicolumn{1}{p{1cm}}{\centering \textbf{Red} $\downarrow$}
    & \multicolumn{1}{p{1cm}}{\centering \textbf{OR} $\downarrow$}
    & \multicolumn{1}{p{1cm}}{\centering \textbf{Dev} $\downarrow$}
    & \multicolumn{1}{p{1cm}}{\centering \textbf{TO} $\downarrow$}
    & \multicolumn{1}{p{1cm}}{\centering \textbf{Blk} $\downarrow$} \\
    \cmidrule{1-15} %
    LAV \cite{lav}     && \cmark & \cmark & $48.41\pm3.40$ & $80.71\pm0.84$ & $0.60\pm0.04$ &$\textbf{0.00}$& $0.50$& $0.19$& $0.07$& $0.20$& $0.07$& $0.01$& $\textbf{0.29}$ \\
    TransFuser \cite{chitta2022transfuser} && \cmark & \cmark & $46.20\pm2.57$& $\textbf{83.61}\pm\textbf{1.16}$& $0.57\pm0.00$& $0.29$& $0.38$& $0.28$& $\textbf{0.04}$& $0.07$& $0.07$& $\textbf{0.00}$& $0.32$ \\
    \cmidrule{1-15} %
    WOR \cite{chen2021learning} && \cmark & \xmark & $17.36\pm2.95$& $43.46\pm2.99$& $0.54\pm0.06$& $0.05$& $0.64$& $0.15$& $0.84$& $0.15$& $0.89$& $0.04$& $0.45$\\
    NEAT \cite{Chitta2021ICCV} && \cmark & \xmark & $24.08\pm3.30$& $59.94\pm0.50$& $0.49\pm0.02$& $0.01$& $0.71$& $0.21$& $0.18$& $0.18$& $\textbf{0.00}$& $0.02$& $2.83$\\
    TCP* \cite{wu2022trajectory} && \cmark & \xmark & $42.86\pm0.63$ & $61.83\pm4.19$ & $0.71\pm0.04$&  $0.01$& $0.42$& $0.11$& $\textbf{0.04}$& $0.08$& $0.01$& $0.01$& $0.75$\\
    \cmidrule{1-15} %

    {CaT (w/o Alignment, Coaching, FD)}  && \cmark & \xmark & $39.48\pm0.67$& $60.96\pm1.65$& $0.68\pm0.01$& $0.03$& $1.29$& $0.20$& $0.08$& $0.84$& $\textbf{0.00}$& $0.02$& $1.49$\\

    {CaT (w/o Alignment, Coaching)}  && \cmark & \xmark & $40.64\pm0.98$& $62.45\pm0.46$& $0.67\pm0.01$& $0.02$& $1.07$& $0.33$& $0.14$& $0.38$& $\textbf{0.00}$& $0.02$& $1.11$\\

    CaT (w/o Coaching, FD, SH) && \cmark & \xmark & $44.10\pm0.40$& $65.84\pm5.55$& $0.72\pm0.03$& $0.01$& $0.26$& $0.05$& $0.11$& $0.15$& $\textbf{0.00}$& $\textbf{0.00}$& $0.66$\\
    CaT (w/o Coaching, SH) && \cmark & \xmark & $49.69\pm2.28$ & $81.10\pm0.58$ & $0.64\pm0.02$ & $0.01$  & $0.78$ & $0.03$ & $0.08$ & $0.20$ & $\textbf{0.00}$ & $0.02$ & $0.51$ \\

    CaT (w/o Coaching)  && \cmark & \xmark & $55.55\pm1.41$& 8$1.97\pm2.34$& $0.68\pm0.01$& $0.02$& $0.30$& $0.05$& $0.07$& $\textbf{0.06}$& $\textbf{0.00}$& $0.02$& $0.35$\\
    CaT && \cmark & \xmark & $\textbf{58.36}\pm\textbf{2.24}$ & $78.79\pm1.50$ & $\textbf{0.77}\pm\textbf{0.02}$ & $0.01$  & $\textbf{0.20}$ & $\textbf{0.02}$ & $0.05$ & $0.30$ & $\textbf{0.00}$ & $0.04$ & $0.44$ \\
  \bottomrule
\multicolumn{14}{l}{\textit{Privileged Agents:}} \\
   \toprule
    RL Expert (Roach) \cite{zhang2021end} && - & - & $60.14\pm2.40$& $85.83\pm0.60$& $0.69\pm0.03$& $0.06$& $0.22$& $\textbf{0.00}$& $0.01$& $\textbf{0.00}$& $\textbf{0.00}$& $0.04$& $\textbf{0.07}$\\
    Rule-based Expert && - & - & $71.96\pm2.13$& $77.46\pm3.11$& $\textbf{0.91}\pm\textbf{0.00}$& $\textbf{0.01}$ & $\textbf{0.06}$& $\textbf{0.00}$& $\textbf{0.00}$& $\textbf{0.00}$& $\textbf{0.00}$& $0.03$& $0.41$\\
    \cmidrule{1-15}
    Basic BEV Agent~\cite{lbc} && - & - & $24.08\pm2.83$& $73.36\pm1.08$& $0.31\pm0.06$& $\textbf{0.01}$& $1.45$& $2.45$& $0.09$& $0.27$& $0.11$& $\textbf{0.01}$& $0.24$\\
    + History and Desired Path && -& -& $52.81\pm1.79$ & $79.34\pm3.65$ & $0.71\pm0.06$ & $0.02$& $0.30$& $\textbf{0.00}$& $\textbf{0.00}$& $\textbf{0.00}$& $\textbf{0.00}$&  $0.02$& $0.53$\\
    + Agent Forecast  && - & - & $65.73\pm0.93$& $83.50\pm1.18$& $0.79\pm0.02$& $\textbf{0.01}$& $0.21$& $\textbf{0.00}$ & $0.02$& $\textbf{0.00}$& $\textbf{0.00}$& $\textbf{0.01}$& $0.28$\\
    + Entity Attention && - & - & $\textbf{73.30}\pm\textbf{1.07}$ & $\textbf{87.44}\pm\textbf{0.28}$& $0.83\pm0.02$& $0.05$& $0.11$& $\textbf{0.00}$& $0.02$& $\textbf{0.00}$& $\textbf{0.00}$& $0.02$& $0.12$\\
    \bottomrule
  \end{tabular}}
  \vspace{-0.15in}
\end{table*}

\section{Experiments}
\label{sec:analysis}
We use the CARLA simulator (version 0.9.10.1) \cite{Dosovitskiy2017CORL} for data generation and closed-loop
evaluation of the proposed CaT framework. We also use open-loop evaluation on nuScenes~\cite{caesar2020nuscenes,zhang2022selfd,hu2021safe} (ADE, FDE, and collision rate). In Carla, we leverage the Longest6 Benchmark \cite{chitta2022transfuser}, which uses the six longest routes of each town (Town01 - Town06) from the set of 76 routes provided by the official CARLA leaderboard \cite{carlaleaderboard} (a total of 36 routes). We note in addition to the standard cross weather and time of day generalization evaluation, the long and dense traffic conditions in Longest6 present the most challenging settings across the various CARLA-based benchmarks. 
To evaluate our models, we follow standard metrics and report Route Completion (RC, in terms of completed route percentage), Infraction Score (IS, a penalty factor over infractions), and Driving Score (DS, computed from the prior two)~\cite{carlaleaderboard,lav,chitta2022transfuser}). 

\subsection{Comparison with Prior Methods}
\vspace{-0.1in}
\boldparagraph{CARLA Results} As shown in Table~\ref{tab:leaderboard}, using feature distillation, safety hints-based BEV for the teacher, and the proposed coaching mechanism, CaT is able to obtain a 58.36\% DS. Specifically, we achieve state-of-the-art performance among all prior agents, including LiDAR-based approaches~\cite{lav,chitta2022transfuser} (by $20.6\%$ in terms of DS, from 48.41\% to 58.36\%, and $28.3\%$ in terms of IS). 
We also note that such approaches provide strong baselines, due to various LiDAR-based safety checks that the agents perform on top of the trained policy, which we do not employ. Moreover, CaT outperforms the prior RGB-only state-of-the-art agent TCP~\cite{wu2022trajectory} by 36.16\% DS. To further validate the generalization of CaT, we discuss an additional benchmark split from TCP~\cite{wu2022trajectory} and LAV~\cite{lav} in the supplementary. We also find that removal of the alignment module degrades DS from $44.10\%$ to $39.48\%$, and is shown to only slightly benefit from feature distillation ($40.64\%$ DS) indicating the effectiveness of the proposed architecture.

\begin{table}[!t]
  \caption{\textbf{Open-Loop Evaluation on nuScenes.}}
  \vspace{-0.1in}
  \label{tab:nusc}
  \centering
  \normalsize
  \resizebox{\textwidth/2-20pt}{!}{\begin{tabular}{p{4.2 cm} |ccc}%
    \toprule
     \textbf{Method} 
    & \multicolumn{1}{p{1.8cm}}{\centering \textbf{ADE (m)} $\downarrow$}
    & \multicolumn{1}{p{1.8cm}}{\centering \textbf{FDE (m)} $\downarrow$}
    & \multicolumn{1}{p{1.8cm}}{\centering \textbf{Coll. (\%)} $\downarrow$}\\
    \cmidrule{1-4}
    BEV Agent & 0.33 & 0.52 & 0.49 \\
    \cmidrule{1-4}
    CaT (w/o Coaching, FD, SH) & 0.48 & 0.43 & 0.68\\
    CaT & \textbf{0.41}  & \textbf{0.36} & \textbf{0.27}\\
    \bottomrule
  \end{tabular}}%
  \vspace{-0.15in}
\end{table}

\boldparagraph{Real-World Evaluation} 
To further analyse the benefits of CaT in realistic driving settings, Table~\ref{tab:nusc} shows open-loop evaluation (ADE, FDE, and Collision rate~\cite{zhang2022selfd}) for nuScenes~\cite{caesar2020nuscenes}. On the official validation split, the privileged agent performs best with an ADE of 0.33. CaT achieves 0.41 ADE, improving by 14.6\% gain over an agent without distillation and coaching. Moreover, we find a significant reduction in the collision rate by 60.3\% compared to the baseline. Additional details regarding nuScenes evaluation can be found in the supplementary.

\subsection{Ablation Studies}
\label{exp:ablation}
\vspace{-0.15cm}
\boldparagraph{Teacher Comparison} Table~\ref{tab:leaderboard} depicts our teacher model ablation. Specifically, we find the proposed BEV channels to drastically improve the privileged agent's performance by simplifying the learning task (73.30\% DS, surpassing the built-in rule-based expert of 71.96\%). We note that 
our behavior cloning agent also significantly outperforms an RL-based expert~\cite{zhang2021end} (60.14\%), simply through an effective BEV design. Thus, our findings apply to real-world scenarios where interactive agent training may be unsafe. We also observe that the improved teacher benefits the student agent. Specifically, incorporating the proposed BEV-based safety hints results in student performance gains, from 49.69\% to 55.55\%.

\begin{table}[!t]
  \caption{\textbf{Ablation Study on Feature Distillation Layers.} All loss terms in Eqn.~\ref{eq:featloss} benefit training with deep feature distillation.}
  \label{tab:distill}
  \vspace{-0.2cm}
  \centering
  \normalsize
  \resizebox{2.9in}{!}{\begin{tabular}{p{3.6cm} |ccc}%
    \toprule
     \textbf{Method} 
    & \multicolumn{1}{p{1cm}}{\centering \textbf{DS} $\uparrow$}
    & \multicolumn{1}{p{1cm}}{\centering \textbf{RC} $\uparrow$}
    & \multicolumn{1}{p{1cm}}{\centering \textbf{IS} $\uparrow$} \\
    \cmidrule{1-4}
    No Distillation & 44.10 & 65.84 & 0.72 \\
    \cmidrule{1-4}
    One Layer~\cite{zhang2021end,wu2022trajectory} & 45.23 & 69.33 & 0.68\\
    \cmidrule{1-4}
    Three Layers $\cL_2$ & 49.31 & 66.92 & 0.78  \\
    Three Layers $\cL_2$ + $\cL_{CD}$ & 51.95 & 62.82 & \textbf{0.87}\\
    Three Layers $\cL_{feat}$  & \textbf{55.55}  & \textbf{81.97} & 0.68  \\
    \bottomrule
  \end{tabular}}%
\end{table}

\begin{figure*}
    \centering
    \setlength{\tabcolsep}{2pt}
    \begin{tabular}{ccccc}
    \multicolumn{4}{c}{\textbf{Image}}&
    \multicolumn{1}{c}{\textbf{BEV}}
    \vspace{-0.05cm}
    \\
    \multicolumn{4}{c}{\includegraphics[height=2.2cm]{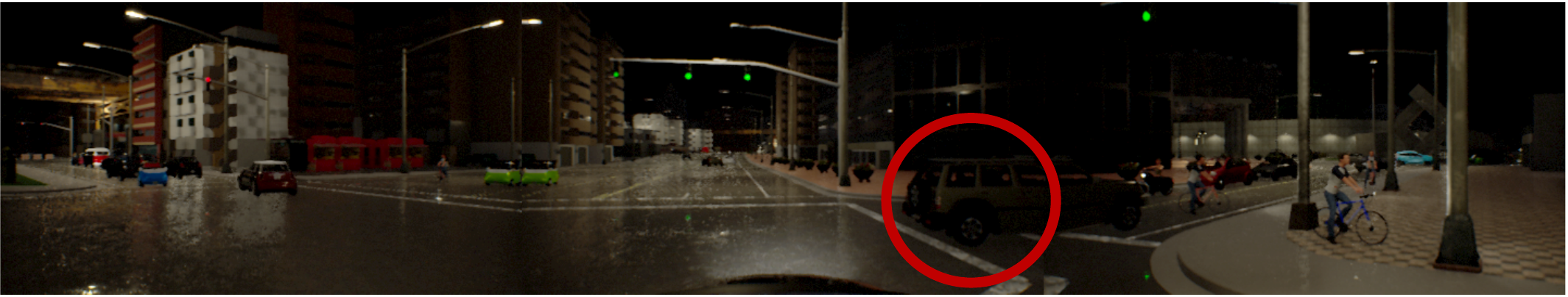}}&
    \multicolumn{1}{c}{\includegraphics[height=2.2cm]{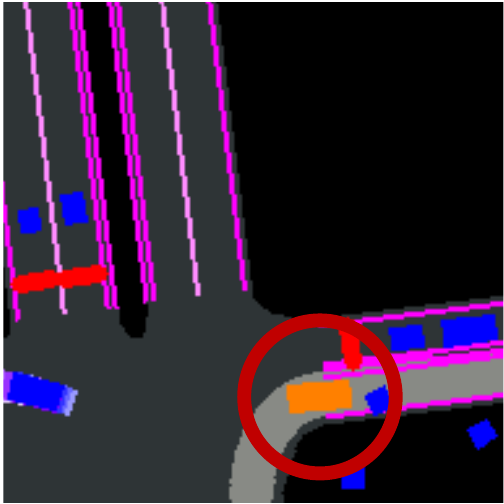}}
    \\
    \textbf{No Alignment Module} & & \textbf{Privileged} & \textbf{Output Distillation} & \textbf{CaT} 
    \\  
  \includegraphics[width=5.5cm,height=2.2cm]{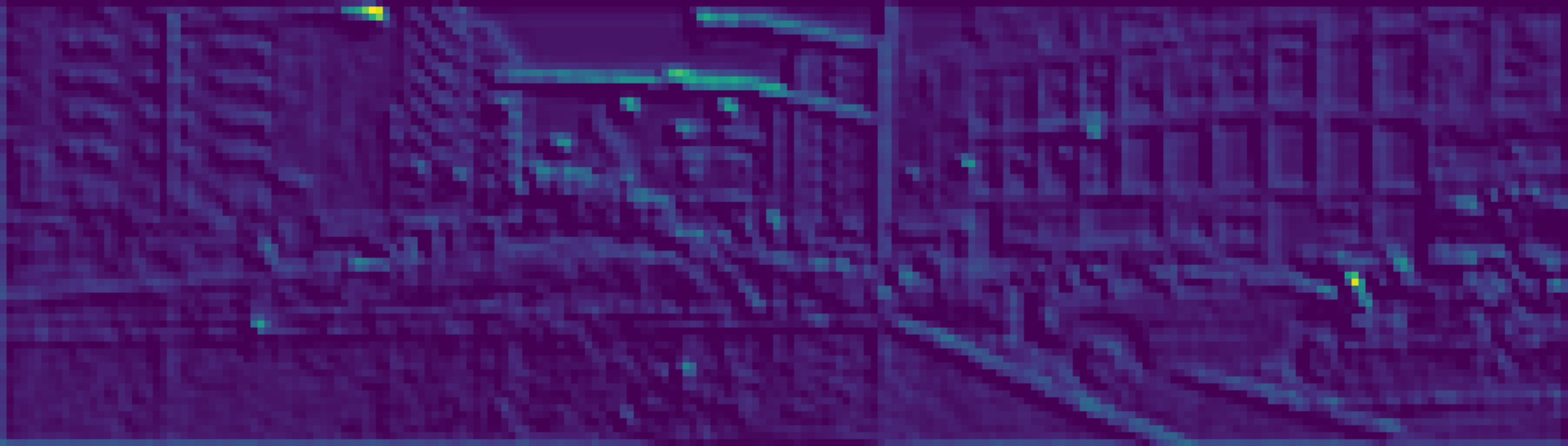} && \includegraphics[width=2.2cm]{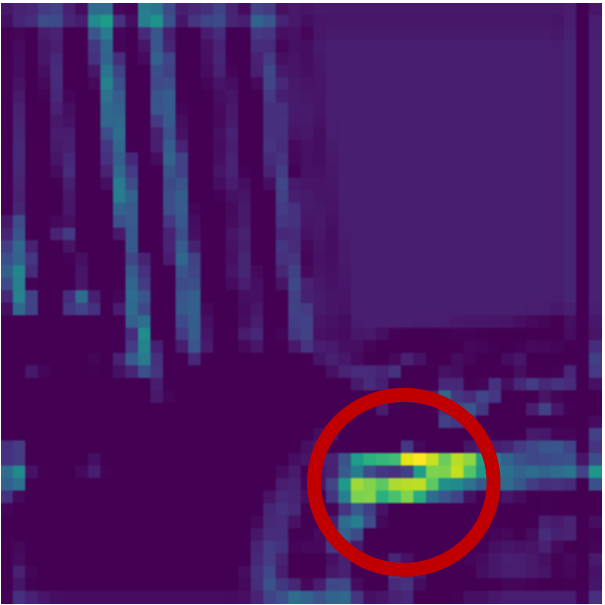} & 
  \includegraphics[width=2.2cm]{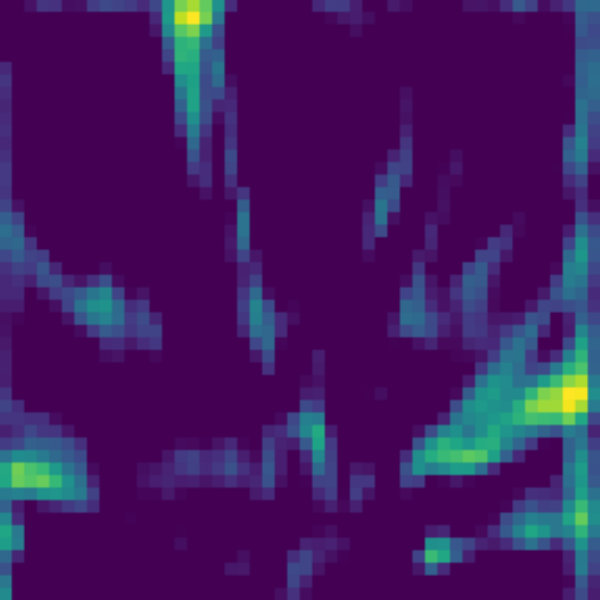} & 
  \includegraphics[width=2.2cm]{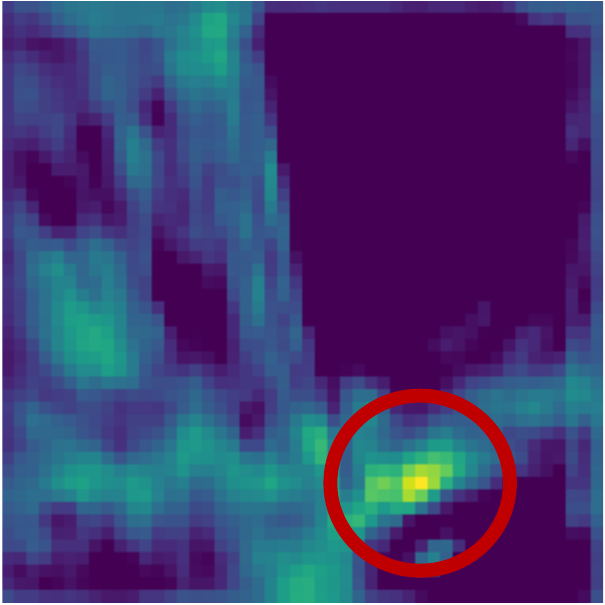} 
    \\

    \includegraphics[width=5.5cm,height=2.2cm]{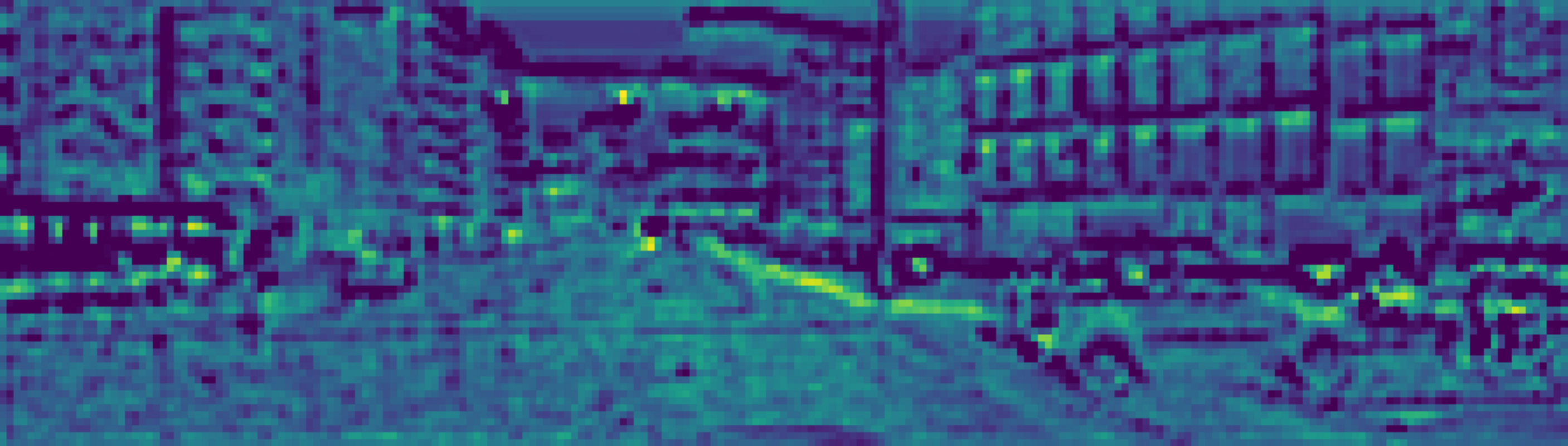} && \includegraphics[width=2.2cm]{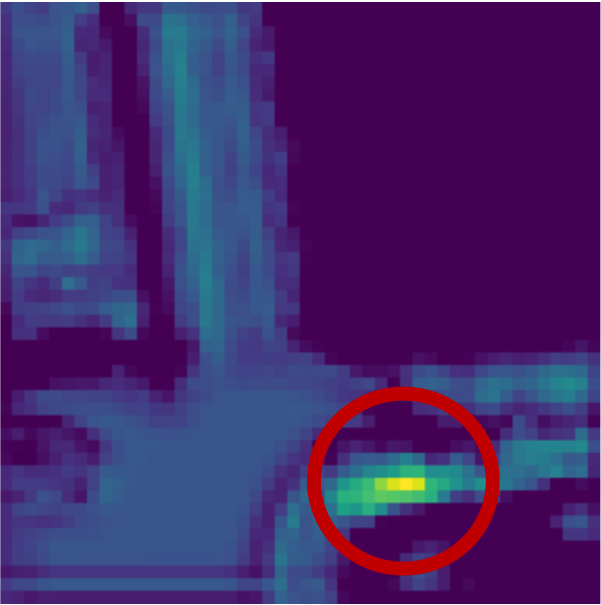} & 
  \includegraphics[width=2.2cm]{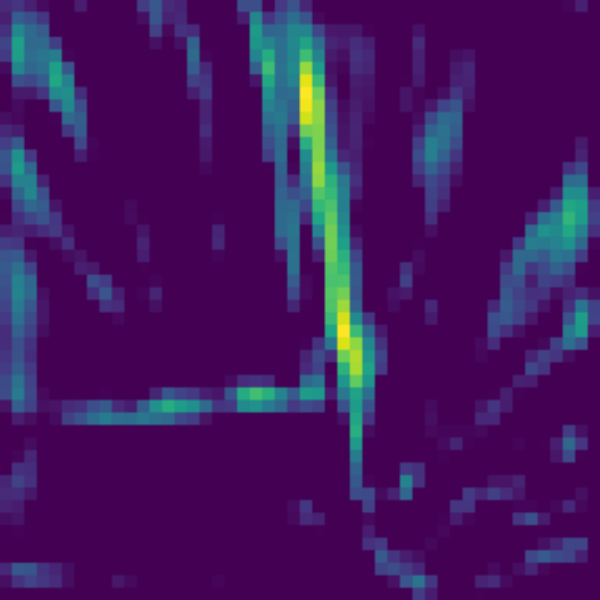} & 
  \includegraphics[width=2.2cm]{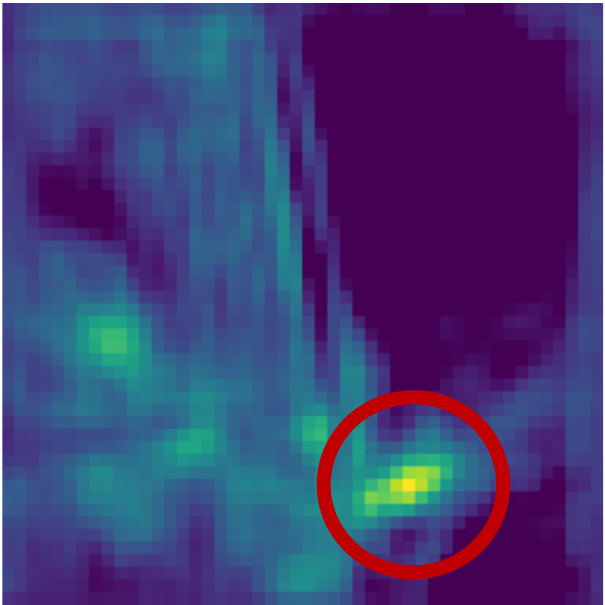} \\
    \end{tabular}
        \vspace{-0.35cm}
    \caption{\textbf{Visualization of Model Input and Features.} We show the BEV ground truth (BEV) and RGB image ground truth (Image) on the top row. Next, we visualize the intermediate layer activations of a sensorimotor agent without the proposed alignment module~\cite{wu2022trajectory} (No Alignment Module), privileged teacher agent (Privileged), student agent trained with output distillation only (Output Distillation), and the proposed agent (CaT). We demonstrate that our teacher agent with proposed knowledge distillation is able to guide the student agent in learning scene-relevant representations and attend to safety-critical entities (circled in red).}
    \label{fig:feature}
    \vspace{-0.35cm}
\end{figure*}

\boldparagraph{Impact of Feature Distillation} 
Table~\ref{tab:distill} presents the benefits of our proposed multi-layer feature distillation framework. We note that prior methods only distill one final fully connected layer (before the prediction head\cite{zhang2021end,wu2022trajectory}, referred to as `one layer' in Table~\ref{tab:distill}). We note that such approaches result in minimal gains, whereas CaT is able to benefit from the extensive feature distillation (improving from 45.23\% to 55.55\%). Incorporating multiple types of feature distillation losses also benefits performance.

\subsection{Feature Visualization}
Fig.~\ref{fig:feature} depicts the learned features of the initial feature distilled layer (out of the three in Fig.~\ref{fig:overview}). We visualize activations from two channels, to better understand the learned representation by the agents. Interestingly, the visualization shows how traditionally trained sensorimotor agents (\ie, without the proposed alignment module) mostly operate in image-space features in this layer despite the BEV waypoint prediction task. This validates our main hypothesis, that such agents have limited ability to learn from a BEV-based agent, even with BEV-based auxiliary prediction tasks. We also find that output distillation provides a weak supervision for critical components of the BEV, including layout and object features. After feature distillation, the features of our CaT model are able to not only keep the driving-related semantic information \ie, drivable road and route, but also focus on safety-critical entities. More visualization examples are provided in the supplementary.

\section{Conclusion}
\label{conclusion}
We present CaT, a novel knowledge distillation method for effectively training a sensorimotor student agent using the supervision from a privileged agent. We leverage an alignment module to better map image features to the BEV space, thus enabling extensive supervision from a BEV-based teacher over the intermediate feature learning. As both agents can be trained offline via imitation learning, our findings are directly relevant to real-world settings where on-policy interactions may be unsafe to perform. 
Finally, we carefully account for inherent differences between the student and the teacher using a student-paced coaching mechanism with various auxiliary supervision tasks. Through the improved knowledge distillation process, our experiments result in a state-of-the-art camera-based agent in CARLA. Our ablation into input and model design guides future directions through interpretable analysis into current limitations of end-to-end driving models. 

\boldparagraph{Acknowledgments}
We thank the Red Hat Collaboratory for supporting this research.

{\small
\bibliographystyle{ieee_fullname}
\bibliography{egbib}
}

\section*{Appendix A. Training Data}
This section provides details regarding our training data collection process and the proposed BEV design.
\subsection*{A.1. Data Collection}
Our training dataset is collected by employing a standard rule-based expert (based on Chitta~\etal~\cite{chitta2022transfuser}). The expert comprises two main components: a PID-based longitudinal and lateral controller (oversampled waypoints from an A$^{*}$ planner) and a set of carefully tuned heuristics to avoid collisions and infractions. The goal of the privileged neural network agent is to learn to imitate the expert-generated waypoints, \ie, plan based on BEV observations. For training, our data is generated from $3{,}200$ predefined short routes spanning all eight towns. We emphasize that the training routes do not overlap with the testing routes (discussed in Sec.~\ref{supsec:eval}) to evaluate generalization. We generate the data at four FPS (Frames Per Second) and randomly shuffle weather and daylight conditions at five-frame intervals. For our camera configuration, we set three RGB cameras at three yaws ($0^{\circ}$, $-60^{\circ}$, and $60^{\circ}$). Each camera's resolution is set as $480\times 270$ with a  $64^{\circ}$ field of view. Examples of the resulting camera views can be seen in Fig. 2 and Fig. 4 of the main paper as well as in our supplementary qualitative results (Sec.~\ref{sec:suppex}).

\subsection*{A.2. BEV with Safety Hints}
Rule-based experts commonly employ a kinematic bicycle model~\cite{polack2017kinematic,chitta2022transfuser,hanselmann2022king,chen2021learning} for predicting possible future collisions with dynamic objects. Despite its simplicity, the kinematics model can be used to efficiently extrapolate future positions and orientations of the ego and surrounding agents and detect an immediate collision risk~\cite{chen2021learning}. While prior work leverages the kinematics model to generate improved expert trajectories, we propose to further incorporate its output for generating safety-driven BEV features for the privileged agent. We hypothesize that the additional BEV features can not only guide privileged agent training (which we show is complex in itself) but also facilitate a more effective teacher model for the sensorimotor student, \ie, due to the improved reasoning over safety and task-relevant components in intricate and dense scenes. As discussed in the main paper, our proposed BEV features result in significant impact on the performance of both the privileged and student agents.

\boldparagraph{Agent Forecast Channels} Our goal is to effectively capture immediate collision and infraction risk for the ego-agent. Thus, our agent forecast channels involve direct sampling from the kinematics model. Specifically, we sample the future states over a horizon of $2.5$s for all surrounding vehicles based on their expert control signals and the bicycle model. We then incorporate the extrapolation as additional input channels for the privileged agent. The agent forecasts channels can therefore help the trained agents reason over potential future collisions. 

\boldparagraph{Entity Attention Channel} To ease correlating observations with actions during training, we also propose to incorporate an explicit \textbf{entity attention channel}. This proposed cue is derived from the objects in the BEV as well as the kinematics model applied to the ego-vehicle. Specifically, we incorporate any object in the current frame with a potential future collision with the ego-vehicle. By improving reasoning over the intentions of other agents, we demonstrate the entity attention cue to reduce collisions and blocks of other agents. As shown in Table 1 of the main paper, a privileged agent trained with object history, desired path, and agent forecast BEV input does not properly learn to avoid collision with vehicles in dense settings (0.21 vehicle collision rate). We find that adding the entity attention channel reduces this rate by nearly half, to 0.11 collision rate. In our implementation, we also sought to leverage this channel to ensure safe interaction with red traffic lights along the path. We do not find this additional cue to provide clear benefits (potentially due to the already low incidence of such events).  Nonetheless, our work suggests an important future direction of input feature design and privileged agent training, \eg, using more sophisticated prediction models~\cite{rudenko2020human,choi2019looking}.

\section*{Appendix B. Implementation}
This section details the training protocol of the proposed models. We also provide additional details regarding the computation of the feature distillation loss terms in Eqn. 7 of the main paper.

\boldparagraph{Training Parameters}
We train the privileged agent for 80 epochs with Adam~\cite{kingma2014adam}. We use a batch size of 512 and an initial learning rate of \num{3e-4}. The learning rate
is reduced by half every 30 epochs.
The sensorimotor CaT model is trained with Adam~\cite{kingma2014adam} for 50 epochs. We use a batch size of 48 and an initial learning rate of \num{1e-3}. The learning rate is reduced by half every 20 epochs. We opt for a simple linear schedule over the $\lambda_i$ (Eqn 8 of the main paper). $\lambda_0$ is initialized at $0.5$ and is linearly reduced over training epochs to $0$ at epoch 15. After the initial coaching stage, training is reduced to standard distillation training over the entire dataset to ensure the final model learns correct predictions.

\boldparagraph{Feature Distillation Loss} Perfect BEV prediction (\eg, 3D distances, occluded regions) is a notoriously challenging task. Thus, while our proposed alignment module can facilitate direct ${L}_2$ matching between internal features of the student and the BEV-based teacher, exact feature matching can be difficult. Moreover, the two agents should not leverage identical processing mechanisms, \ie, due to the varying inputs and the inherent uncertainty in perception-based decision-making. To better capture the noisy BEV feature prediction task, we incorporate two additional auxiliary terms in Eqn. 7 of the main paper. As shown in Table 2 of the main paper, the two terms holistically combine to improve over basic direct feature matching with an ${L}_2$ loss. To account for inherent noise and differences in the underlying feature extraction processes between the two agents, we compute an additional loss term (2$^{nd}$ term in Eqn. 7 of the main paper) by introducing \textit{three feature distillation modules} (see our Figure 2 overview in the main paper), one for each of the three overall distillation feature maps. Each feature distillation module first performs bilinear upsampling for the original feature map. Next, the feature map is processed with four, six, and eight $3\times3$ convolutional layers (each followed by batch normalization and ReLU), for the first, second, and third feature distillation map, respectively (\ie, $\cF_1$, $\cF_2$, $\cF_3$ in Eq. 7 of the main paper, with the third feature map having the lowest resolution). We then compute a second ${L}_2$ loss term over the outputs of the distillation models as part of $\cL_{feat}$. The distillation modules enable more effective knowledge distillation by guiding (\ie, instead of exactly matching with the highly privileged teacher) the student's internal features.

\boldparagraph{Chamfer Distance (CD) Loss} Given the aligned feature maps, we also explore the benefits of incorporating further regularization during the feature distillation process, computed over the feature maps. Intuitively, while the student and teacher feature maps may not have to perfectly align, layout and coarse object properties should be preserved. As fine-grained BEV details and distances may be difficult to obtain for the student, the Chamfer Distance~\cite{paschalidou2021neural} provides a comparison mechanism beyond ${L}_2$ that can better capture underlying BEV feature structure. We leverage a Spatial Soft Argmax~\cite{levine2016end,lbc} \textit{over the distilled feature maps} to extract a set of feature keypoints. We empirically found a small set of 10 keypoints (from each feature map) to be sufficient. Given a set of 10 student feature map keypoints $\bP_{kp}^s$ and 10 teacher feature map keypoints $\bP_{kp}^t$, the Chamfer distance can be computed via
 \begin{equation*}
     \mathcal{L}_{CD}=\sum_{\bp^{s}\in \bP_{kp}^{s}} \min_{\bp^{t}\in \bP_{kp}^{t}}\|\bp^{s}-\bp^{t}\|^2_2+
     \sum_{\bp^{t}\in \bP_{kp}^{t}} \min_{\bp^{s}\in \bP_{kp}^{s}}\|\bp^{s}-\bp^{t}\|^2_2
\end{equation*}
We empirically demonstrate this term to further regularize the challenging feature matching task of the student.

\section*{Appendix C. Evaluation Benchmark}
\label{supsec:eval}
\boldparagraph{Longest6 Benchmark}
Our ablation studies are conducted using the challenging Longest6 benchmark (the offline counterpart of the CARLA Leaderboard~\cite{carlaleaderboard,chitta2022transfuser}). In contrast to prior CARLA-based benchmarks, the Longest6 settings incorporate longer driving routes. Moreover, surrounding vehicles (including bicycles) are spawned at \textit{all of the possible spawning points defined by the CARLA map} at the onset of each episode. This introduces dense and difficult traffic conditions (even for privileged agents, as demonstrated by our analysis) while also resulting in harsher evaluation settings compared to the official leaderboard. Pedestrian scenarios are designed to stress test model safety and include frequent and unexpected crossings. Each test route is evaluated with unique ambient conditions, \ie, using a unique combination of weather (cloudy, wet, mid-rain, wet-cloudy, hard-rain, soft-rain) and time of day (night, twilight, dawn, morning, noon, sunset). Multiple traffic scenarios from the official CARLA Leaderboard are included during evaluation, \ie, control loss, dynamic object avoidance, and traffic negotiation. The scenarios are triggered at predefined location along the testing routes.

\section*{Appendix D. Additional Analysis}
\label{sec:suppex}

\begin{table}[!t]
  \caption{\textbf{Evaluation on Town02 \& Town05 Benchmark.}}
  \label{tab:0205}
  \centering
  \normalsize
  \resizebox{\textwidth/2-50pt}{!}
  {\begin{tabular}{p{2 cm} |ccc}%
    \toprule
     \textbf{Method} 
    & \multicolumn{1}{p{1cm}}{\centering \textbf{DS} $\uparrow$}
    & \multicolumn{1}{p{1cm}}{\centering \textbf{RC} $\uparrow$}
    & \multicolumn{1}{p{1cm}}{\centering \textbf{IS} $\uparrow$} \\
    \cmidrule{1-4}
    LAV~\cite{lav} & 45.20 & 91.55 & 0.49  \\
    TCP~\cite{wu2022trajectory} & 57.01 & 85.27 & 0.67\\
    \cmidrule{1-4}
    CaT & \textbf{66.70}  & \textbf{92.14} & \textbf{0.73}  \\
    \bottomrule
  \end{tabular}}%
\end{table}

\boldparagraph{Impact of auxiliary tasks} In Table~\ref{tab:auxiliary}, we further study the impact of different auxiliary task losses in Eqn.~\ref{eqn:loss}. We find that CaT benefits from the proposed auxiliary tasks. Specifically, the command prediction task improves the DS by $5.5\%$ (from 48.07\% to 50.70\%) while segmentation task improves DS by $10.7\%$ to 53.23\%. We observe that our method benefit from the proposed multiple auxiliary tasks.

\boldparagraph{Discrepancies} The discrepancies in Table~\ref{tab:leaderboard} from results reported in ~\cite{chitta2022transfuser} are due to us re-running the up-to-date github models released by authors, ensuring fair comparison and complete reproducibility. We note that we train the ensemble-based TCP agent using the publicly available code provided by the authors (there is no pre-trained model).

\begin{table}[!t]
  \caption{\textbf{Auxiliary Tasks Ablation.}}
  \label{tab:auxiliary}
  \centering
  \normalsize
  \resizebox{3.2in}{!}
  {\begin{tabular}{p{5 cm} |ccc}%
    \toprule
     \textbf{Method} 
    & \multicolumn{1}{p{1cm}}{\centering \textbf{DS} $\uparrow$}
    & \multicolumn{1}{p{1cm}}{\centering \textbf{RC} $\uparrow$}
    & \multicolumn{1}{p{1cm}}{\centering \textbf{IS} $\uparrow$} \\
    \cmidrule{1-4}
    
    CaT (w/o Coaching, $\mathcal{L}_{seg}$, $\mathcal{L}_{{cmd}}$) & $48.07\pm2.30$ & $73.46\pm0.87$ & $0.70\pm0.02$  \\
    
    CaT (w/o Coaching, $\mathcal{L}_{seg}$) & $50.70\pm2.87$ & $82.41\pm0.78$ & $0.64\pm0.04$\\

    CaT (w/o Coaching, $\mathcal{L}_{{cmd}}$) & $\textbf{53.23}\pm\textbf{1.11}$ & $\textbf{82.97}\pm\textbf{1.07}$ & $\textbf{0.66}\pm\textbf{0.00}$ \\

    \bottomrule
  \end{tabular}}%
\end{table}

\boldparagraph{Additional Benchmark}
To further validate the generalization of CaT, we discuss an additional benchmark split from TCP~\cite{wu2022trajectory} and LAV~\cite{lav} in Table~\ref{tab:0205} where Town02 and Town05 are not seen in training. CaT achieves a $66.7\%$ DS ($92.1\%$ RC) on this benchmark, which is $12.8\%$ higher than an ensemble TCP model and $47.6\%$ higher than LAV.

\boldparagraph{Different Coaching Mechanisms}
In addition to the coaching mechanism in the main paper, we investigate two alternative training strategies as shown in Table~\ref{tab:coach}. First, instead of initial coaching with a $\lambda_i$ schedule followed by regular training, we implement a periodic coaching strategy where we fix $\lambda_i = 0.5$ but alternate between coaching and traditional distillation. This strategy is analogous to real-world instruction, where the student may alternate between attempting tasks on their own followed by the coach intermittently intervening to provide easier targets. We find this strategy to degrade student model performance, potentially due to the difficulty of learning in the early stages of training. Moreover, emphasizing hard samples, \ie, as in traditional hard negative mining, can degrade performance (44.92\% DS vs. 55.55\% DS without hard example mining). This affirms our hypothesis that the student can benefit from incremental scaffolding due to the difficulty of the sensorimotor learning task. Table~\ref{tab:coach} also investigates the impact of the initial weight $\lambda_0$ in our student-paced coaching algorithm. A higher value for $\lambda_0$ translates to a more student-centered (but noisy) targets. We find $\lambda_0=0.5$ to provide the best driving performance. Increasing the $\lambda_0$ further hinders performance as the student cannot get sufficient supervision from the teacher at the crucial early stage of training. We note that this proposed mechanism is more efficient than DAGGER-based coaching as done by He~\etal~\cite{he2012imitation}. Our approach does not require potentially dangerous on-policy interactions either.

\begin{table}[!t]
  \caption{\textbf{Coaching Method Ablation.} We compare our student-paced coaching algorithm with a periodic coaching strategy (Periodic, repeatedly alternating between $\lambda_i=0.5$ and $\lambda_i=0$) as well as a hard mining strategy, where the student agent is exposed to more samples at high difficulty. We also depict the impact of the initial weight $\lambda_0$ in our student-paced coaching algorithm showing $\lambda_0=0.5$ to give best performance.}
  \label{tab:coach}
  \centering
  \normalsize
\begin{tabular}{p{3.7cm} |ccc}%
    \toprule
    \textbf{Method} 
    & \multicolumn{1}{p{0.9cm}}{\centering \textbf{DS} $\uparrow$}
    & \multicolumn{1}{p{0.9cm}}{\centering \textbf{RC} $\uparrow$}
    & \multicolumn{1}{p{0.9cm}}{\centering \textbf{IS} $\uparrow$} \\
    \cmidrule{1-4}
    Periodic & $48.97$ & $75.16$ & $0.70$  \\
    \cmidrule{1-4}
    Hard Mining & $44.92$ & $73.78$ & $0.64$  \\
    \cmidrule{1-4}
    Student-paced ($\lambda_0=0$) & $55.55$ & $81.97$ & $0.68$ \\
    Student-paced ($\lambda_0=0.5$) & $\textbf{58.36}$ & $\textbf{78.79}$ & $\textbf{0.77}$ \\
    Student-paced ($\lambda_0=0.8$) & $51.76$ & $79.16$ & $0.69$  \\
    \bottomrule
  \end{tabular}%
\end{table}

\begin{figure*}
    \centering
    \setlength{\tabcolsep}{2pt}
    \begin{tabular}{ccccc}

    \multicolumn{4}{c}{\textbf{Image}}&
    \multicolumn{1}{c}{\textbf{BEV}}
    
    \\
    \multicolumn{4}{c}{\includegraphics[height=2cm]{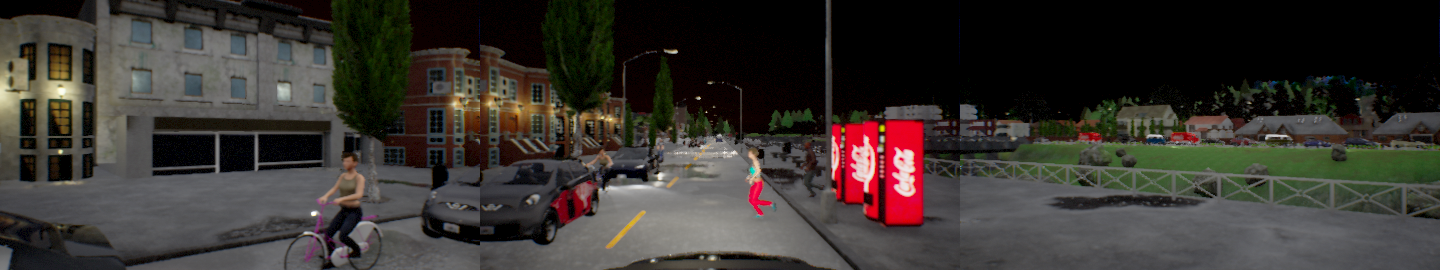}}&
    \multicolumn{1}{c}{\includegraphics[height=2cm]{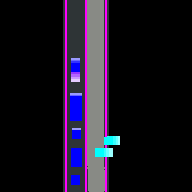}}
 
    \\
      \textbf{No Alignment Module} & & \textbf{Privileged} & \textbf{Output Distillation} & \textbf{CaT} 
    \\  
  \includegraphics[width=5.3cm,height=2cm]{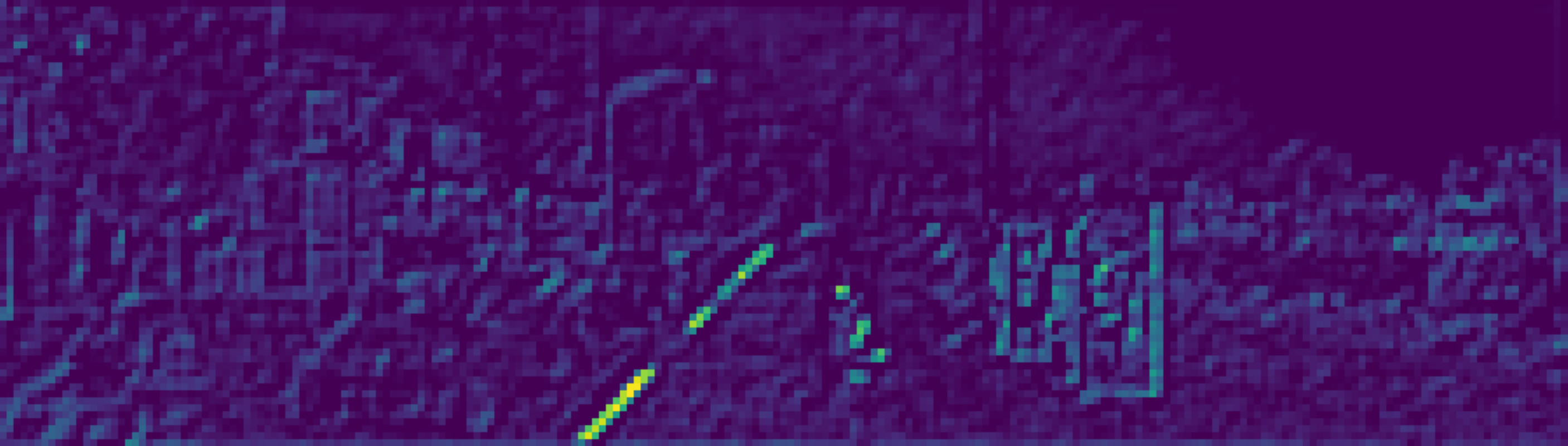} && \includegraphics[width=2cm]{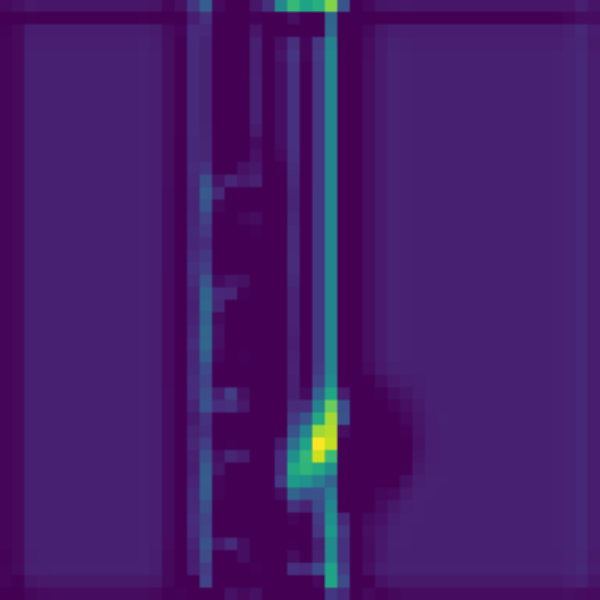} & 
  \includegraphics[width=2cm]{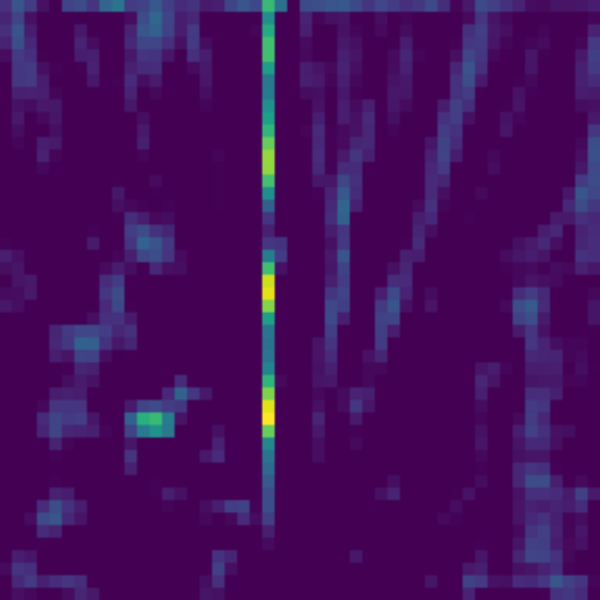} & 
  \includegraphics[width=2cm]{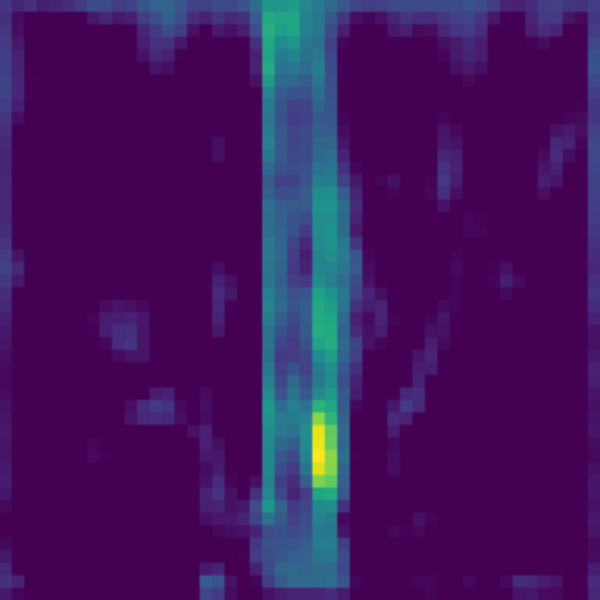} 
    \\

  \includegraphics[width=5.3cm,height=2cm]{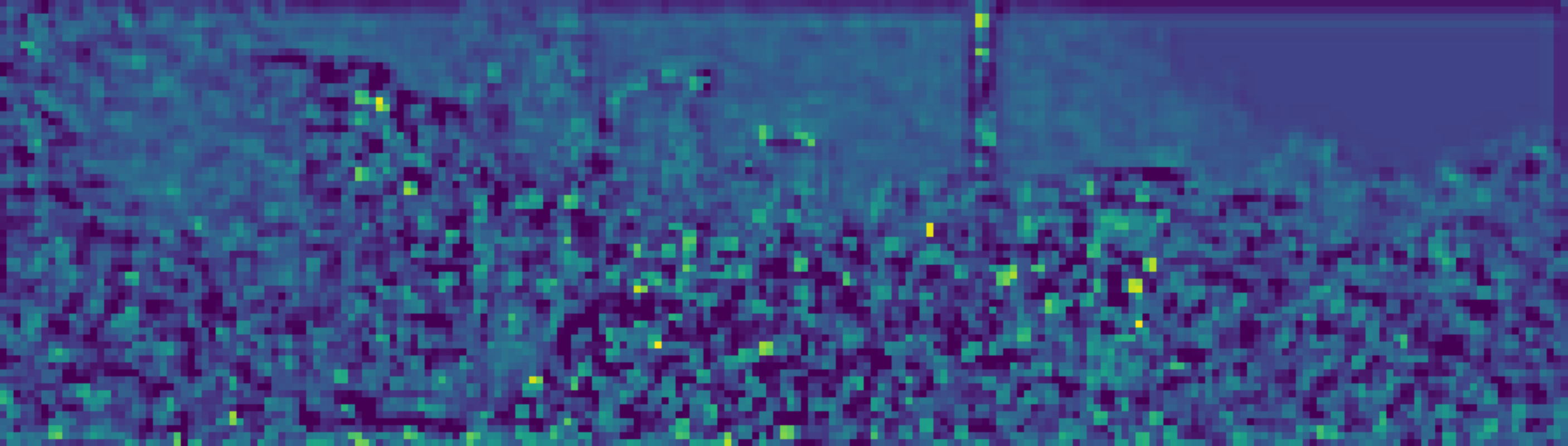} && \includegraphics[width=2cm]{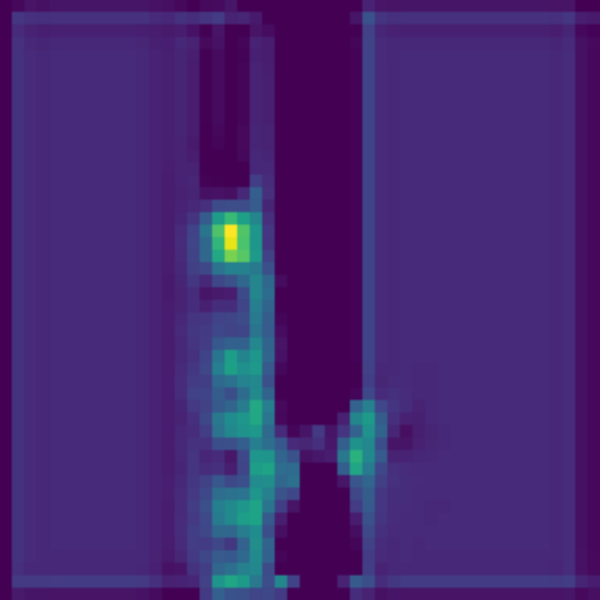} & 
  \includegraphics[width=2cm]{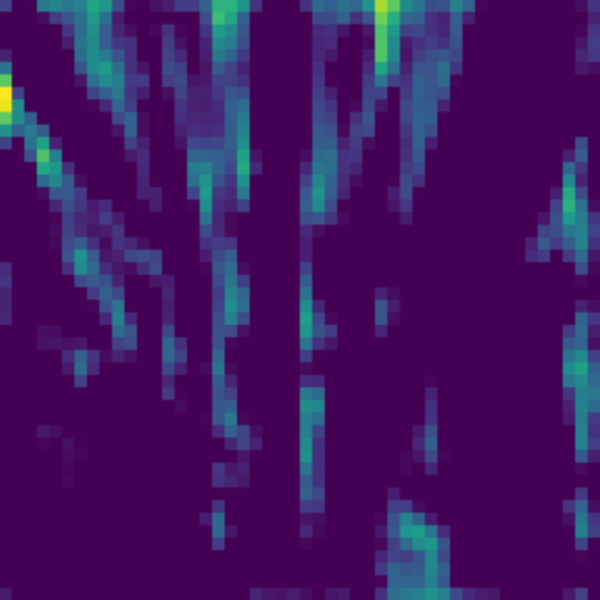} & 
  \includegraphics[width=2cm]{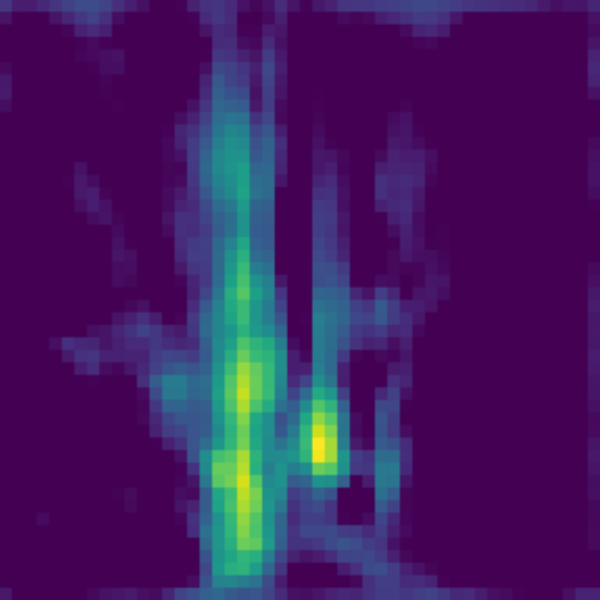} 
    \\
    \midrule

    \includegraphics[width=5.3cm,height=2cm]{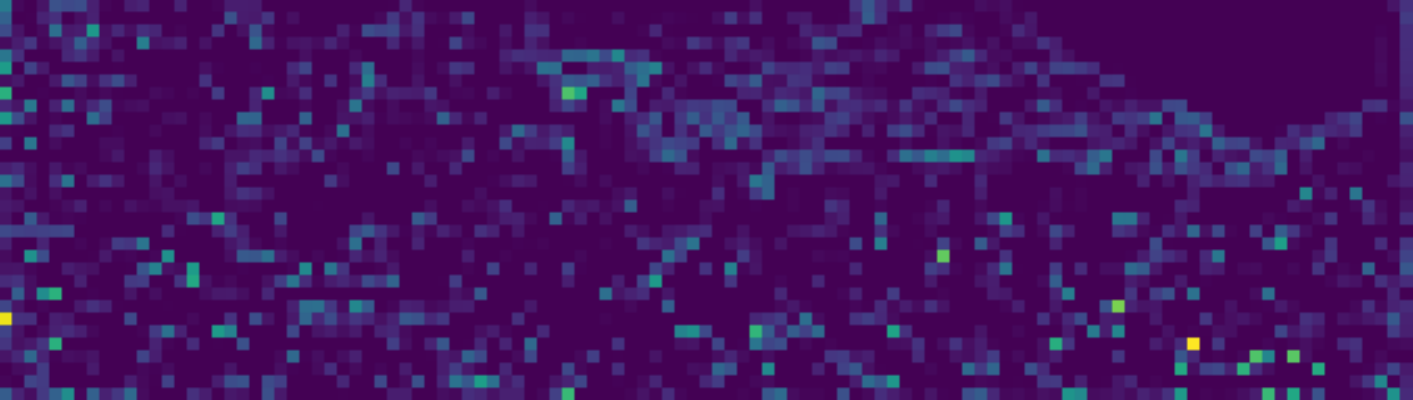} && \includegraphics[width=2cm]{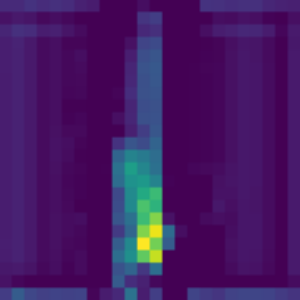} & 
  \includegraphics[width=2cm]{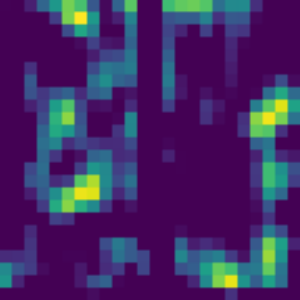} & 
  \includegraphics[width=2cm]{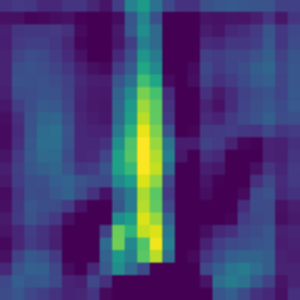} 
    \\
    
    \includegraphics[width=5.3cm,height=2cm]{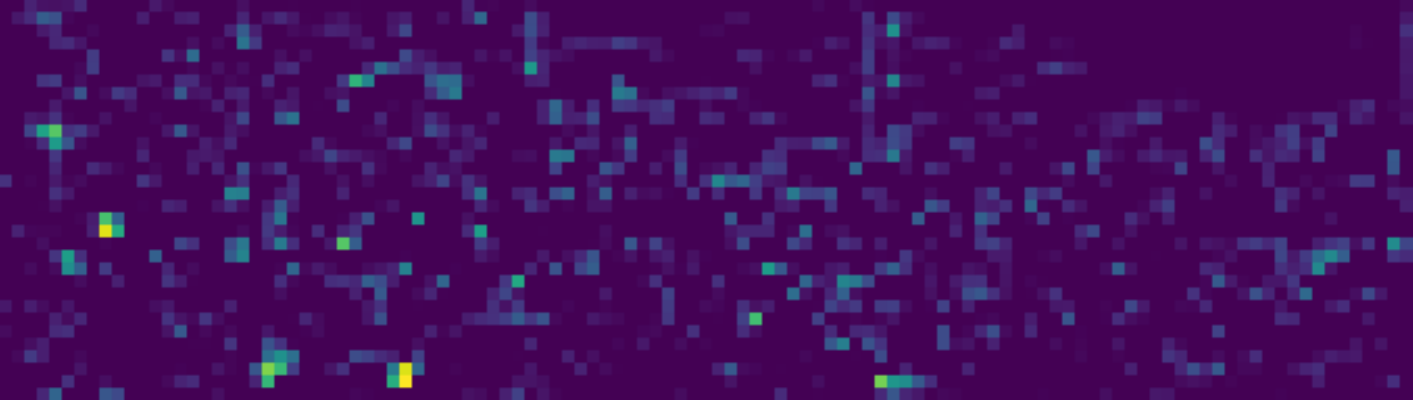} && \includegraphics[width=2cm]{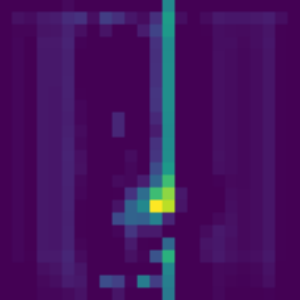} & 
  \includegraphics[width=2cm]{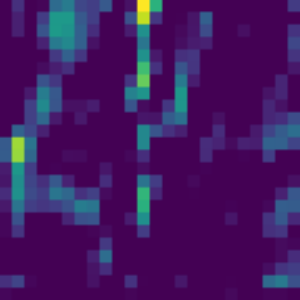} & 
  \includegraphics[width=2cm]{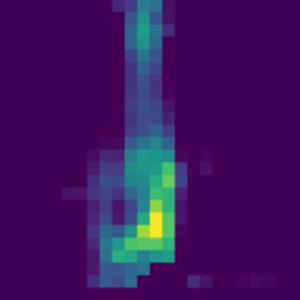} 
    \\
    \end{tabular}
    \caption{\textbf{Visualization of Model Input and Features.} Analysis of night-time driving with a pedestrian abruptly emerging from the right. We show the BEV ground truth (BEV) and RGB image ground truth (Image) on the top row. Next, we visualize the first distillation feature map $\cF_1$ (the second and third rows) and the second distillation feature map $\cF_2$ (the fourth and fifth rows) of a sensorimotor agent without the proposed alignment module (\ie, a baseline traditional RGB model~\cite{wu2022trajectory,lbc}) (No Alignment Module), privileged teacher agent (Privileged), student agent trained with output distillation only (Output Distillation), and the proposed agent (CaT).}
    \label{fig:feature_a}
\end{figure*}

\begin{figure*}
    \centering
    \setlength{\tabcolsep}{2pt}
    \begin{tabular}{ccccc}

    \multicolumn{4}{c}{\textbf{Image}}&
    \multicolumn{1}{c}{\textbf{BEV}}
    
    \\
    \multicolumn{4}{c}{\includegraphics[height=2cm]{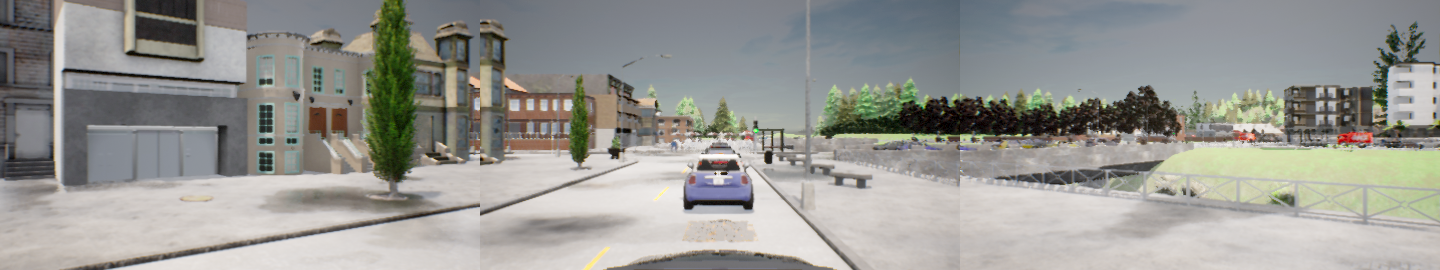}}&
    \multicolumn{1}{c}{\includegraphics[height=2cm]{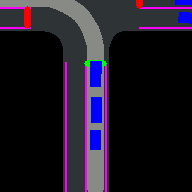}}
 
    \\
      \textbf{No Alignment Module} & & \textbf{Privileged} & \textbf{Output Distillation} & \textbf{CaT} 
    \\  
  \includegraphics[width=5.3cm,height=2cm]{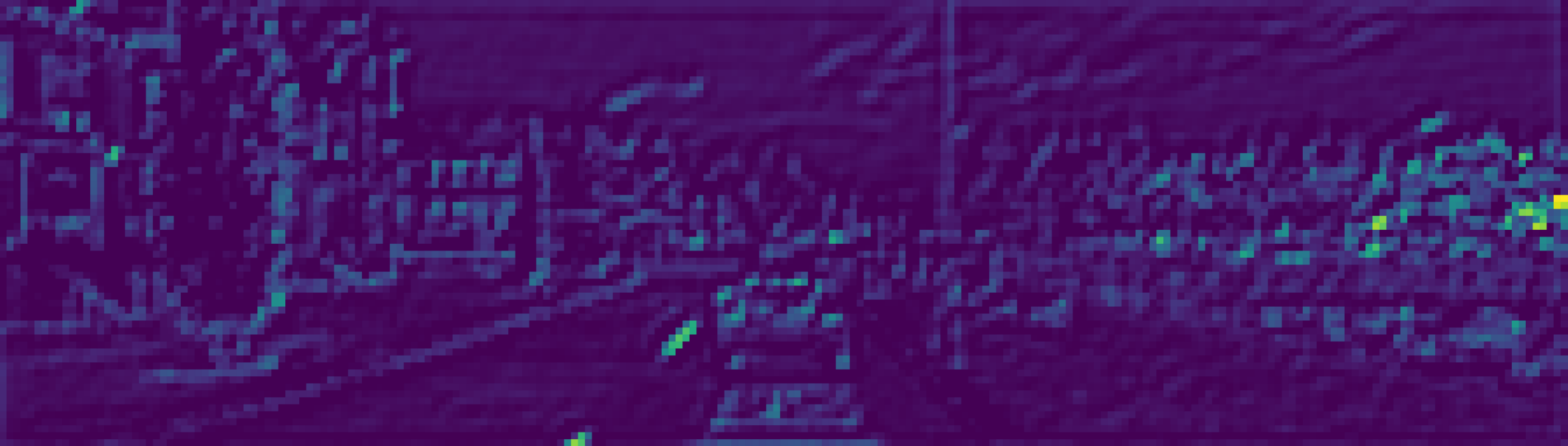} && \includegraphics[width=2cm]{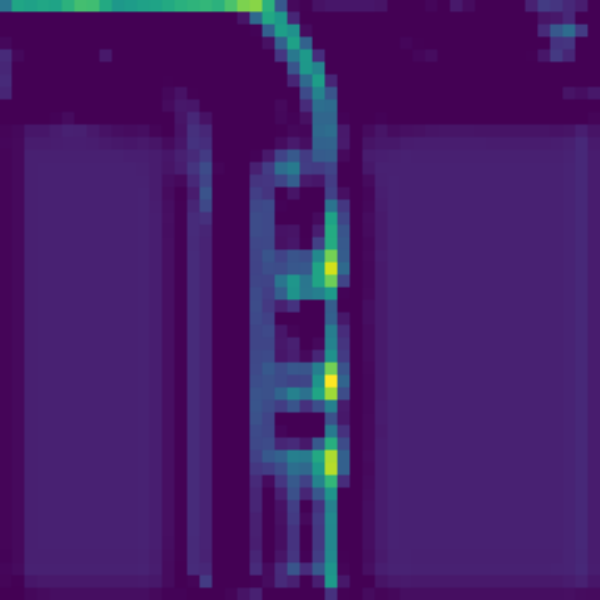} & 
  \includegraphics[width=2cm]{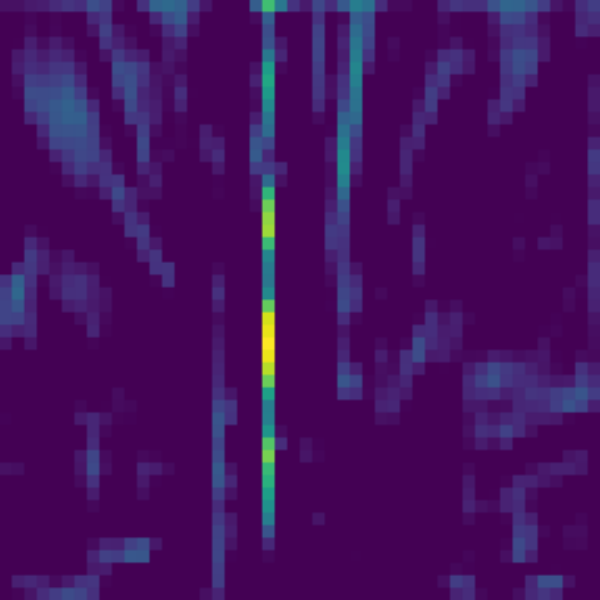} & 
  \includegraphics[width=2cm]{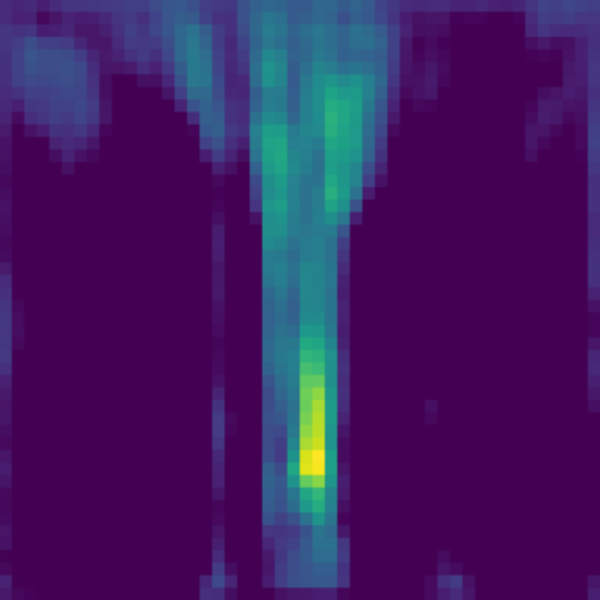} 
    \\

  \includegraphics[width=5.3cm,height=2cm]{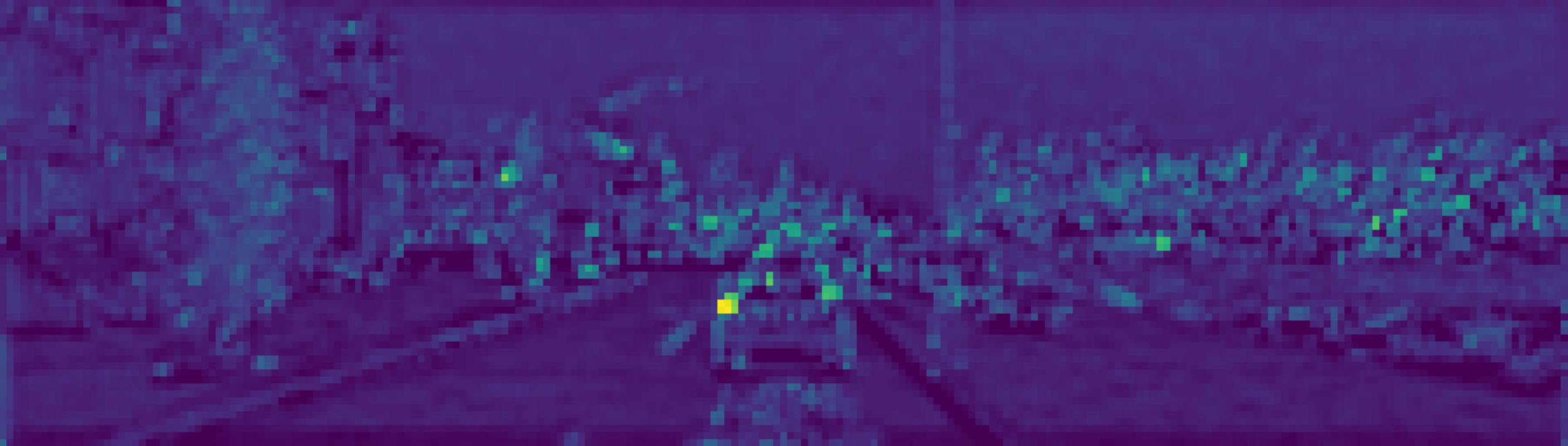} && \includegraphics[width=2cm]{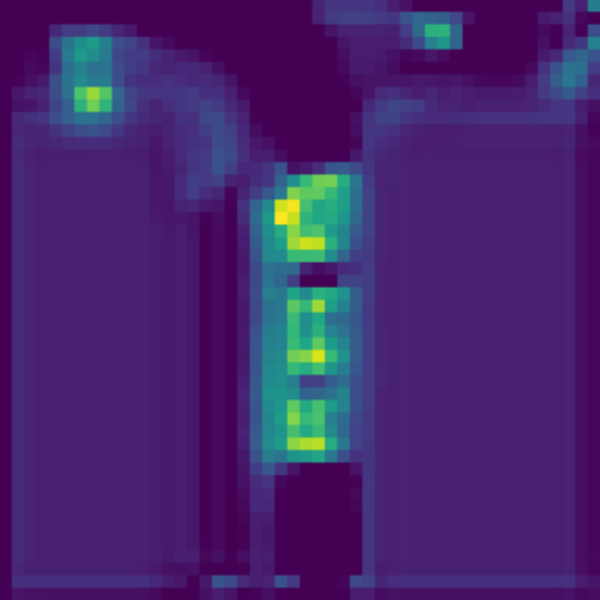} & 
  \includegraphics[width=2cm]{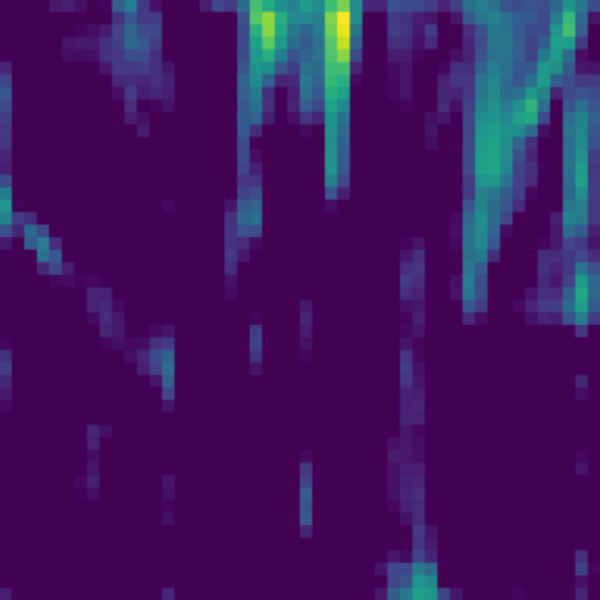} & 
  \includegraphics[width=2cm]{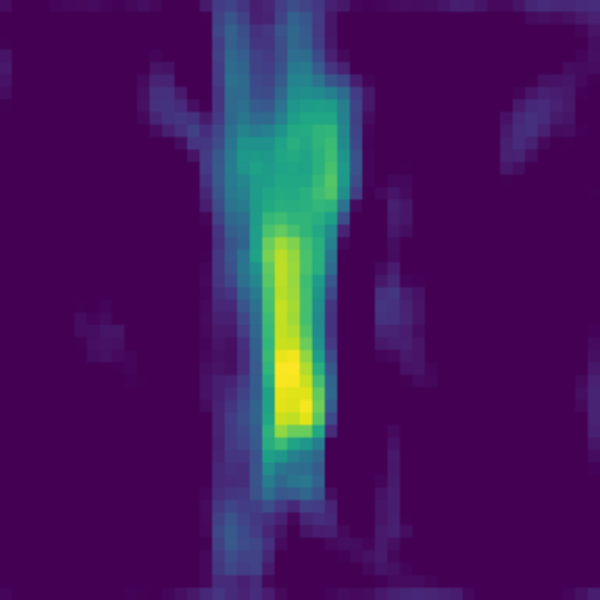} 
    \\
    
    \midrule
    
  \includegraphics[width=5.3cm,height=2cm]{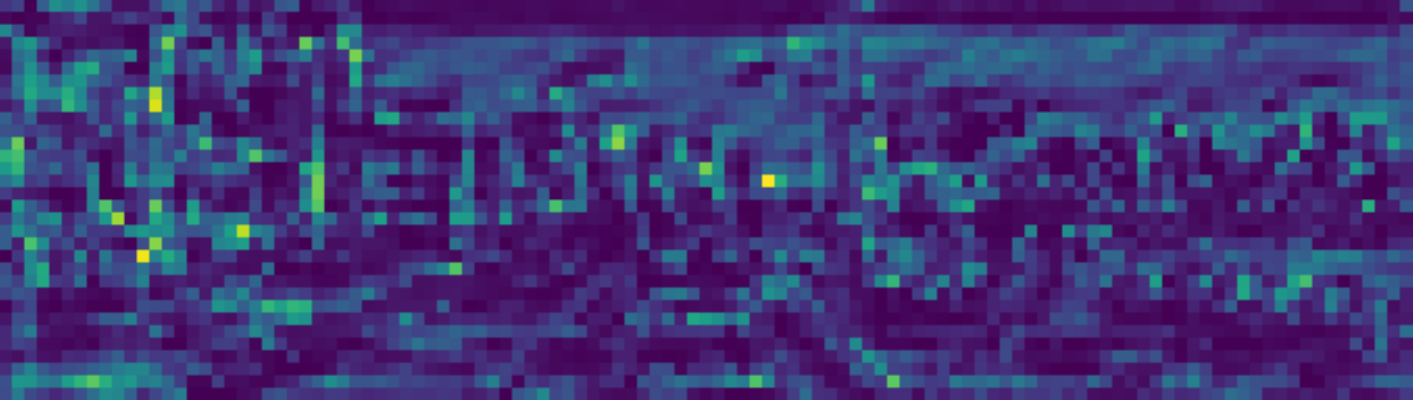} && \includegraphics[width=2cm]{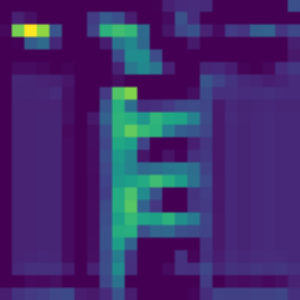} & 
  \includegraphics[width=2cm]{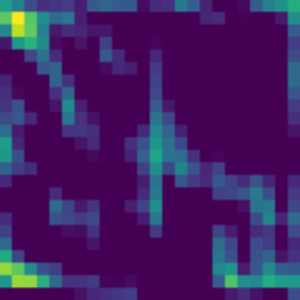} & 
  \includegraphics[width=2cm]{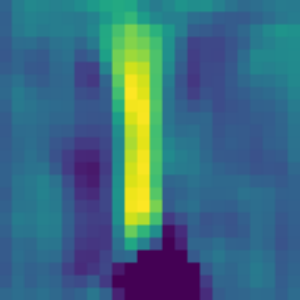} 
    \\

  \includegraphics[width=5.3cm,height=2cm]{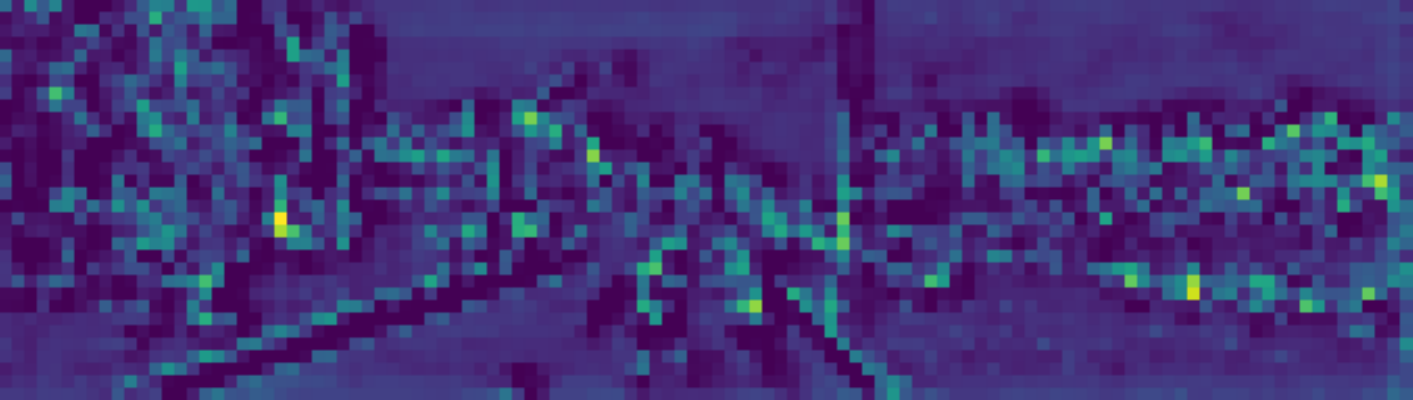} && \includegraphics[width=2cm]{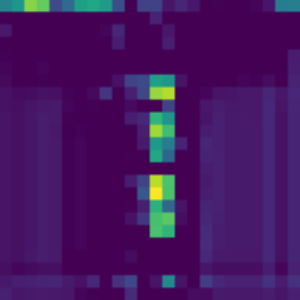} & 
  \includegraphics[width=2cm]{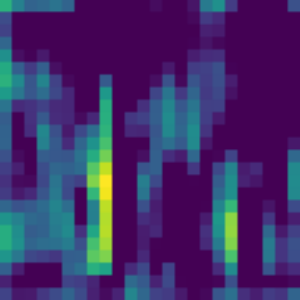} & 
  \includegraphics[width=2cm]{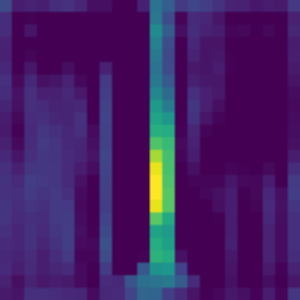} 
    \\

\end{tabular}
    \caption{\textbf{Visualization of Model Input and Features.} Analysis of a scenario where the agent is following the leading vehicle on wet noon.}
    \label{fig:feature_b}
\end{figure*}

\begin{figure*}
    \centering
    \setlength{\tabcolsep}{2pt}
    \begin{tabular}{ccccc}
    \multicolumn{4}{c}{\textbf{Image}}&
    \multicolumn{1}{c}{\textbf{BEV}}
    
    \\
    \multicolumn{4}{c}{\includegraphics[height=2cm]{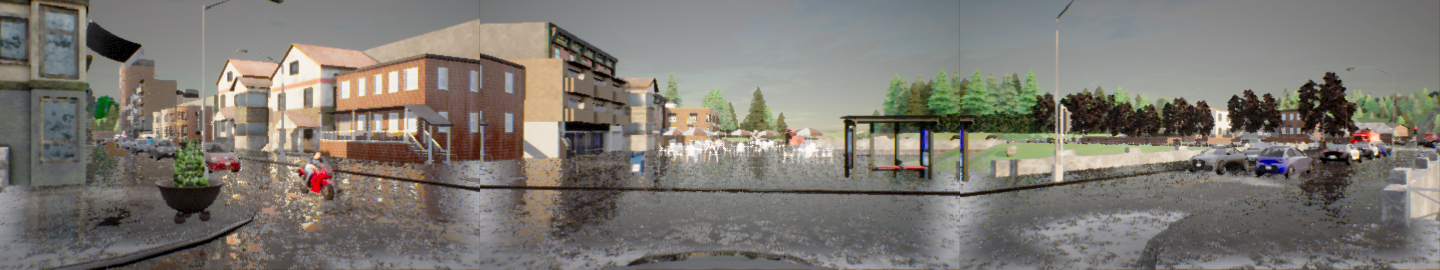}}&
    \multicolumn{1}{c}{\includegraphics[height=2cm]{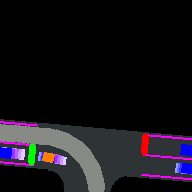}}
 
    \\
      \textbf{No Alignment Module} & & \textbf{Privileged} & \textbf{Output Distillation} & \textbf{CaT} 
    \\  
  \includegraphics[width=5.3cm,height=2cm]{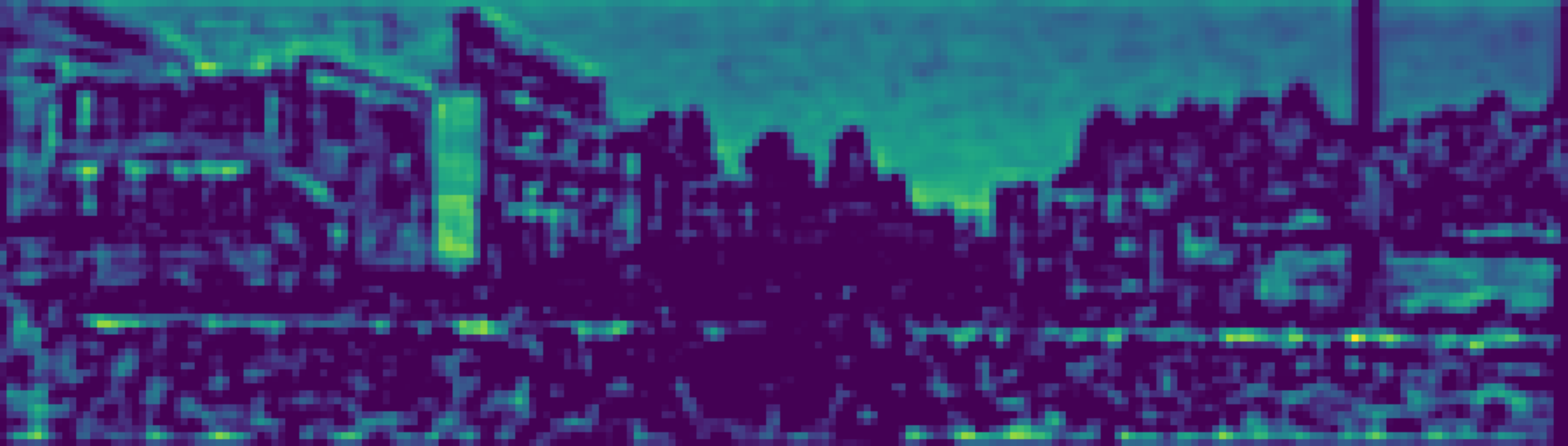} && \includegraphics[width=2cm]{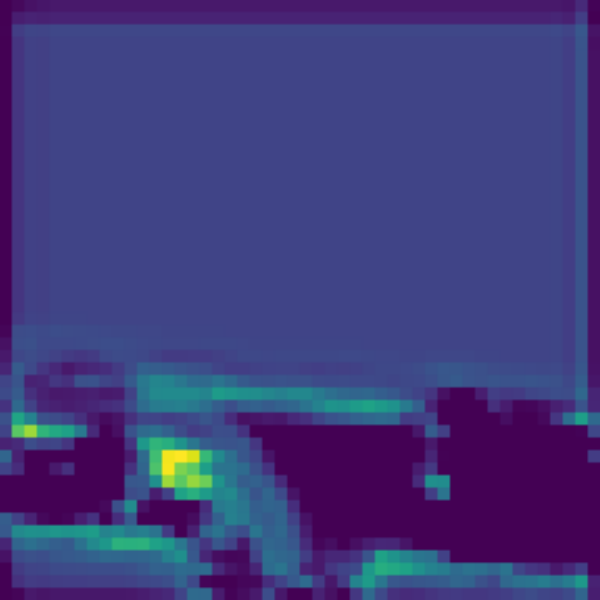} & 
  \includegraphics[width=2cm]{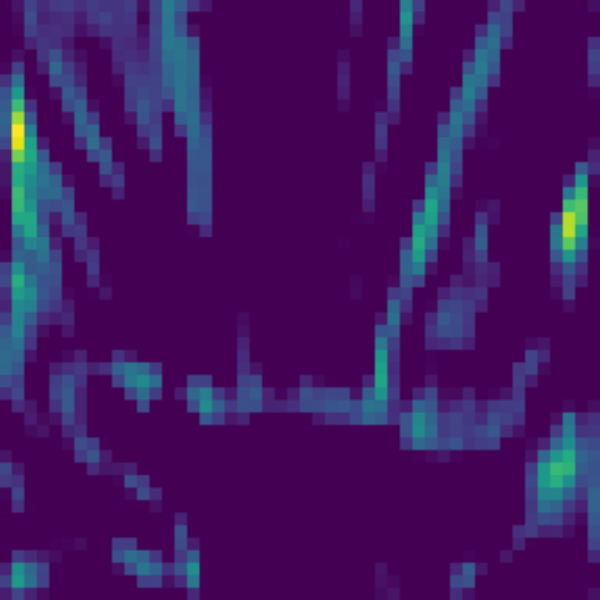} & 
  \includegraphics[width=2cm]{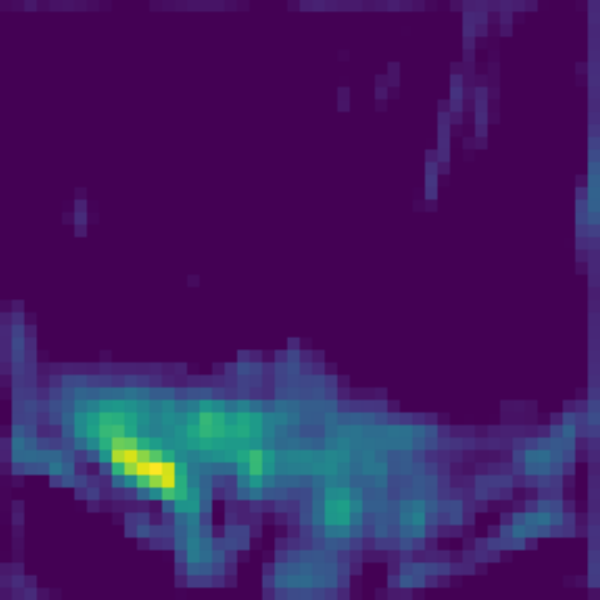} 
    \\

  \includegraphics[width=5.3cm,height=2cm]{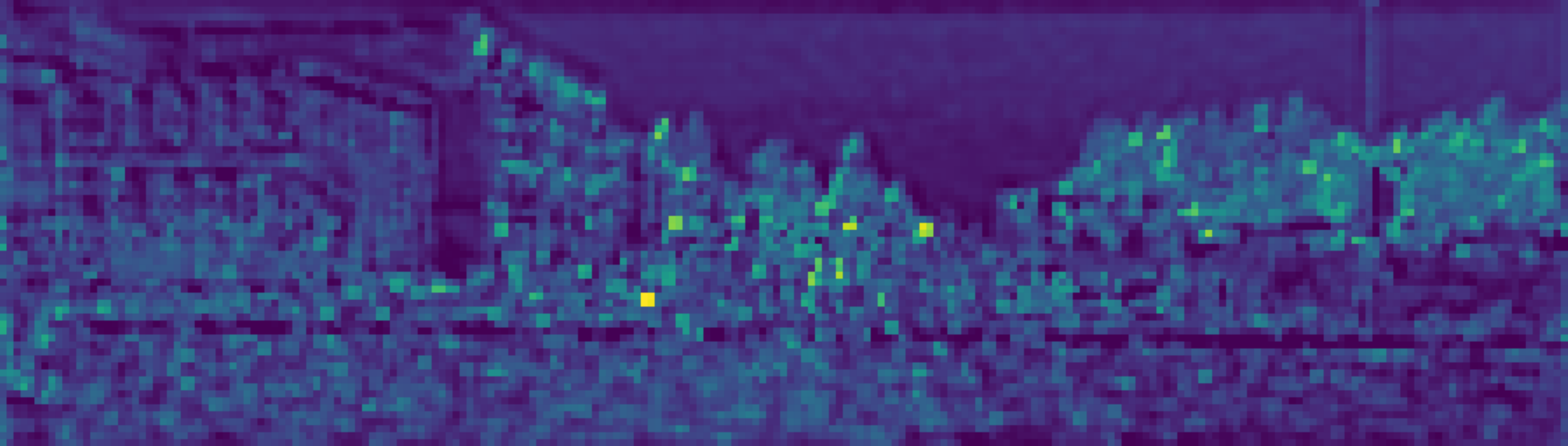} && \includegraphics[width=2cm]{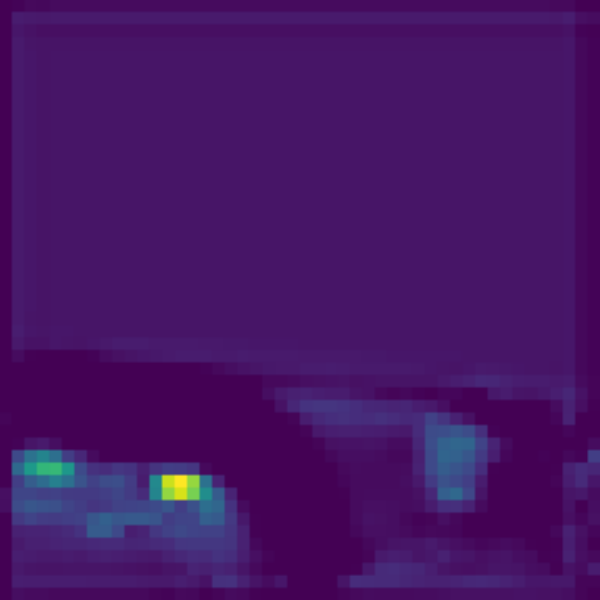} & 
  \includegraphics[width=2cm]{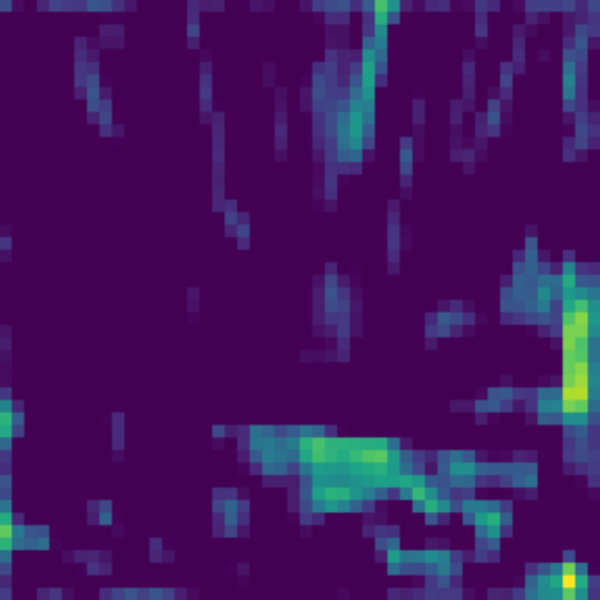} & 
  \includegraphics[width=2cm]{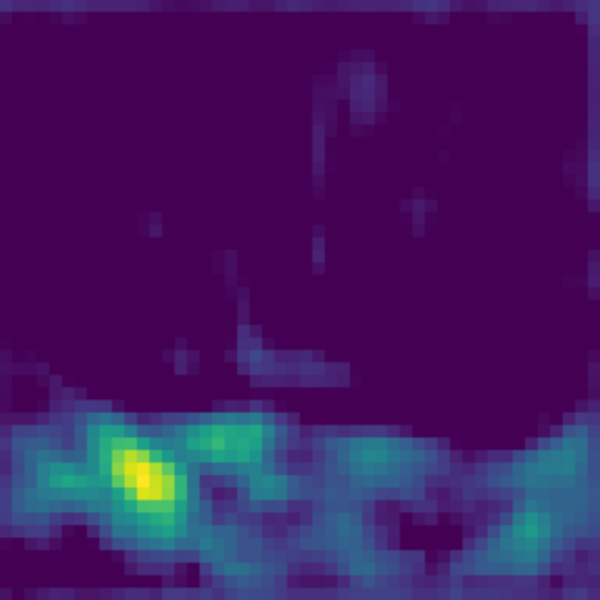} 
    \\
    
    \midrule
    
    \includegraphics[width=5.3cm,height=2cm]{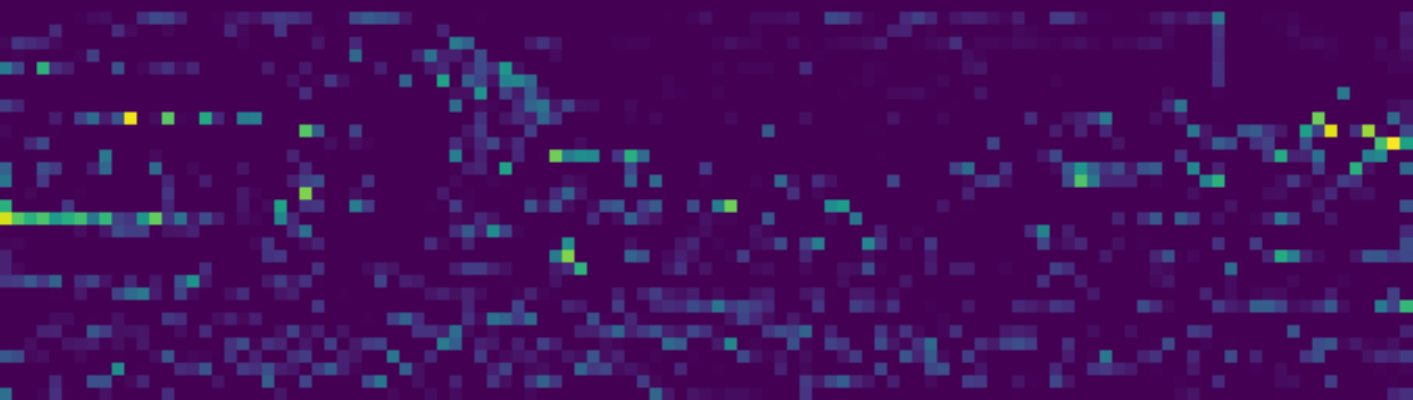} && \includegraphics[width=2cm]{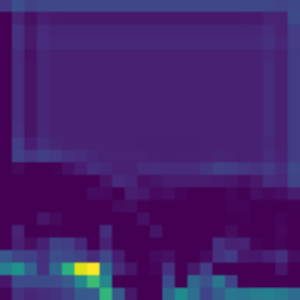} & 
  \includegraphics[width=2cm]{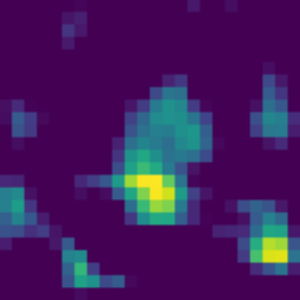} & 
  \includegraphics[width=2cm]{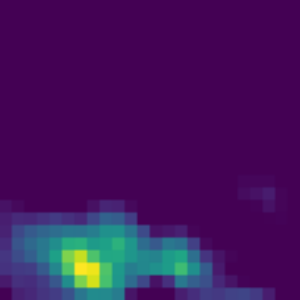} 
    \\

  \includegraphics[width=5.3cm,height=2cm]{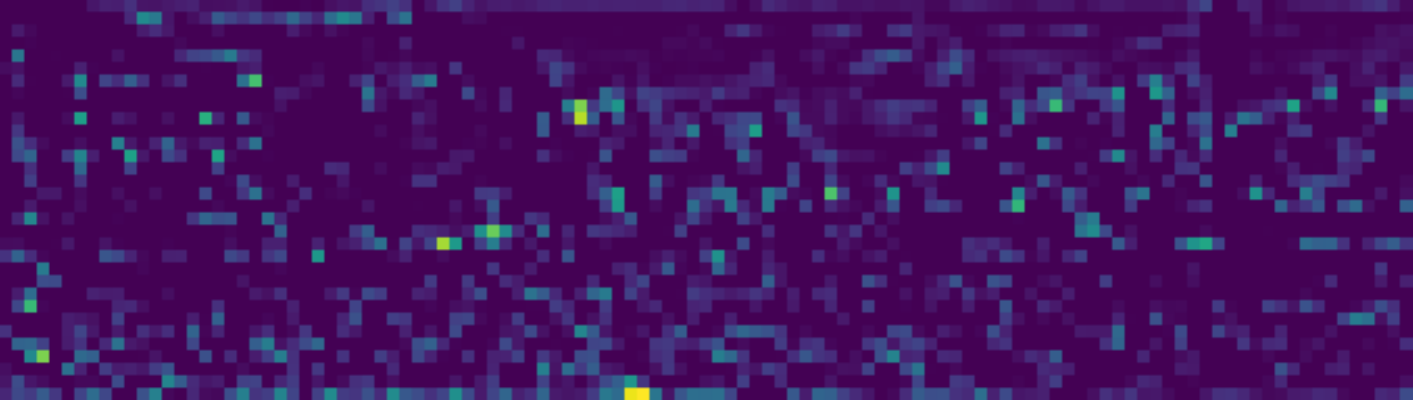} && \includegraphics[width=2cm]{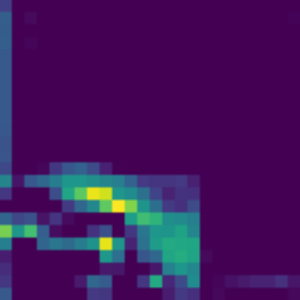} & 
  \includegraphics[width=2cm]{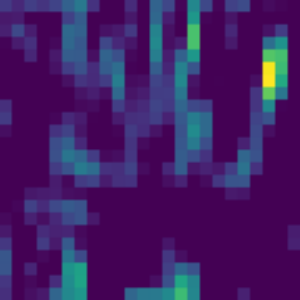} & 
  \includegraphics[width=2cm]{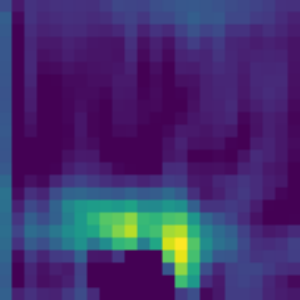} 
    \\

\end{tabular}
    \caption{\textbf{Visualization of Model Input and Features.} Analysis of a scenario where the agent is turning left at an intersection with a motorcycle coming from the left in hard-rain conditions.}
    \label{fig:feature_c}
\end{figure*}

\begin{figure*}
    \centering
    \setlength{\tabcolsep}{2pt}
    \begin{tabular}{ccccc}
    \multicolumn{4}{c}{\textbf{Image}}&
    \multicolumn{1}{c}{\textbf{BEV}}

\\
    \multicolumn{4}{c}{\includegraphics[height=2cm]{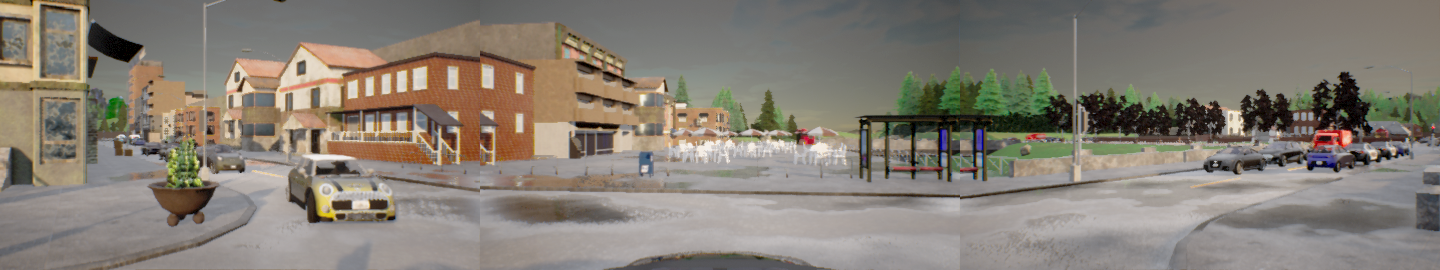}}&
    \multicolumn{1}{c}{\includegraphics[height=2cm]{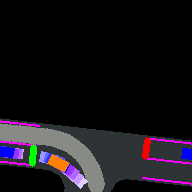}}
 
    \\
      \textbf{No Alignment Module} & & \textbf{Privileged} & \textbf{Output Distillation} & \textbf{CaT} 
    \\  
  \includegraphics[width=5.3cm,height=2cm]{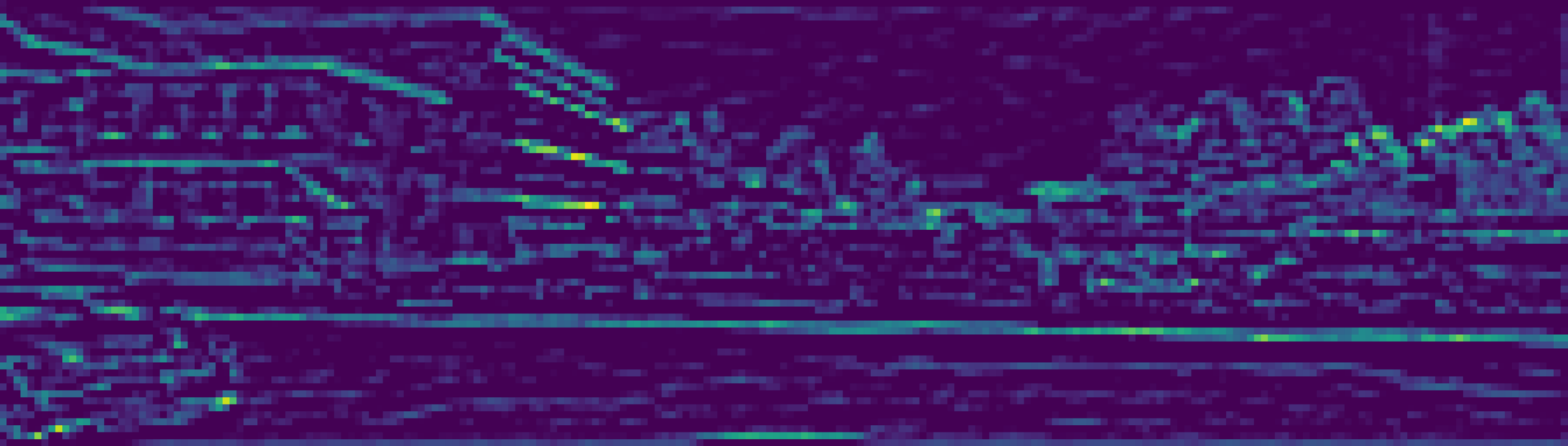} && \includegraphics[width=2cm]{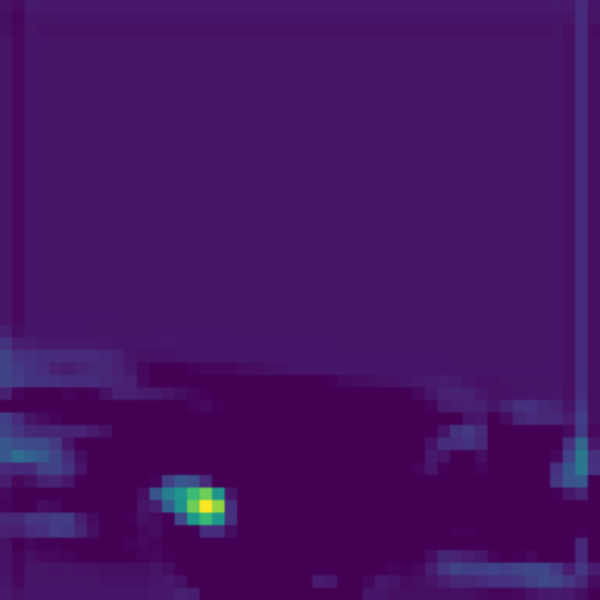} & 
  \includegraphics[width=2cm]{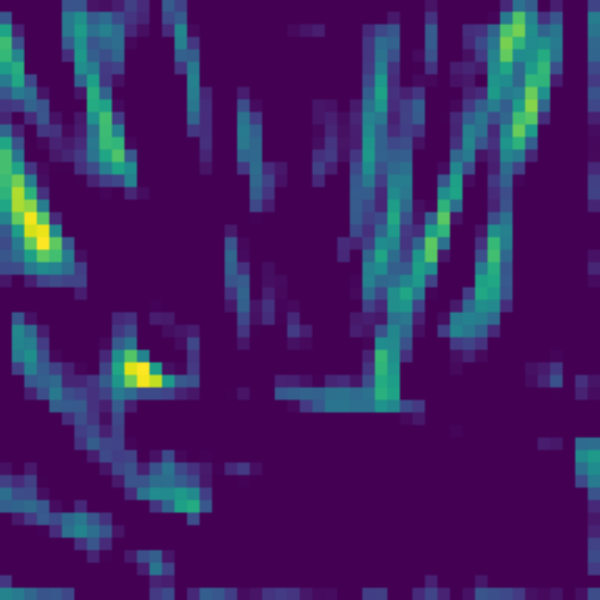} & 
  \includegraphics[width=2cm]{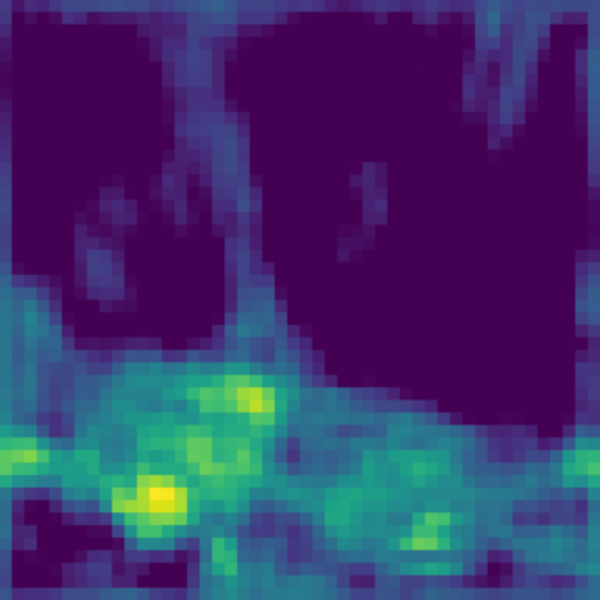} 
    \\

  \includegraphics[width=5.3cm,height=2cm]{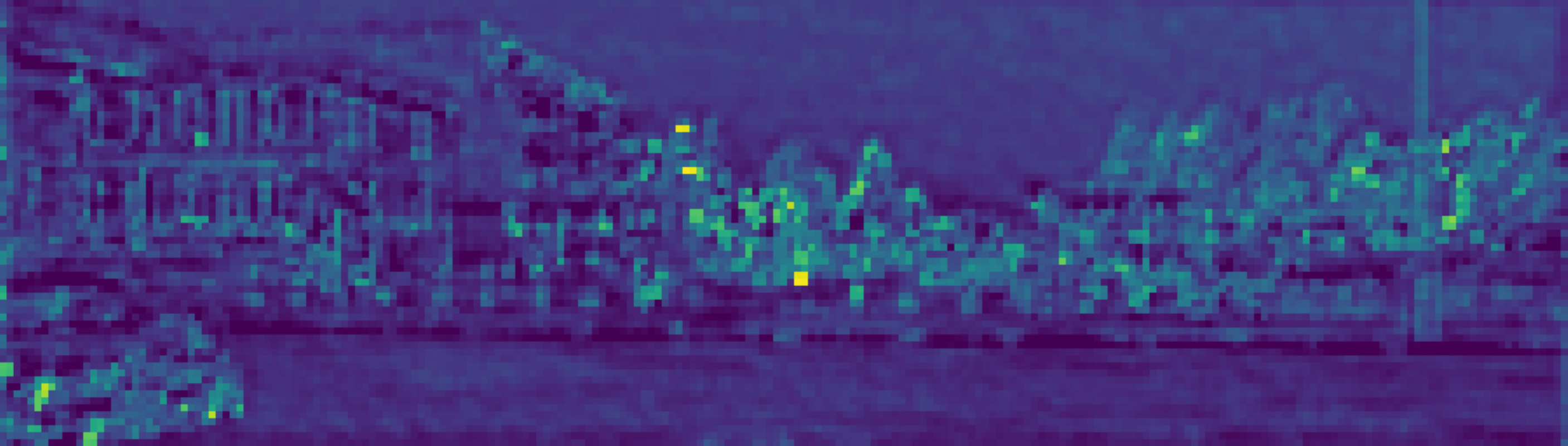} && \includegraphics[width=2cm]{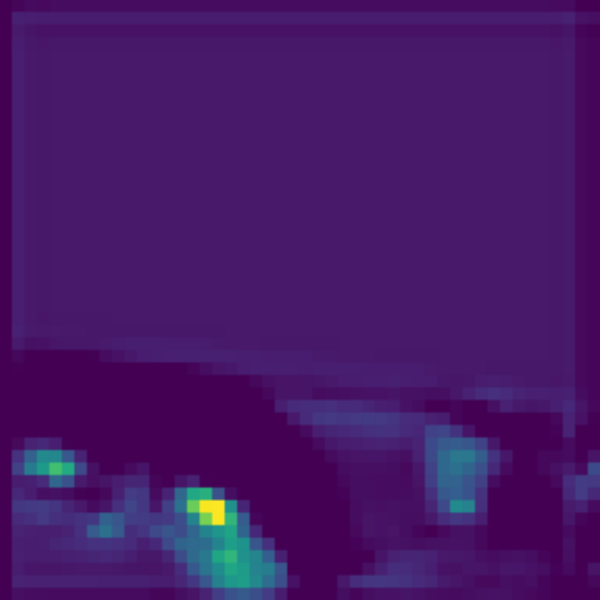} & 
  \includegraphics[width=2cm]{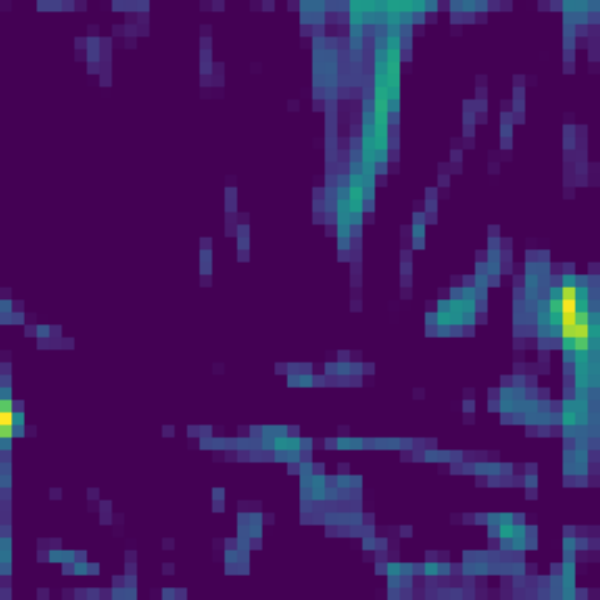} & 
  \includegraphics[width=2cm]{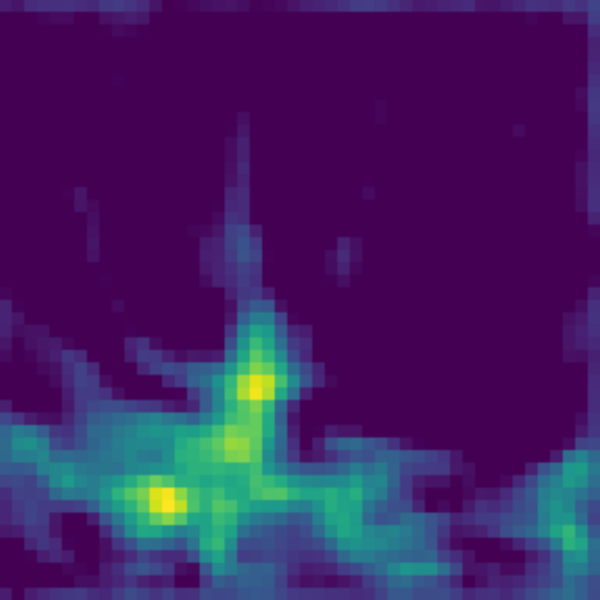} 
    \\
    
    \midrule
    
    \includegraphics[width=5.3cm,height=2cm]{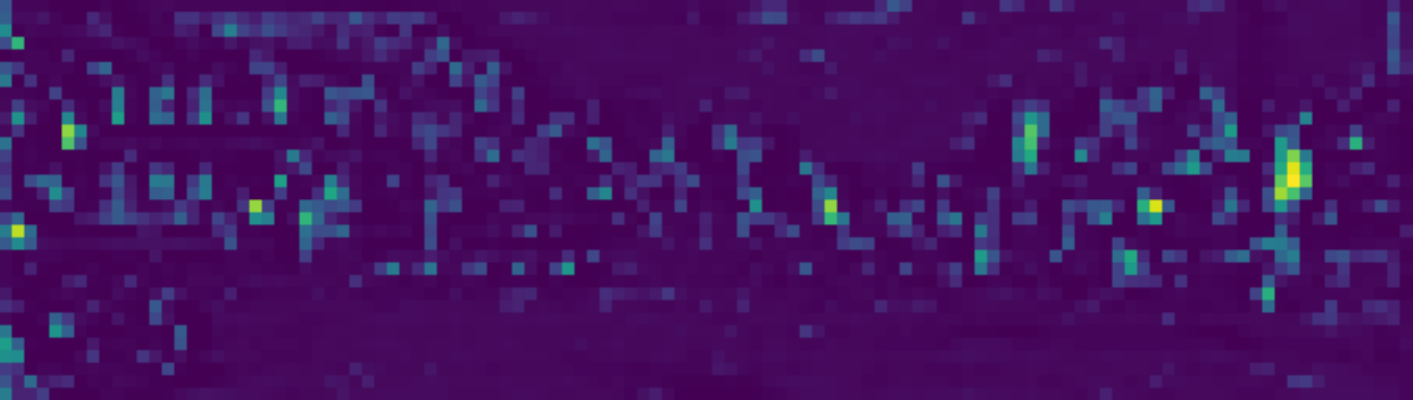} && \includegraphics[width=2cm]{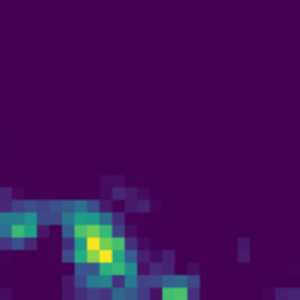} & 
  \includegraphics[width=2cm]{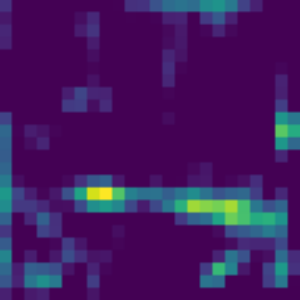} & 
  \includegraphics[width=2cm]{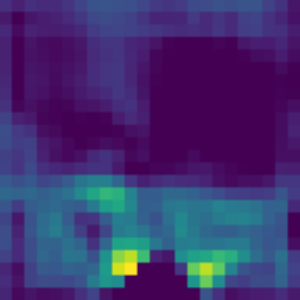} 
    \\

  \includegraphics[width=5.3cm,height=2cm]{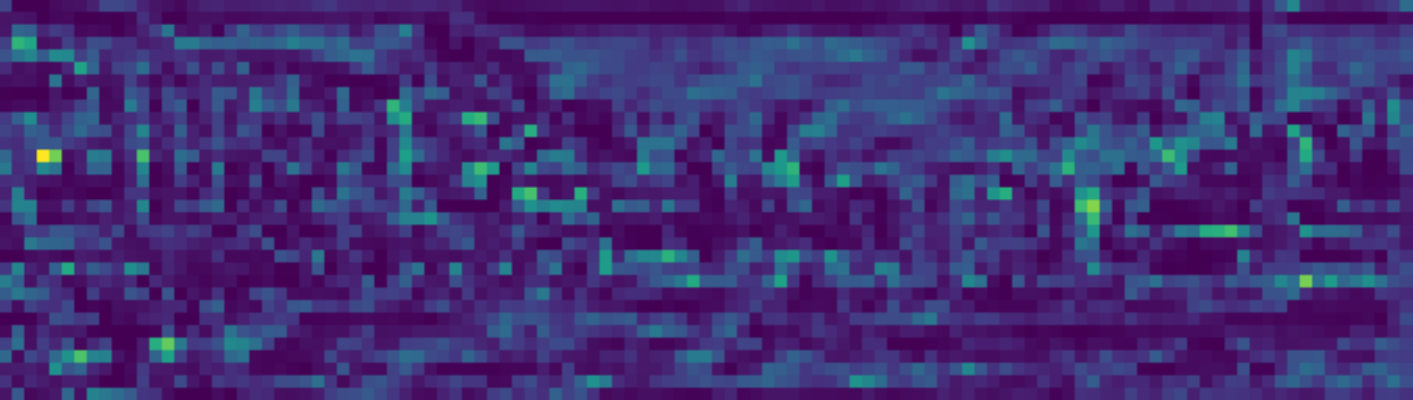} && \includegraphics[width=2cm]{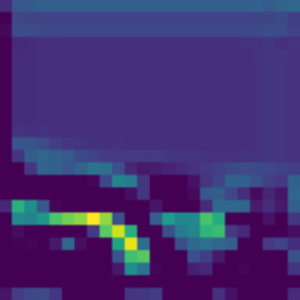} & 
  \includegraphics[width=2cm]{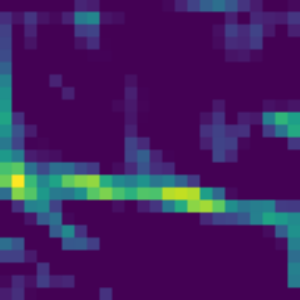} & 
  \includegraphics[width=2cm]{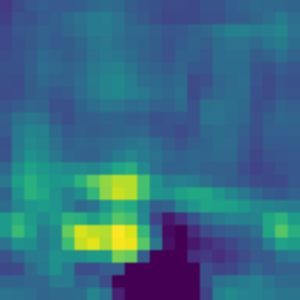} 
    \\

\end{tabular}
    \caption{\textbf{Visualization of Model Input and Features.} Analysis of a scenario where the agent is turning left at an intersection with traffic coming from the left in wet sunset conditions.}
    \label{fig:feature_d}
\end{figure*}

\begin{figure*}
    \centering
    \setlength{\tabcolsep}{2pt}
    \begin{tabular}{ccccc}
    \multicolumn{4}{c}{\textbf{Image}}&
    \multicolumn{1}{c}{\textbf{BEV}}
    
    \\
    \multicolumn{4}{c}{\includegraphics[height=2cm]{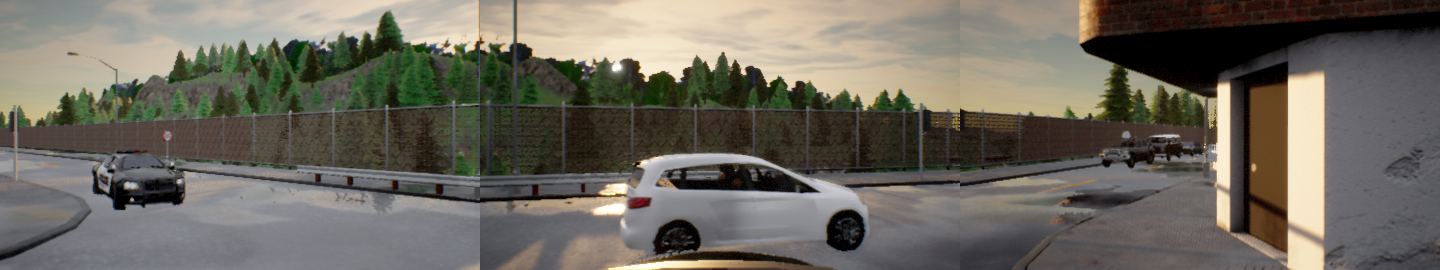}}&
    \multicolumn{1}{c}{\includegraphics[height=2cm]{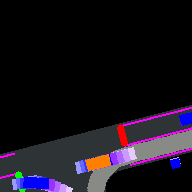}}
 
    \\
      \textbf{No Alignment Module} & & \textbf{Privileged} & \textbf{Output Distillation} & \textbf{CaT} 
    \\  
  \includegraphics[width=5.3cm,height=2cm]{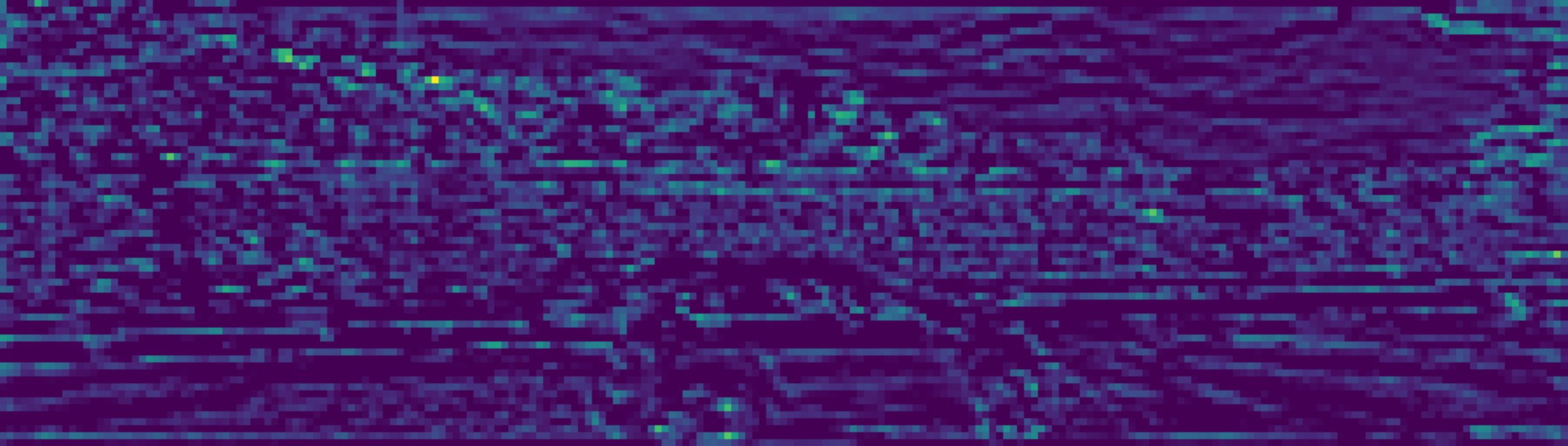} && \includegraphics[width=2cm]{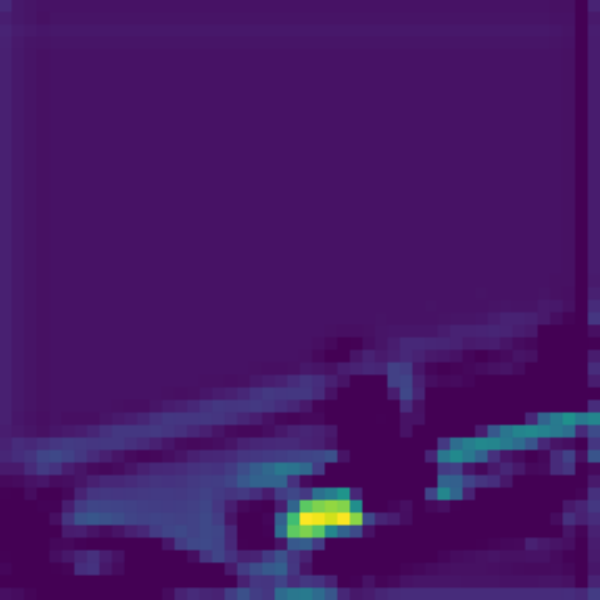} & 
  \includegraphics[width=2cm]{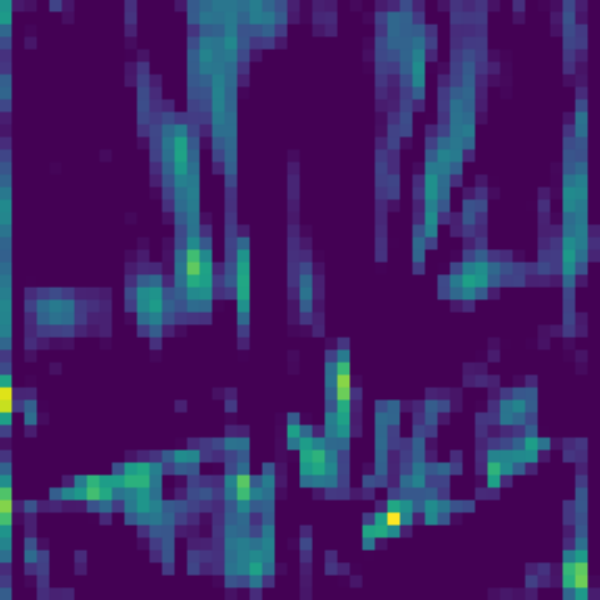} & 
  \includegraphics[width=2cm]{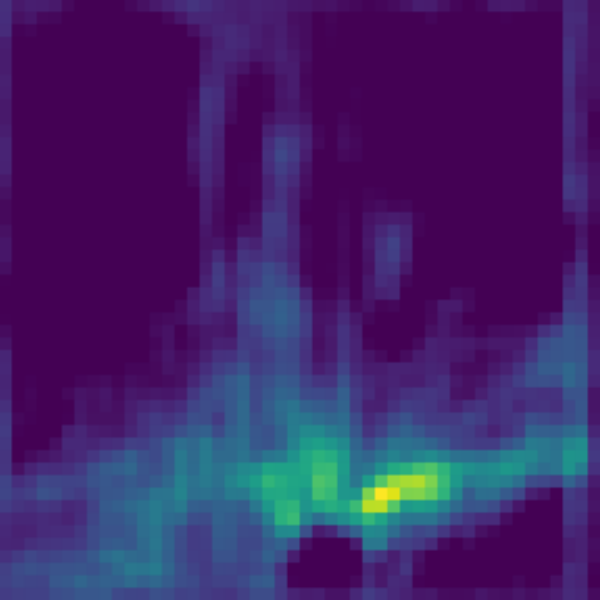} 
    \\

  \includegraphics[width=5.3cm,height=2cm]{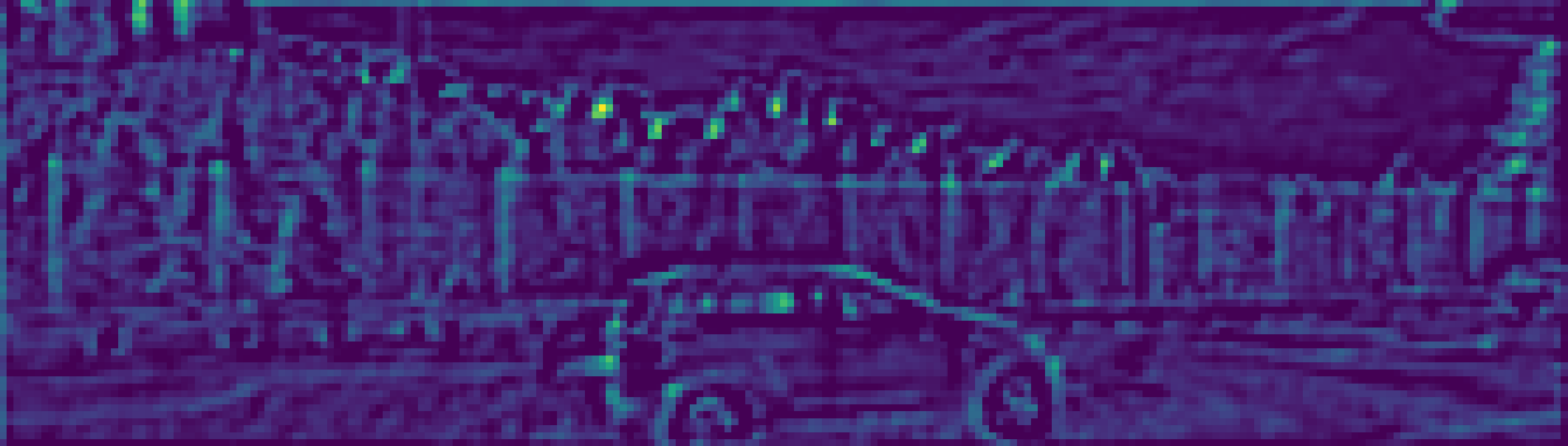} && \includegraphics[width=2cm]{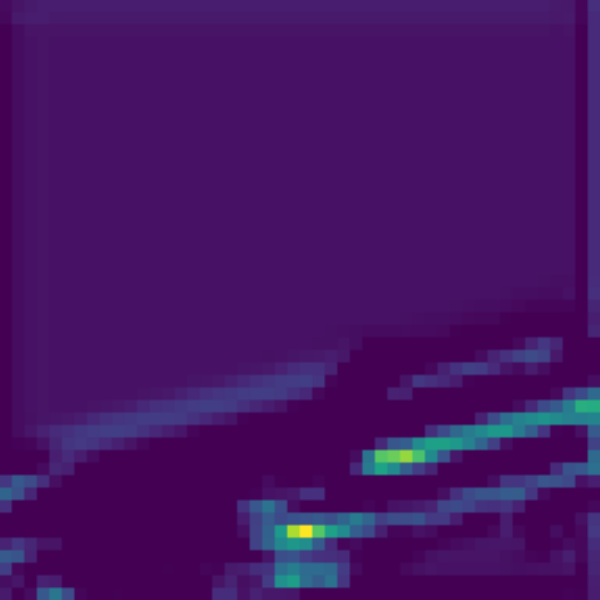} & 
  \includegraphics[width=2cm]{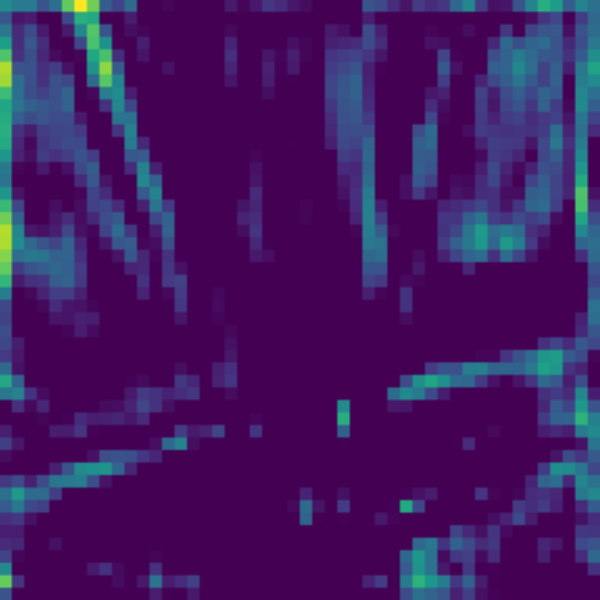} & 
  \includegraphics[width=2cm]{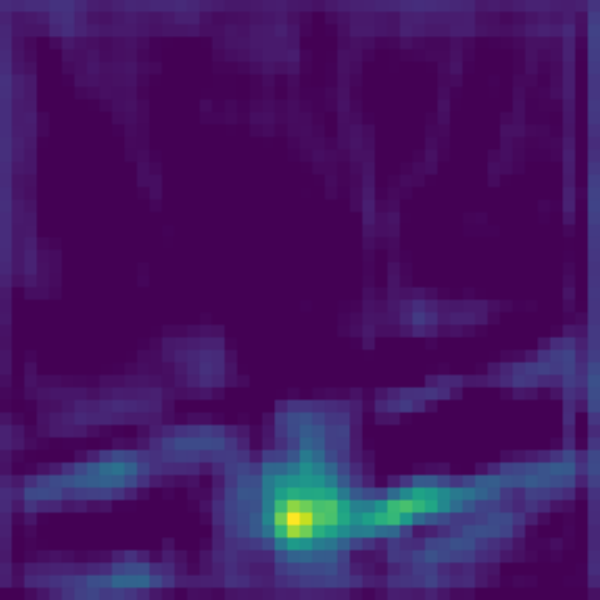} 
    \\
    
    \midrule
    
    \includegraphics[width=5.3cm,height=2cm]{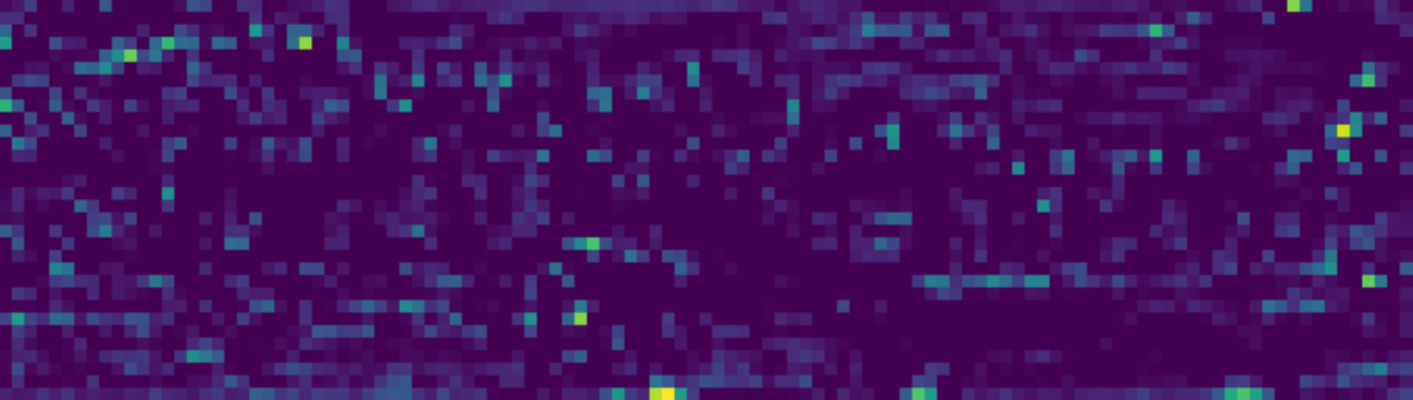} && \includegraphics[width=2cm]{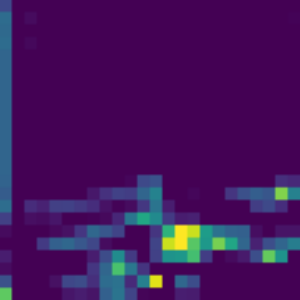} & 
  \includegraphics[width=2cm]{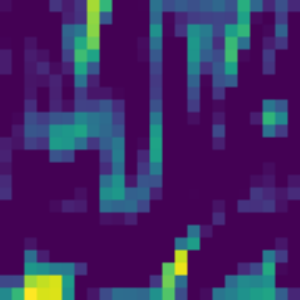} & 
  \includegraphics[width=2cm]{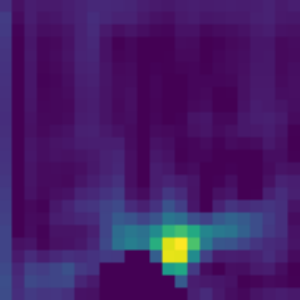} 
    \\

  \includegraphics[width=5.3cm,height=2cm]{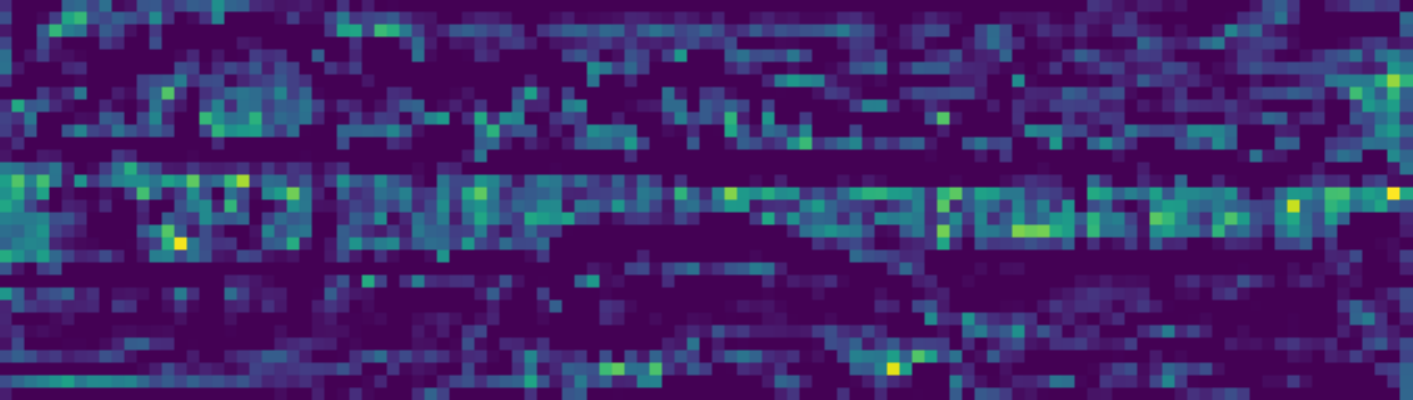} && \includegraphics[width=2cm]{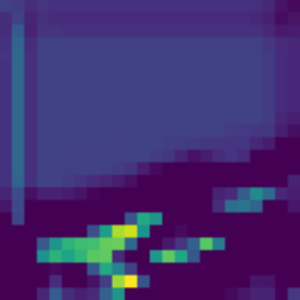} & 
  \includegraphics[width=2cm]{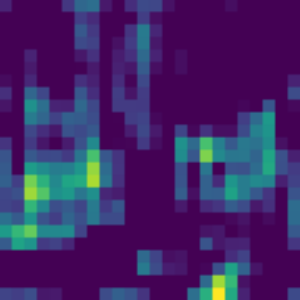} & 
  \includegraphics[width=2cm]{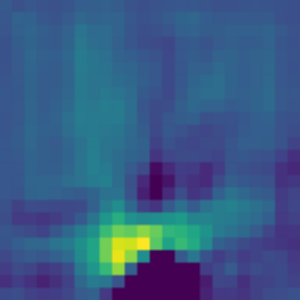} 
    \\

\end{tabular}
    \caption{\textbf{Visualization of Model Input and Features.} Analysis of a scenario where the agent is turning right at an intersection yielding to traffic in wet sunset conditions.}
    \label{fig:feature_e}
\end{figure*}

\begin{figure*}
    \centering
    \setlength{\tabcolsep}{2pt}
    \begin{tabular}{ccccc}
    \multicolumn{4}{c}{\textbf{Image}}&
    \multicolumn{1}{c}{\textbf{BEV}}

    \\
    \multicolumn{4}{c}{\includegraphics[height=2cm]{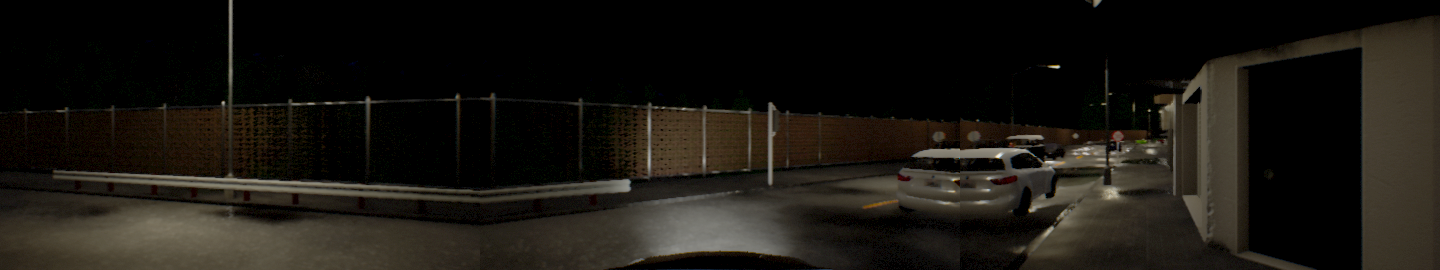}}&
    \multicolumn{1}{c}{\includegraphics[height=2cm]{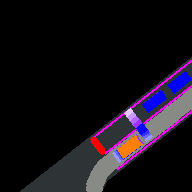}}
 
    \\
      \textbf{No Alignment Module} & & \textbf{Privileged} & \textbf{Output Distillation} & \textbf{CaT} 
    \\  
  \includegraphics[width=5.3cm,height=2cm]{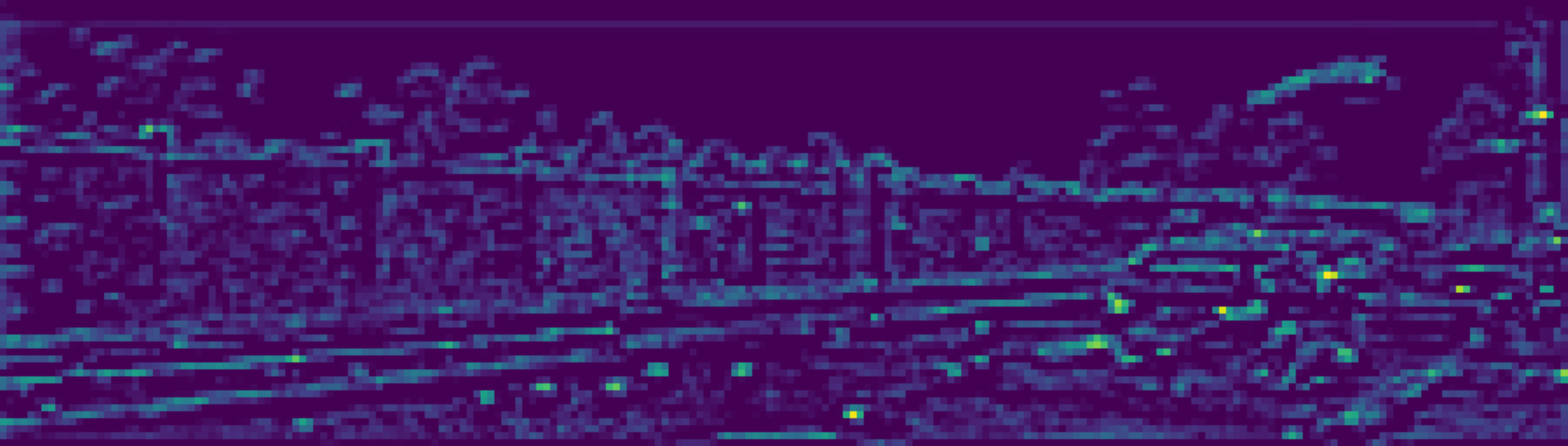} && \includegraphics[width=2cm]{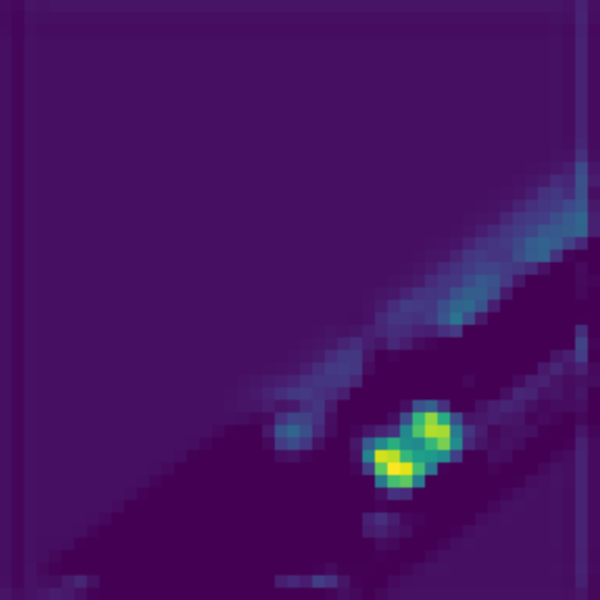} & 
  \includegraphics[width=2cm]{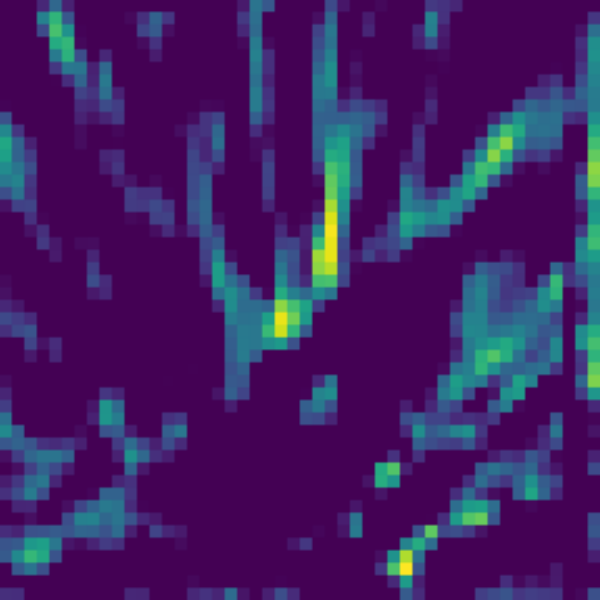} & 
  \includegraphics[width=2cm]{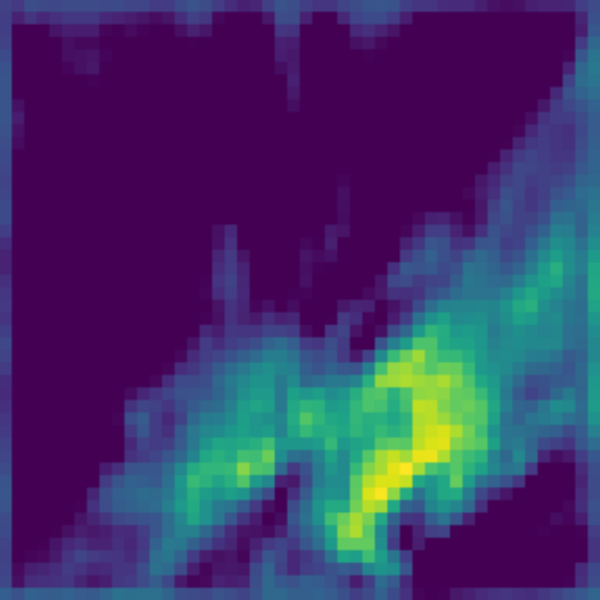} 
    \\

  \includegraphics[width=5.3cm,height=2cm]{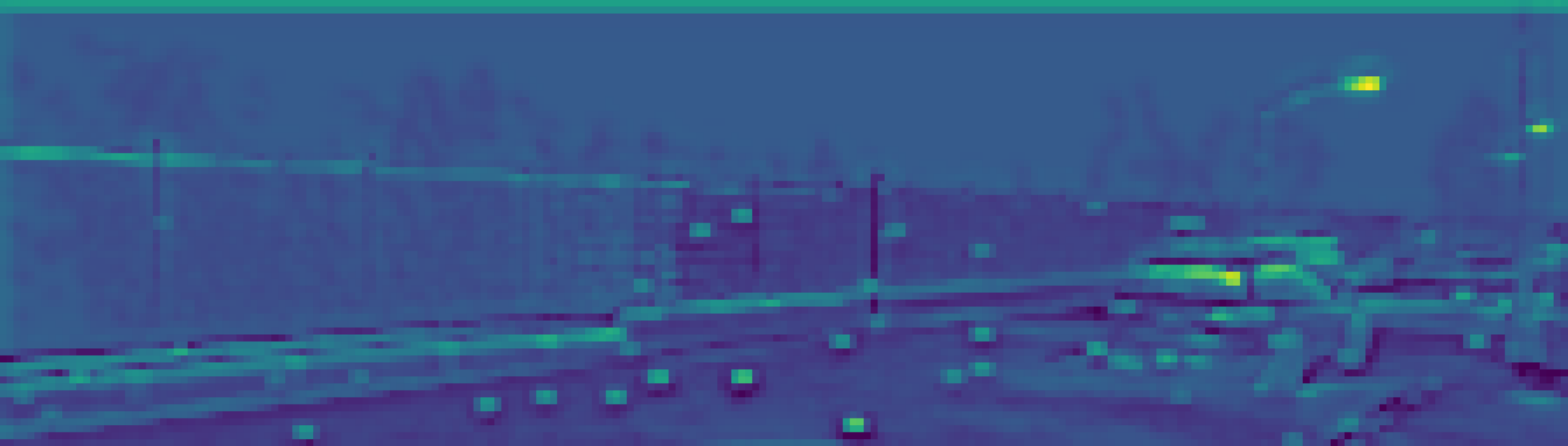} && \includegraphics[width=2cm]{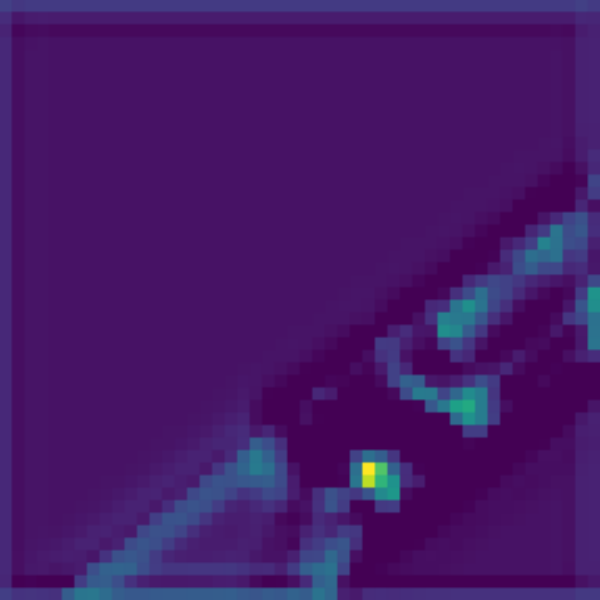} & 
  \includegraphics[width=2cm]{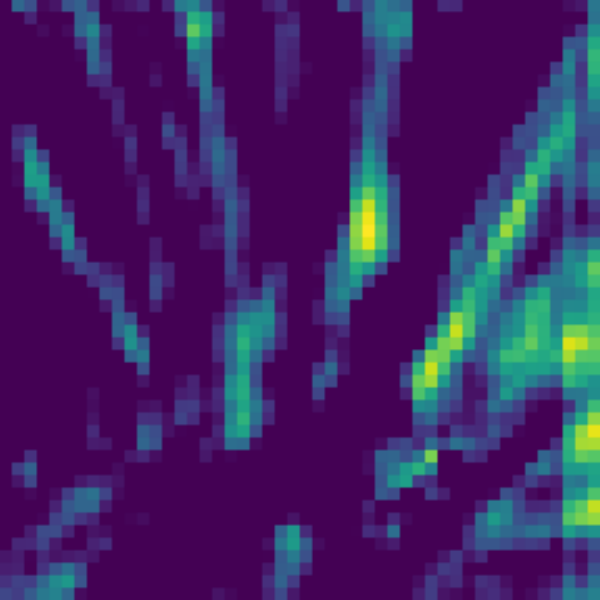} & 
  \includegraphics[width=2cm]{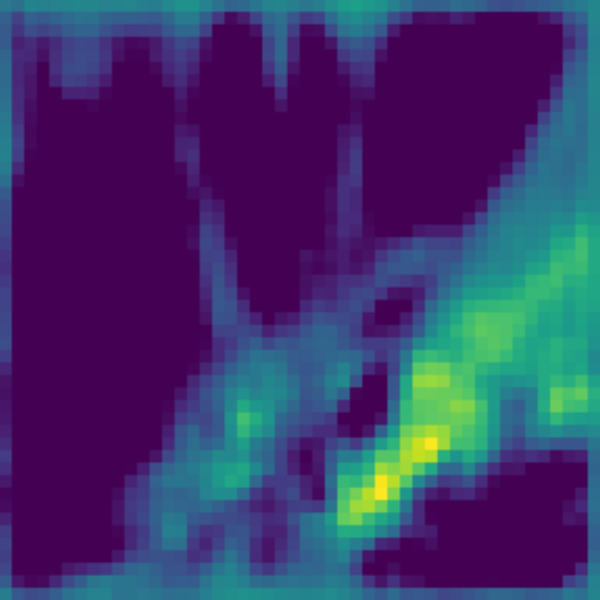} 
    \\

    \midrule
    
    \includegraphics[width=5.3cm,height=2cm]{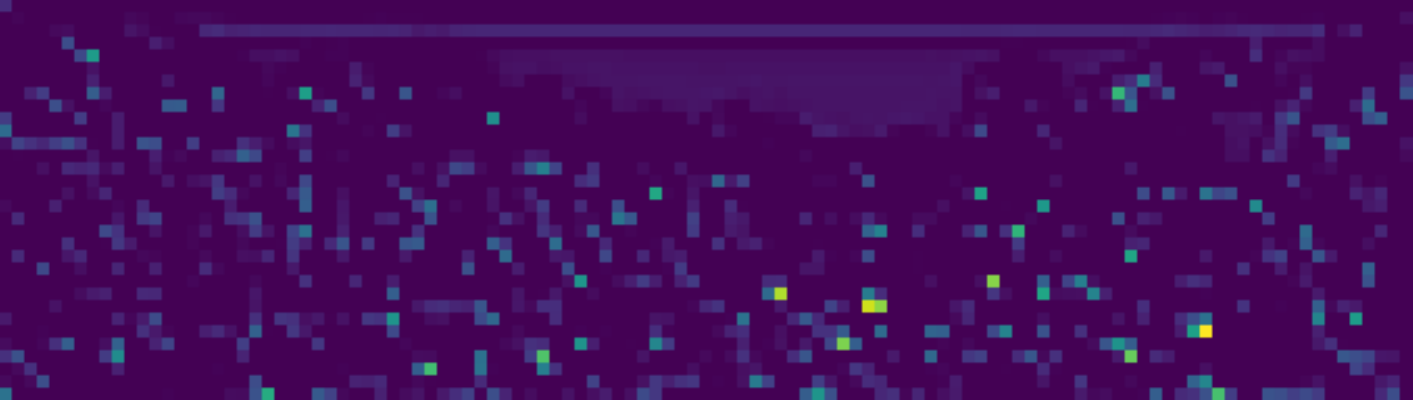} && \includegraphics[width=2cm]{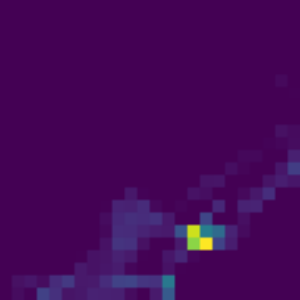} & 
  \includegraphics[width=2cm]{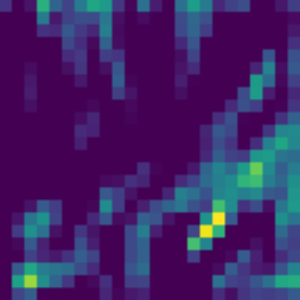} & 
  \includegraphics[width=2cm]{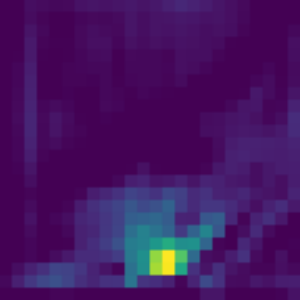} 
    \\

  \includegraphics[width=5.3cm,height=2cm]{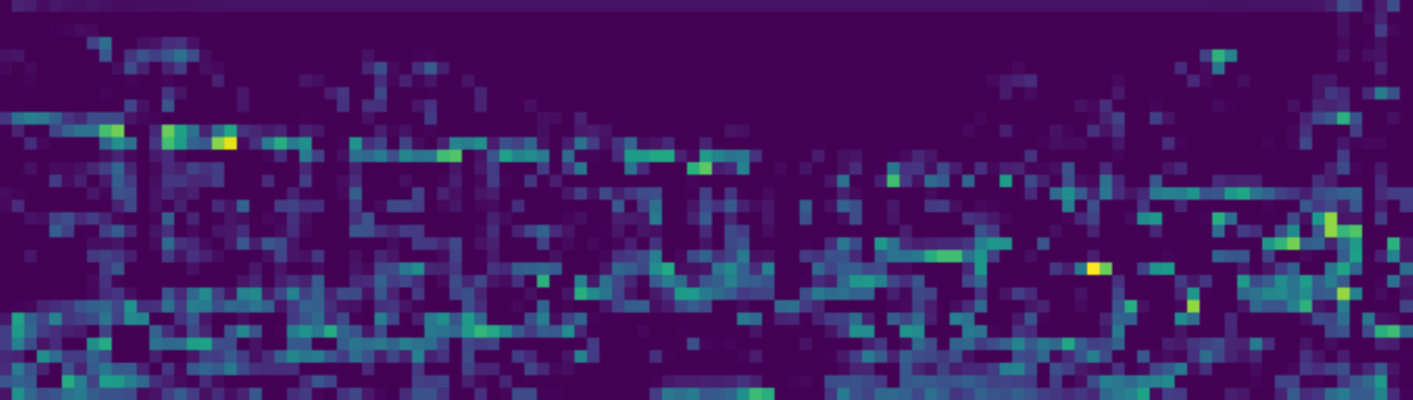} && \includegraphics[width=2cm]{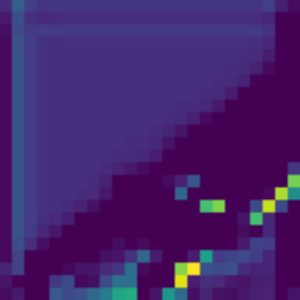} & 
  \includegraphics[width=2cm]{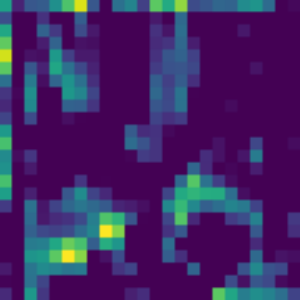} & 
  \includegraphics[width=2cm]{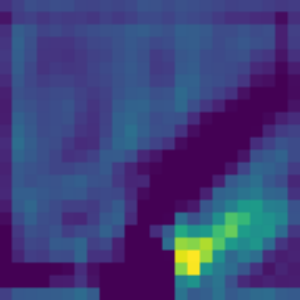} 
    \\

\end{tabular}
    \caption{\textbf{Visualization of Model Input and Features.} Analysis of a scenario where the agent is turning left at an intersection at night with low visibility.}
    \label{fig:feature_f}
\end{figure*}

\begin{figure*}
    \centering
    \setlength{\tabcolsep}{2pt}
    \begin{tabular}{ccccc}
    \multicolumn{4}{c}{\textbf{Image}}&
    \multicolumn{1}{c}{\textbf{BEV}}

    \\
    \multicolumn{4}{c}{\includegraphics[height=2cm]{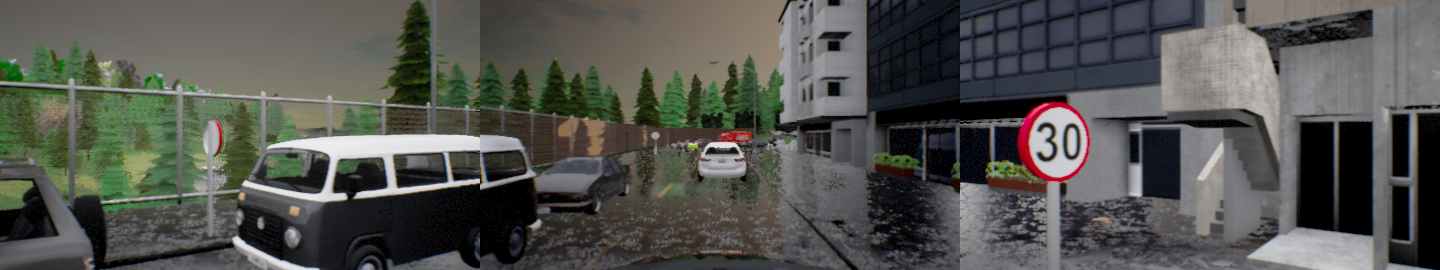}}&
    \multicolumn{1}{c}{\includegraphics[height=2cm]{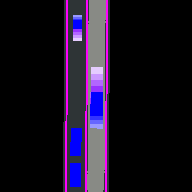}}
 
    \\
      \textbf{No Alignment Module} & & \textbf{Privileged} & \textbf{Output Distillation} & \textbf{CaT} 
    \\  
  \includegraphics[width=5.3cm,height=2cm]{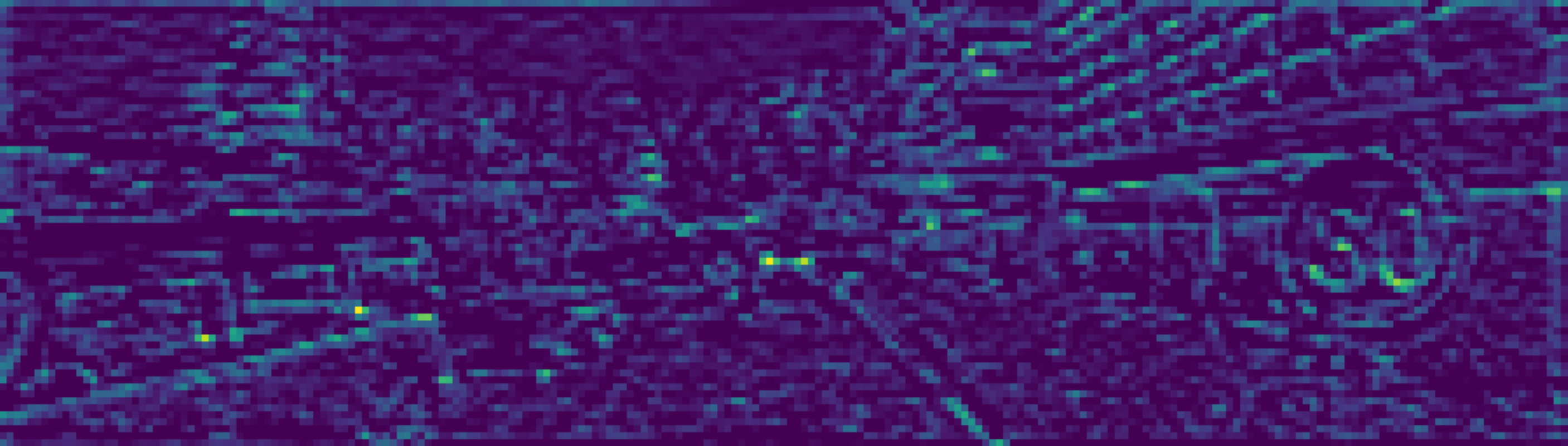} && \includegraphics[width=2cm]{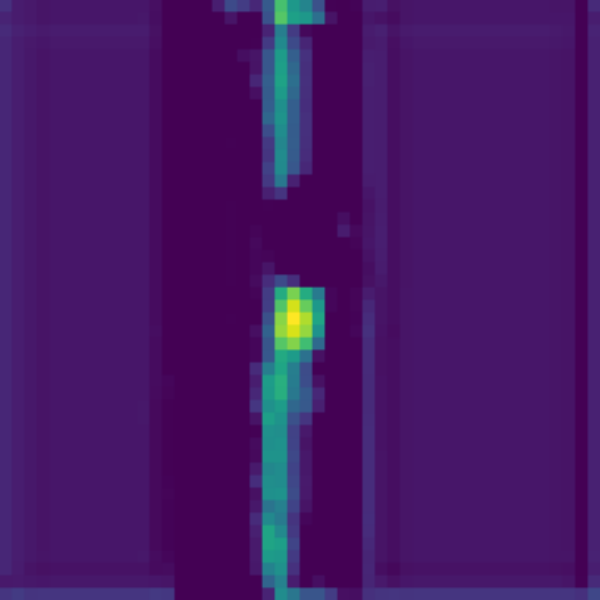} & 
  \includegraphics[width=2cm]{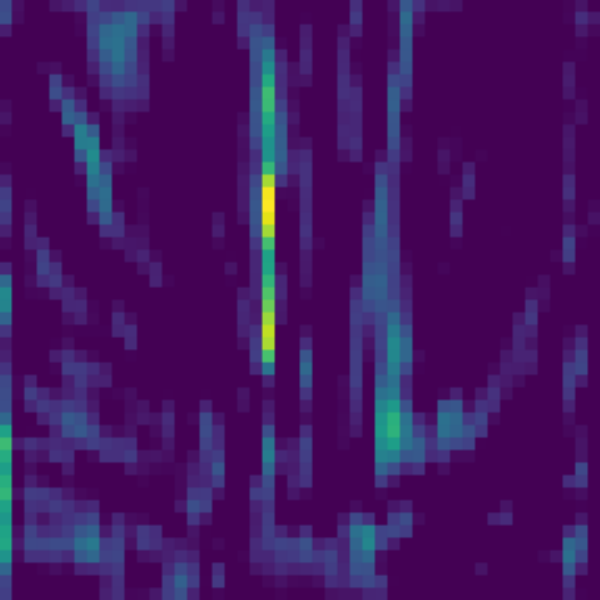} & 
  \includegraphics[width=2cm]{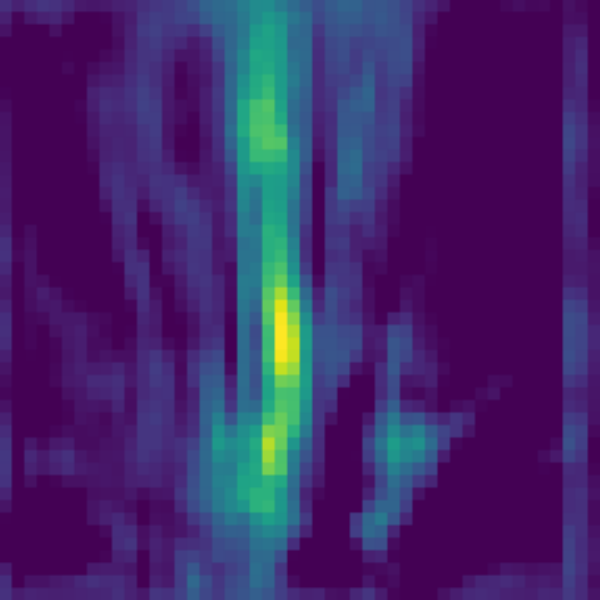} 
    \\

  \includegraphics[width=5.3cm,height=2cm]{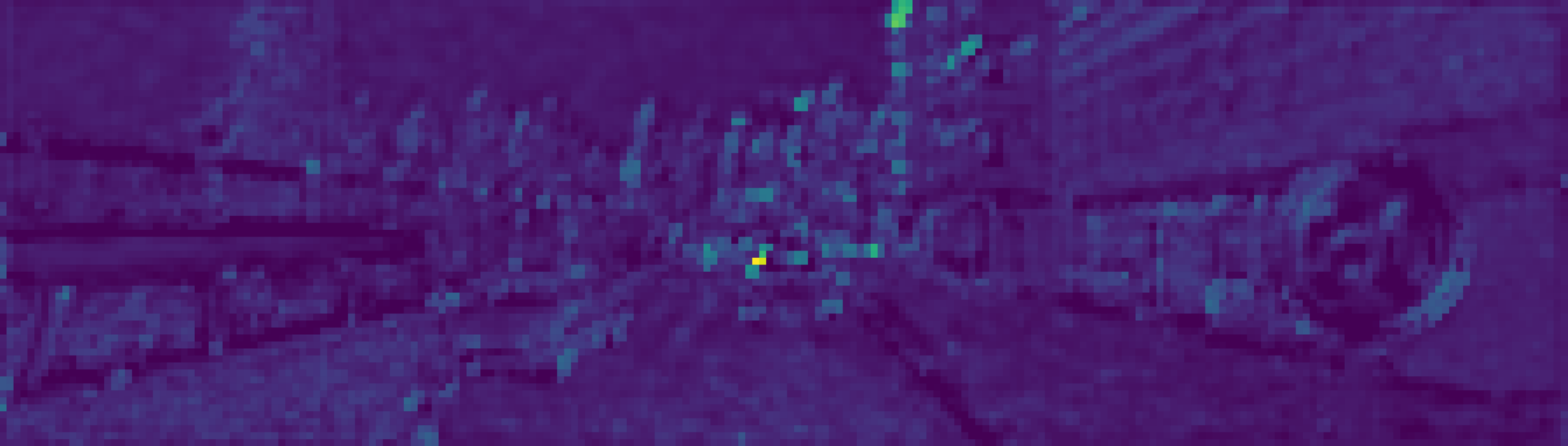} && \includegraphics[width=2cm]{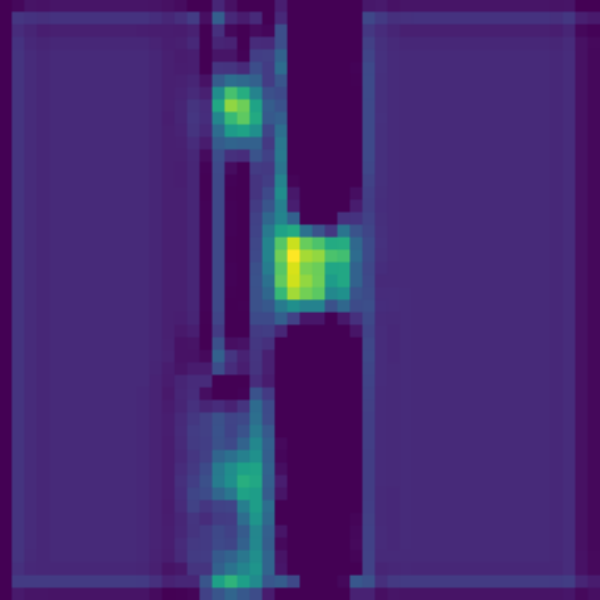} & 
  \includegraphics[width=2cm]{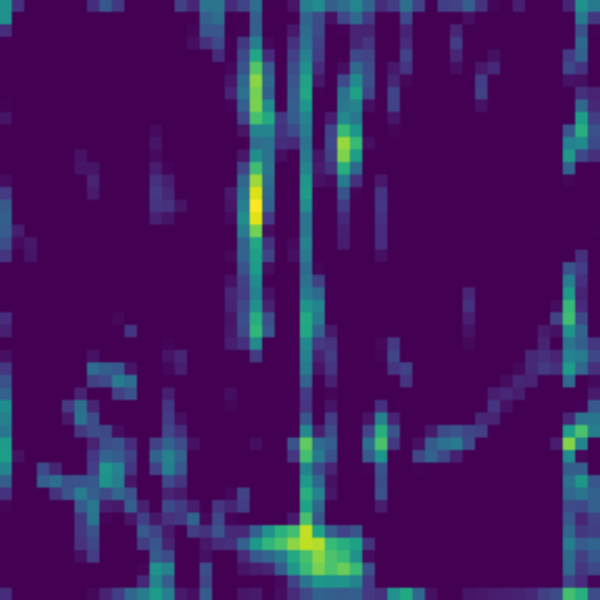} & 
  \includegraphics[width=2cm]{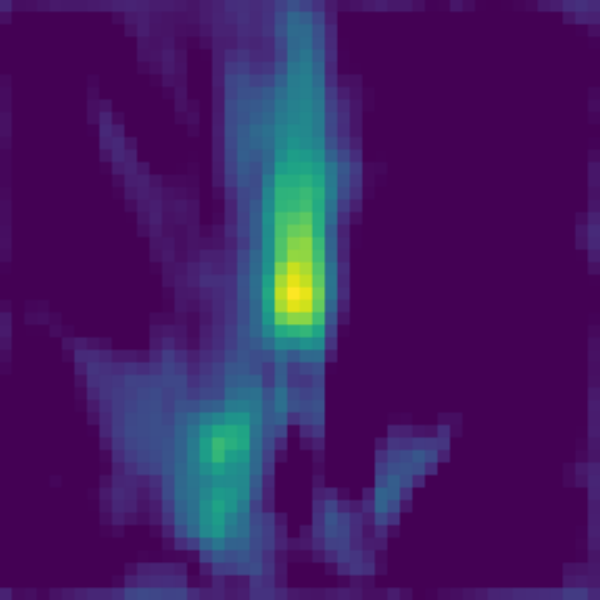} 
    \\
    
    \midrule
    
    \includegraphics[width=5.3cm,height=2cm]{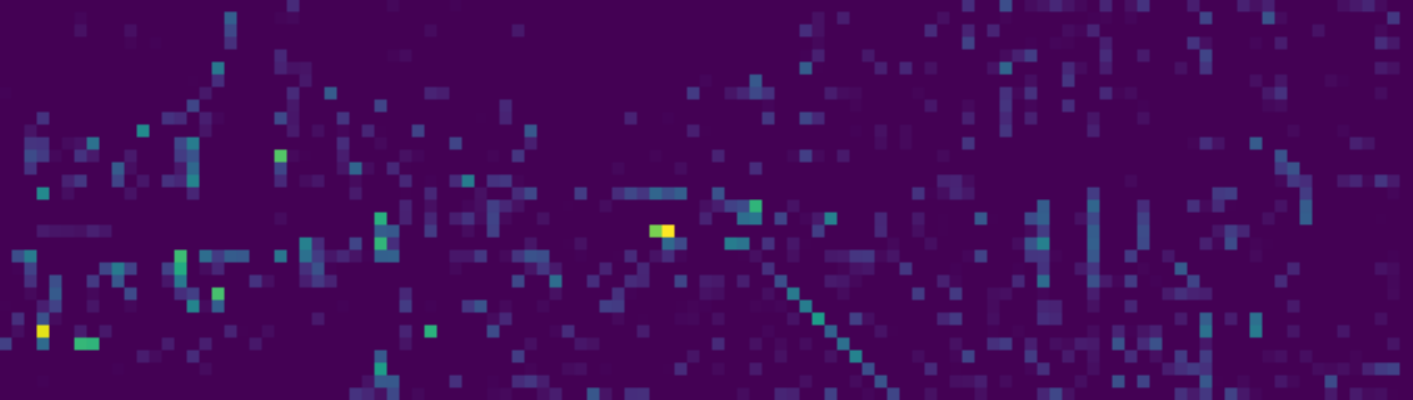} && \includegraphics[width=2cm]{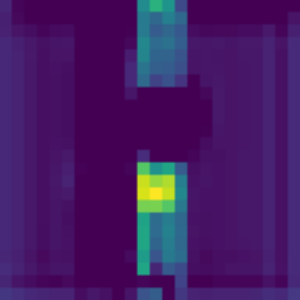} & 
  \includegraphics[width=2cm]{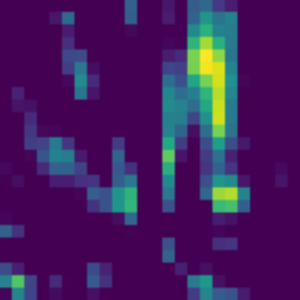} & 
  \includegraphics[width=2cm]{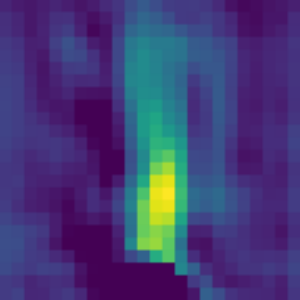} 
    \\

  \includegraphics[width=5.3cm,height=2cm]{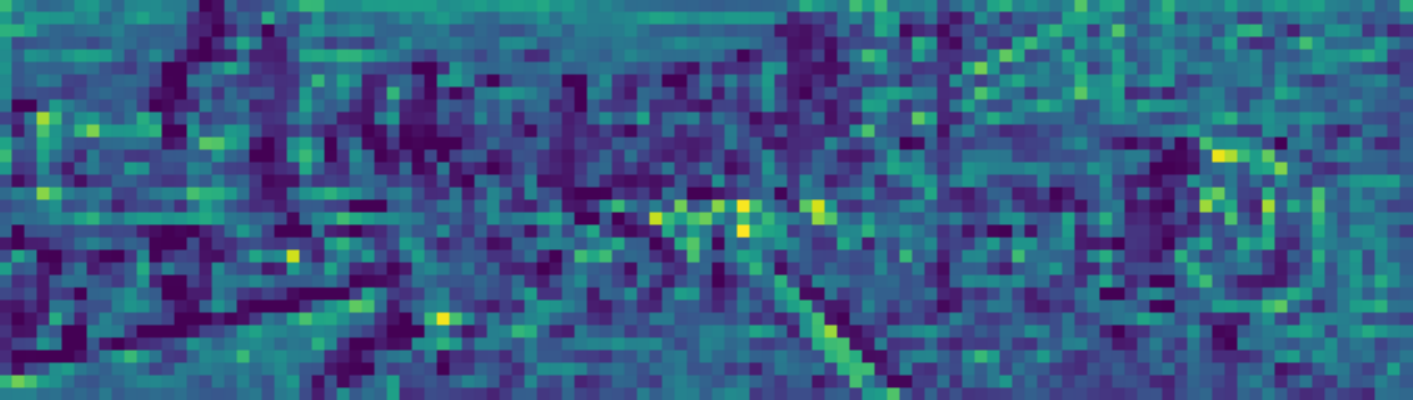} && \includegraphics[width=2cm]{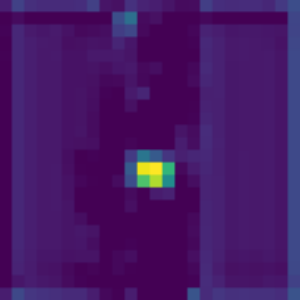} & 
  \includegraphics[width=2cm]{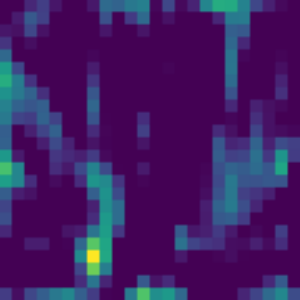} & 
  \includegraphics[width=2cm]{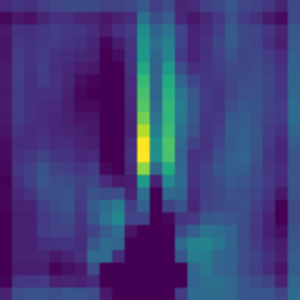} 
    \\

\end{tabular}
    \caption{\textbf{Visualization of Model Input and Features.} Analysis of a scenario where the agent is following the lane in hard-rain conditions.}
    \label{fig:feature_g}
\end{figure*}

\begin{figure*}
    \centering
    \setlength{\tabcolsep}{2pt}
    \begin{tabular}{ccccc}
    \multicolumn{4}{c}{\textbf{Image}}&
    \multicolumn{1}{c}{\textbf{BEV}}

    \\
    \multicolumn{4}{c}{\includegraphics[height=2cm]{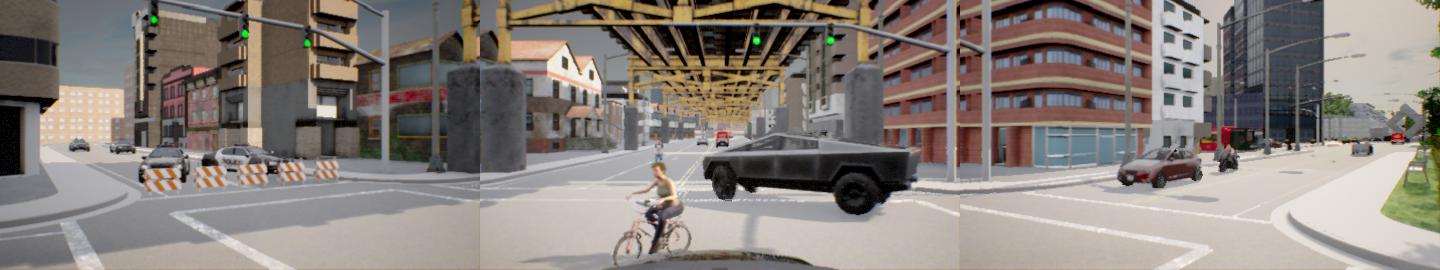}}&
    \multicolumn{1}{c}{\includegraphics[height=2cm]{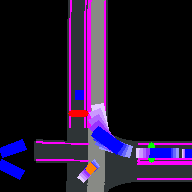}}
 
    \\
      \textbf{No Alignment Module} & & \textbf{Privileged} & \textbf{Output Distillation} & \textbf{CaT} 
    \\  
  \includegraphics[width=5.3cm,height=2cm]{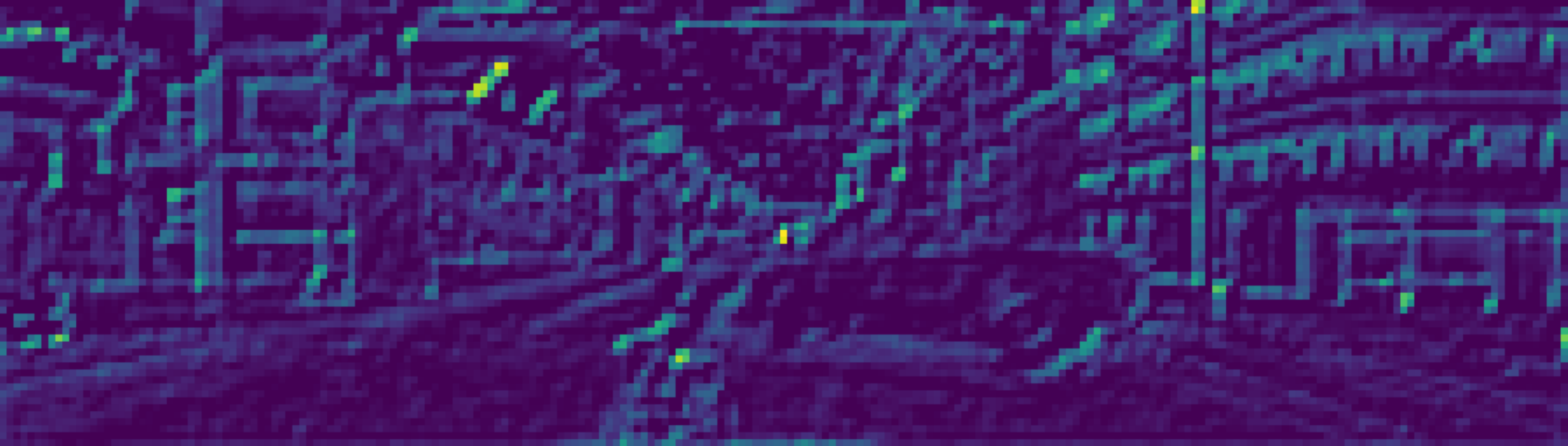} && \includegraphics[width=2cm]{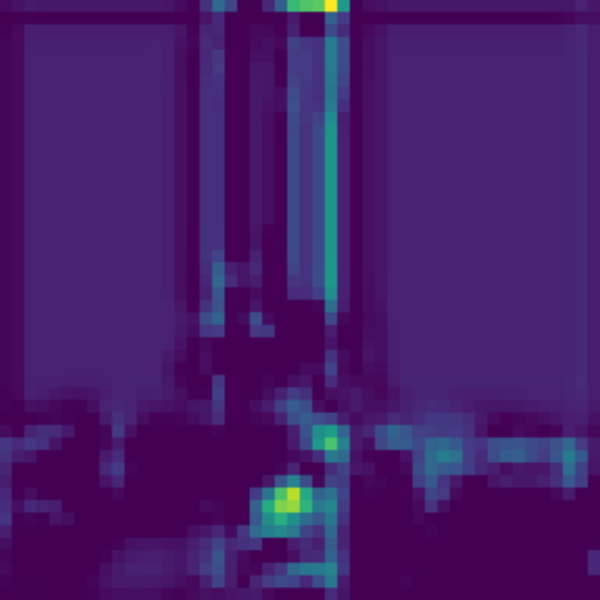} & 
  \includegraphics[width=2cm]{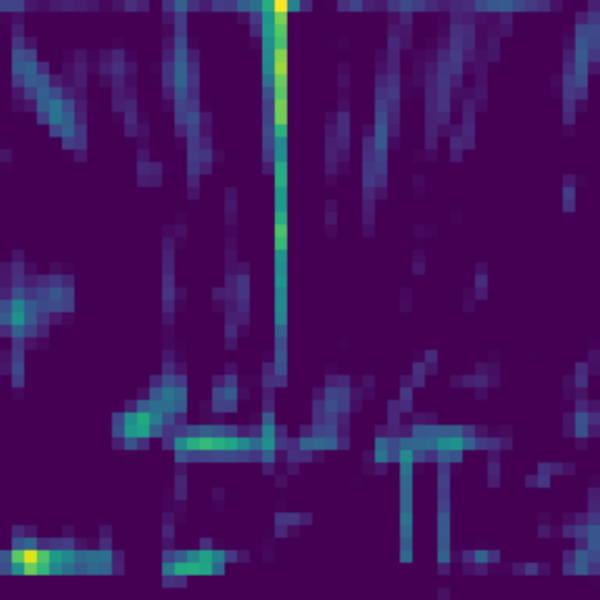} & 
  \includegraphics[width=2cm]{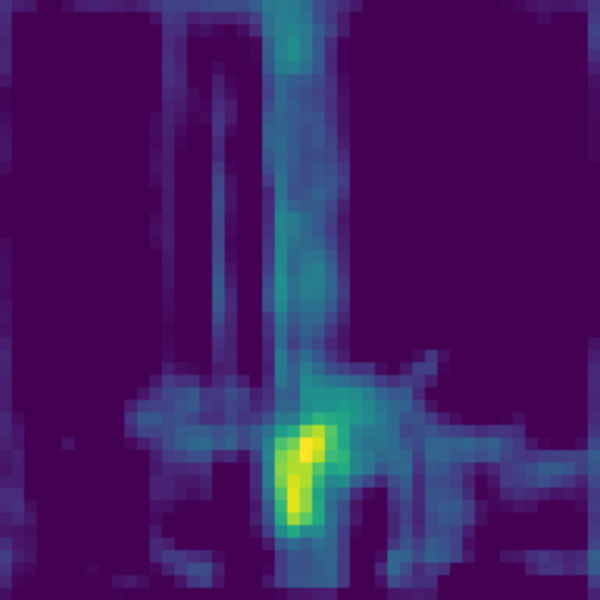} 
    \\

  \includegraphics[width=5.3cm,height=2cm]{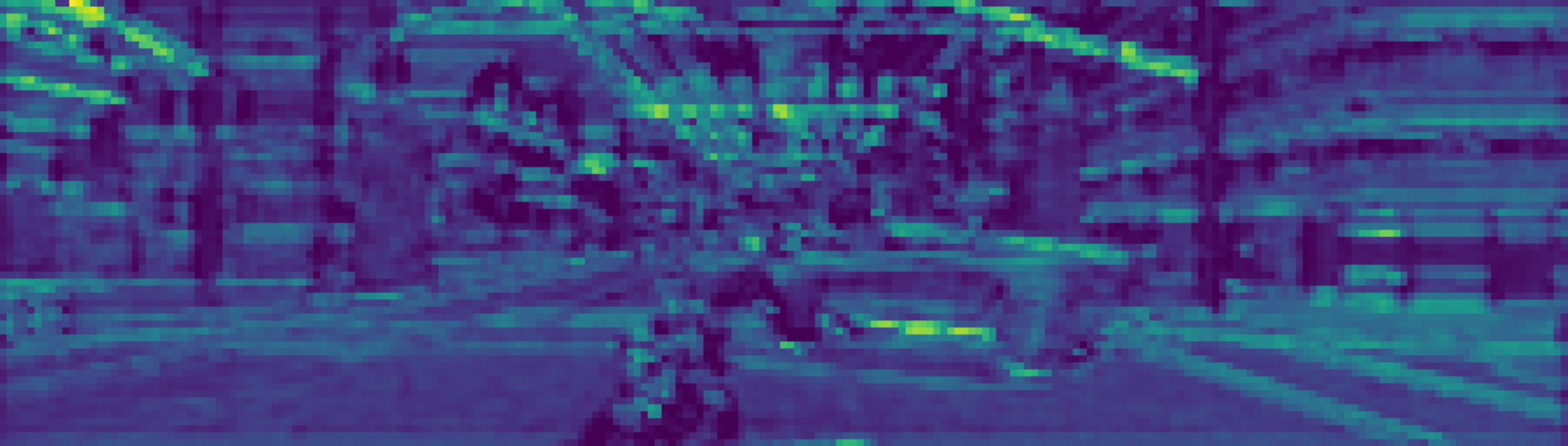} && \includegraphics[width=2cm]{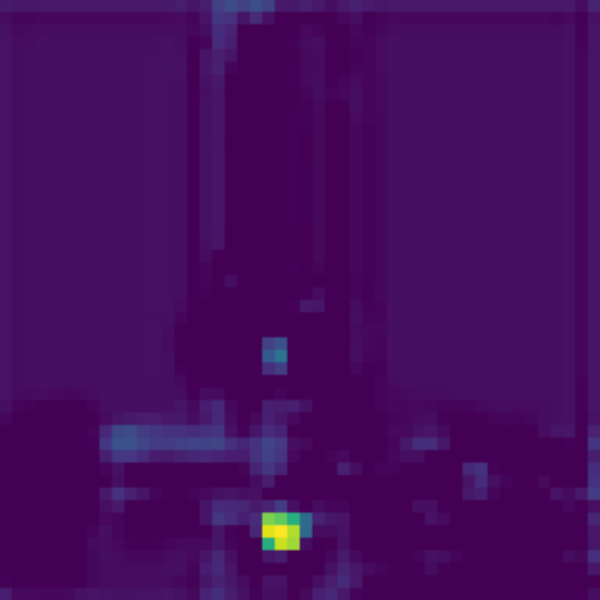} & 
  \includegraphics[width=2cm]{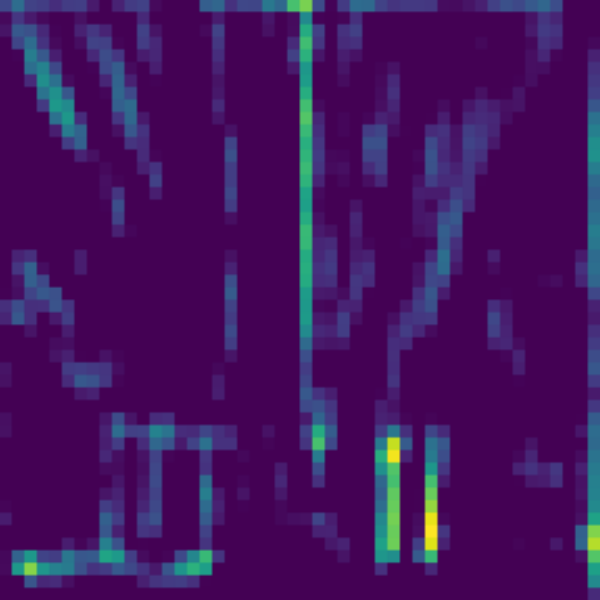} & 
  \includegraphics[width=2cm]{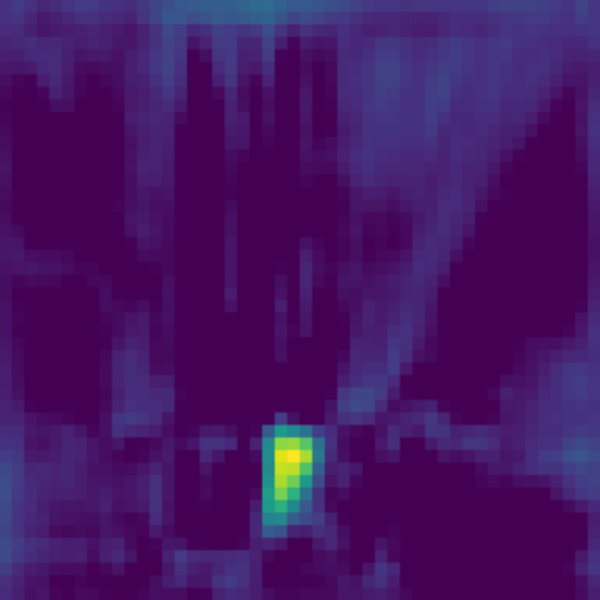} 
    \\
    
    \midrule

    \includegraphics[width=5.3cm,height=2cm]{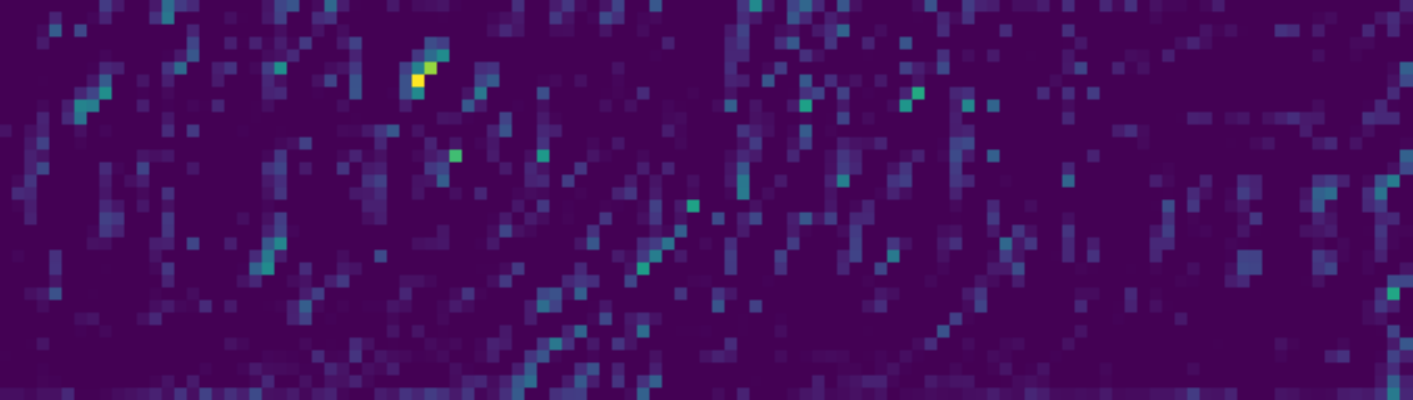} && \includegraphics[width=2cm]{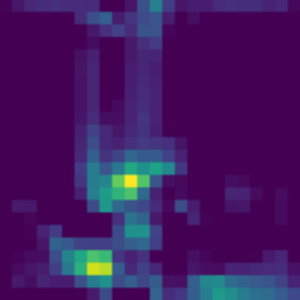} & 
  \includegraphics[width=2cm]{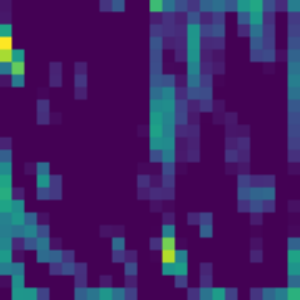} & 
  \includegraphics[width=2cm]{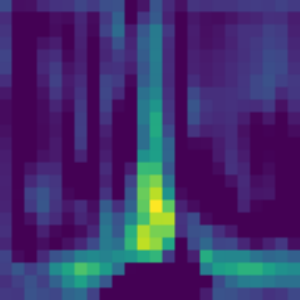} 
    \\

  \includegraphics[width=5.3cm,height=2cm]{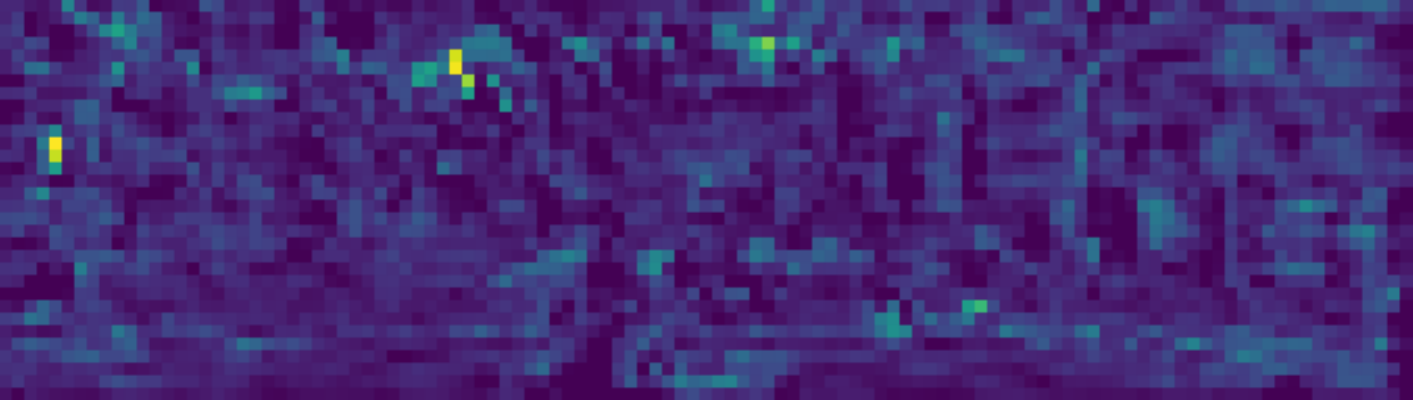} && \includegraphics[width=2cm]{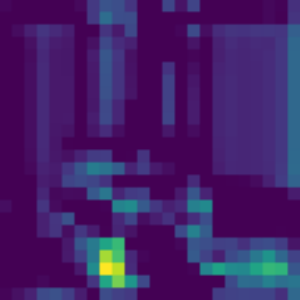} & 
  \includegraphics[width=2cm]{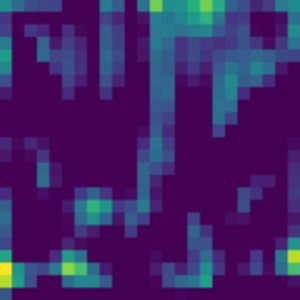} & 
  \includegraphics[width=2cm]{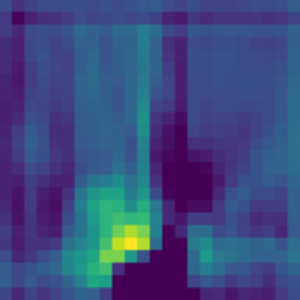} 
    \\

\end{tabular}
    \caption{\textbf{Visualization of Model Input and Features.} Analysis of a scenario where the agent is driving forward at an intersection, yielding to traffic coming from the right.}
    \label{fig:feature_h}
\end{figure*}

\begin{figure*}
    \centering
    \setlength{\tabcolsep}{2pt}
    \begin{tabular}{ccccc}
    \multicolumn{4}{c}{\textbf{Image}}&
    \multicolumn{1}{c}{\textbf{BEV}}
    
    \\
    \multicolumn{4}{c}{\includegraphics[height=2cm]{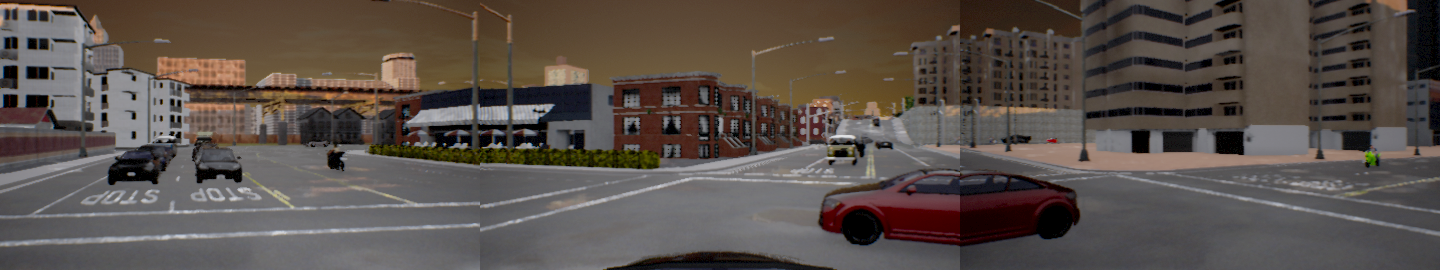}}&
    \multicolumn{1}{c}{\includegraphics[height=2cm]{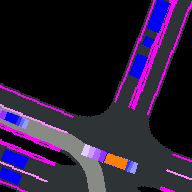}}
 
    \\
      \textbf{No Alignment Module} & & \textbf{Privileged} & \textbf{Output Distillation} & \textbf{CaT} 
    \\  
  \includegraphics[width=5.3cm,height=2cm]{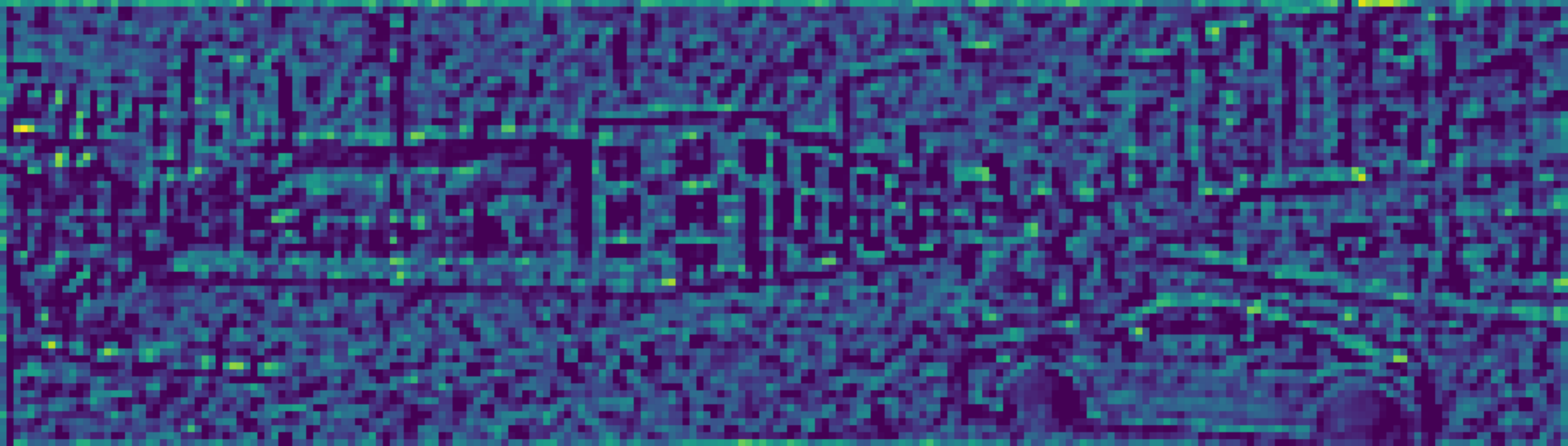} && \includegraphics[width=2cm]{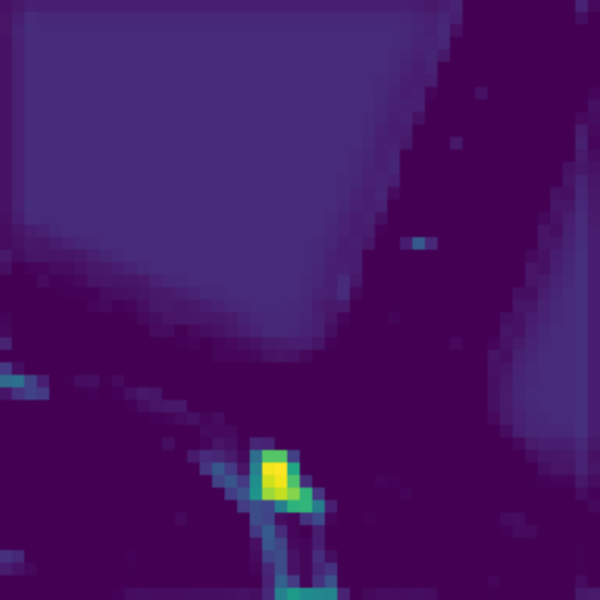} & 
  \includegraphics[width=2cm]{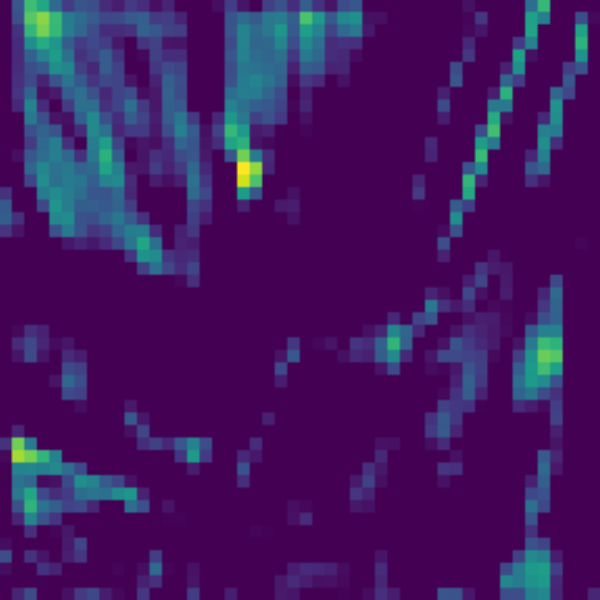} & 
  \includegraphics[width=2cm]{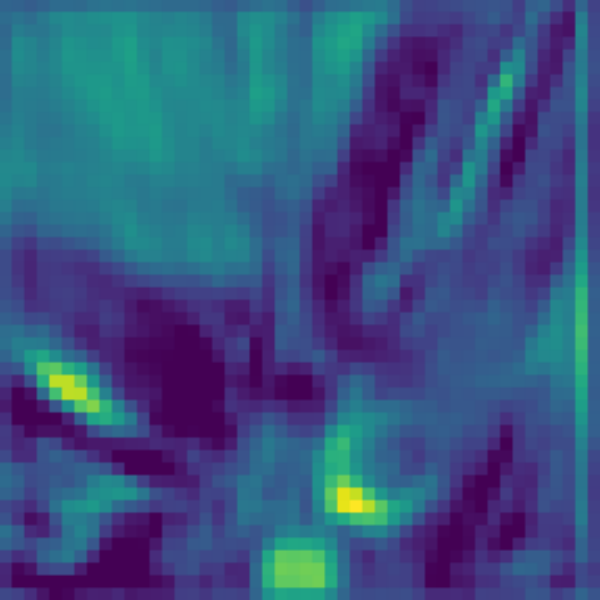} 
    \\

  \includegraphics[width=5.3cm,height=2cm]{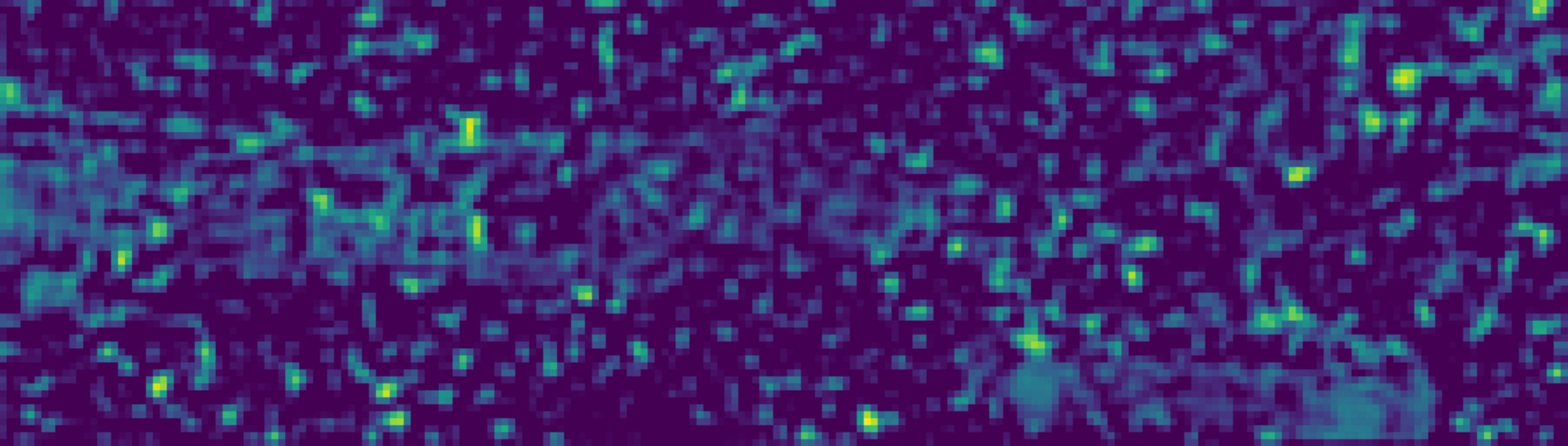} && \includegraphics[width=2cm]{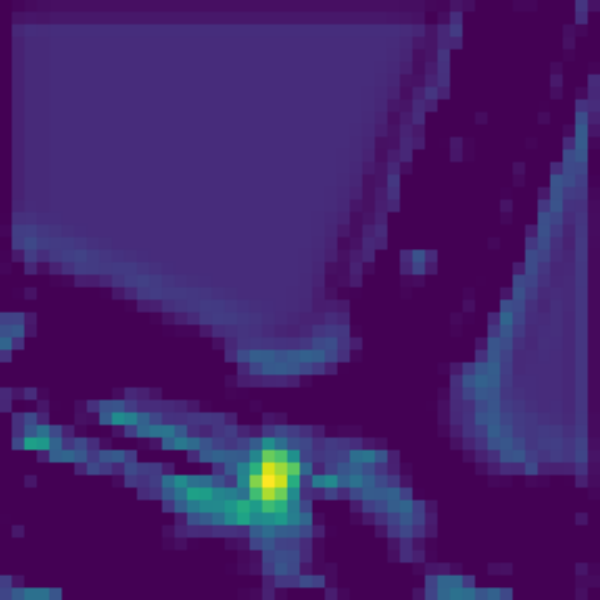} & 
  \includegraphics[width=2cm]{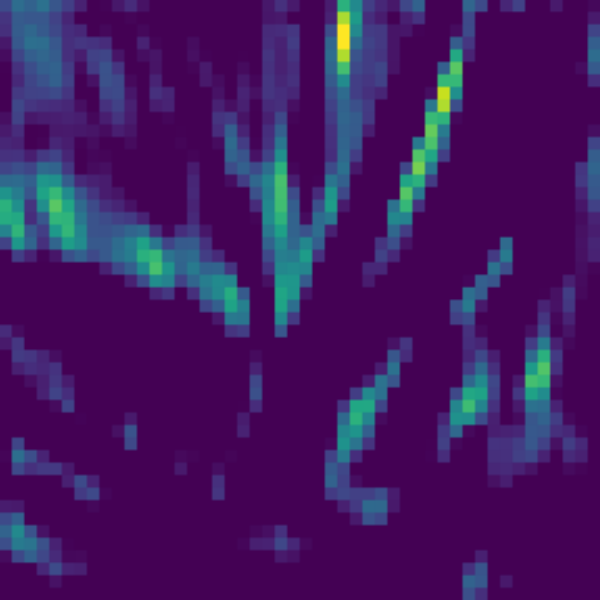} & 
  \includegraphics[width=2cm]{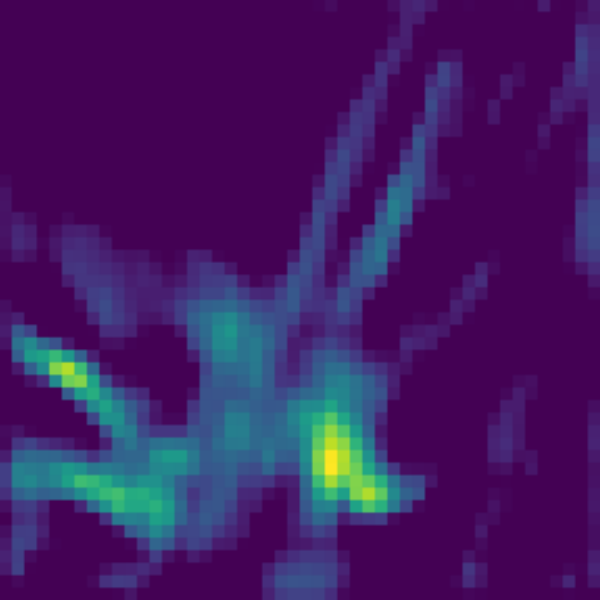} 
    \\
    
    \midrule

    \includegraphics[width=5.3cm,height=2cm]{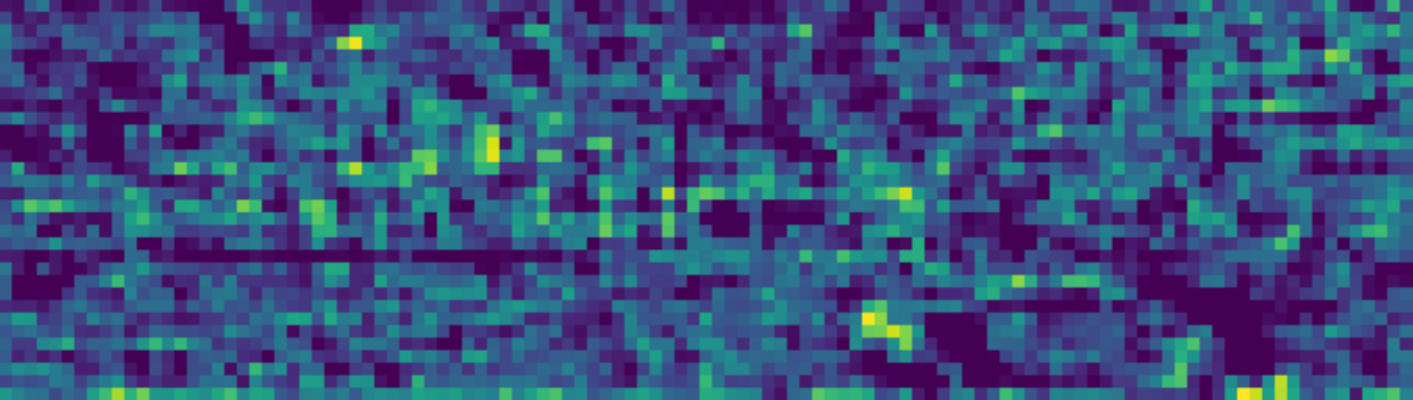} && \includegraphics[width=2cm]{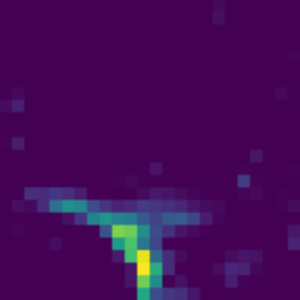} & 
  \includegraphics[width=2cm]{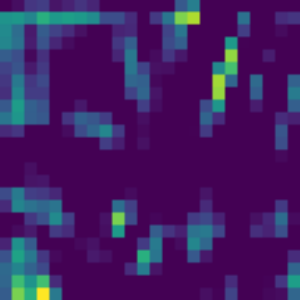} & 
  \includegraphics[width=2cm]{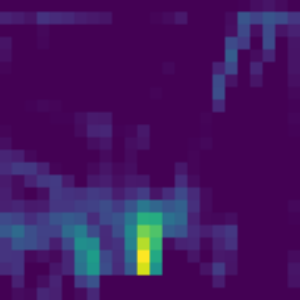} 
    \\

  \includegraphics[width=5.3cm,height=2cm]{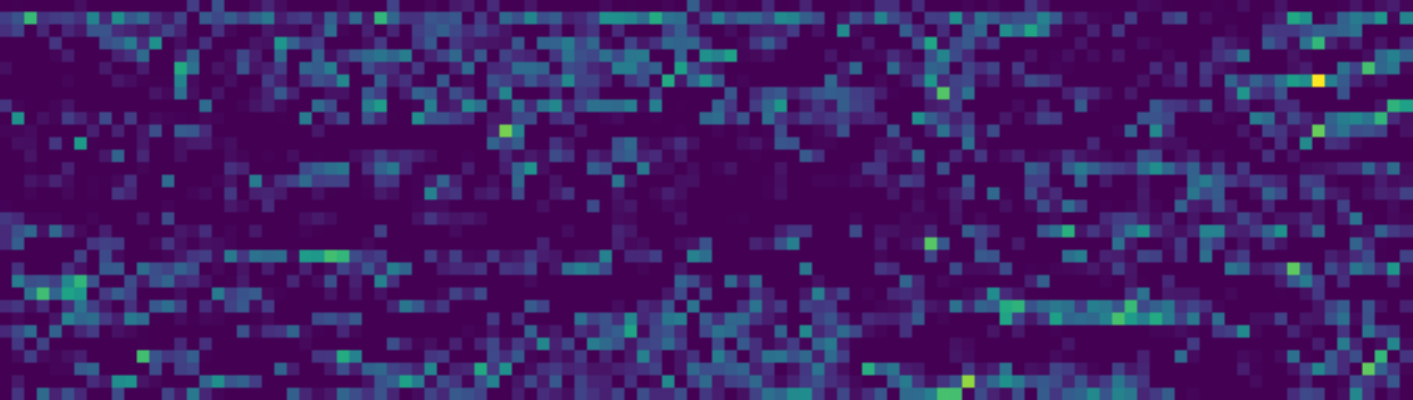} && \includegraphics[width=2cm]{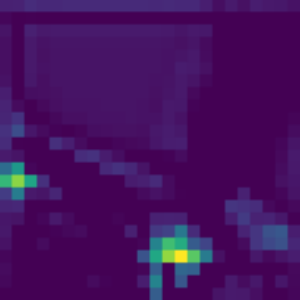} & 
  \includegraphics[width=2cm]{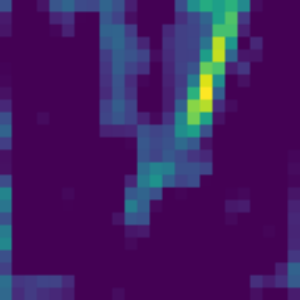} & 
  \includegraphics[width=2cm]{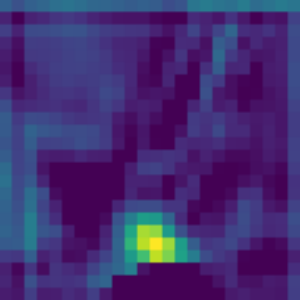} 
    \\

\end{tabular}
    \caption{\textbf{Visualization of Model Input and Features.} Analysis of a scenario where the agent is turning left at an intersection, yielding to traffic coming from the right at dawn.}
    \label{fig:feature_i}
\end{figure*}

\begin{figure*}
    \centering
    \setlength{\tabcolsep}{2pt}
    \begin{tabular}{ccccc}
    \multicolumn{4}{c}{\textbf{Image}}&
    \multicolumn{1}{c}{\textbf{BEV}}

    \\
    \multicolumn{4}{c}{\includegraphics[height=2cm]{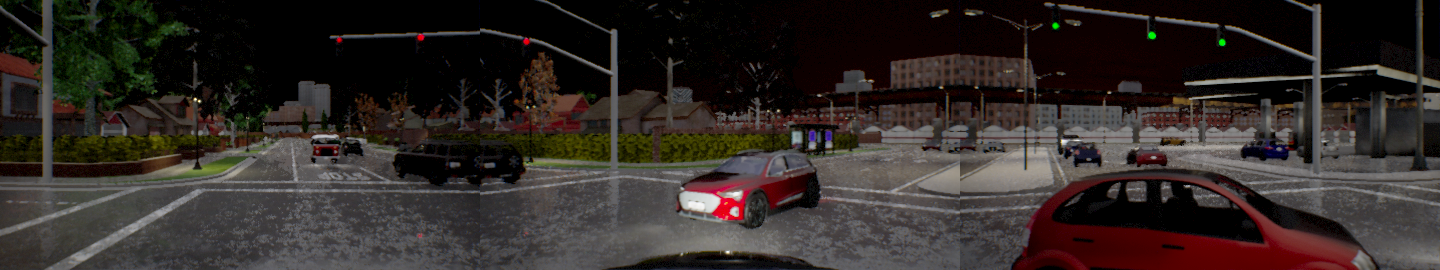}}&
    \multicolumn{1}{c}{\includegraphics[height=2cm]{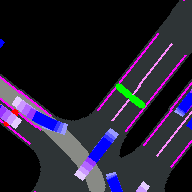}}
 
    \\
      \textbf{No Alignment Module} & & \textbf{Privileged} & \textbf{Output Distillation} & \textbf{CaT} 
    \\  
  \includegraphics[width=5.3cm,height=2cm]{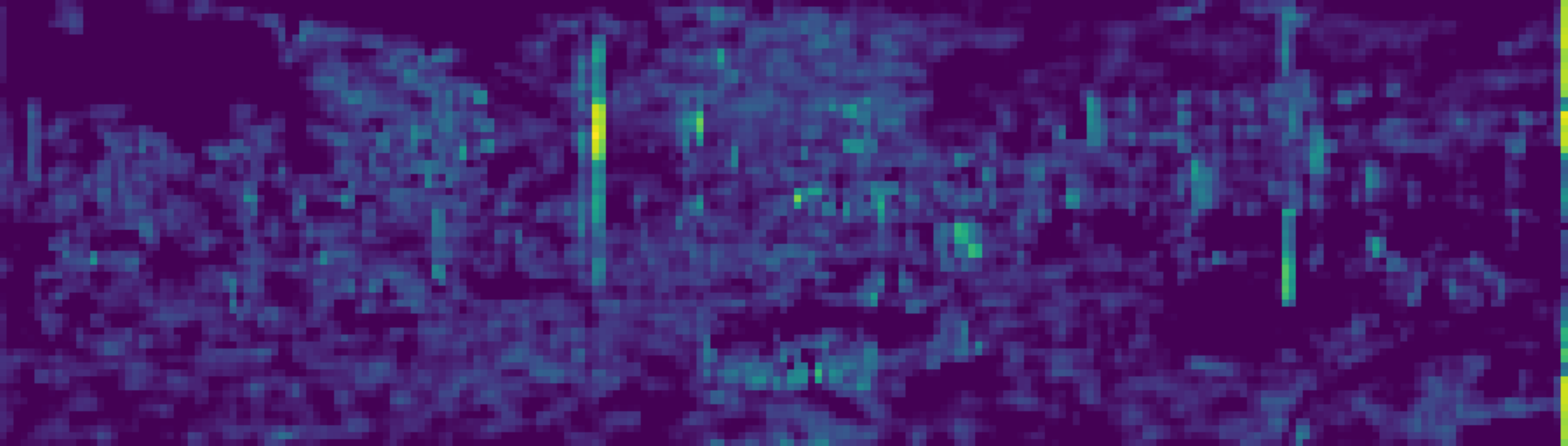} && \includegraphics[width=2cm]{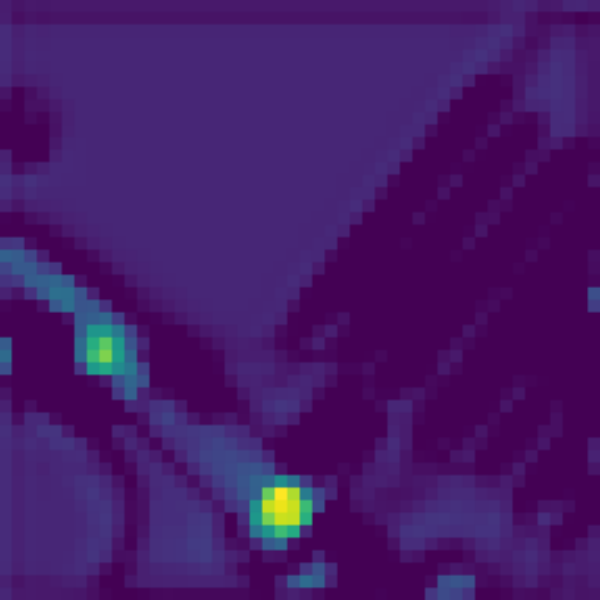} & 
  \includegraphics[width=2cm]{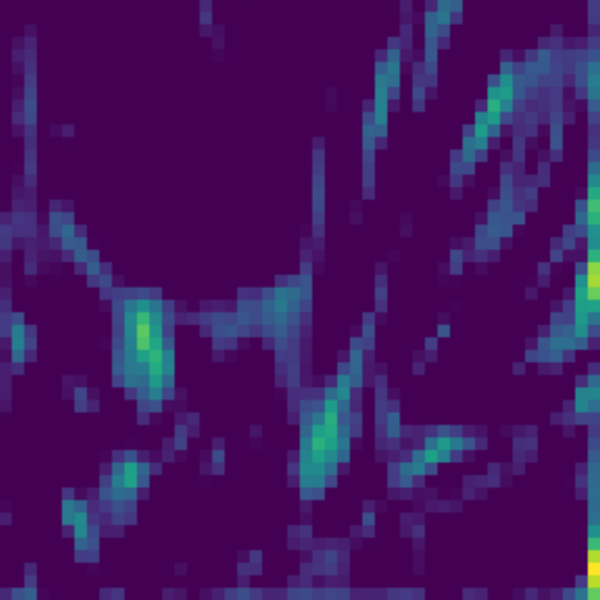} & 
  \includegraphics[width=2cm]{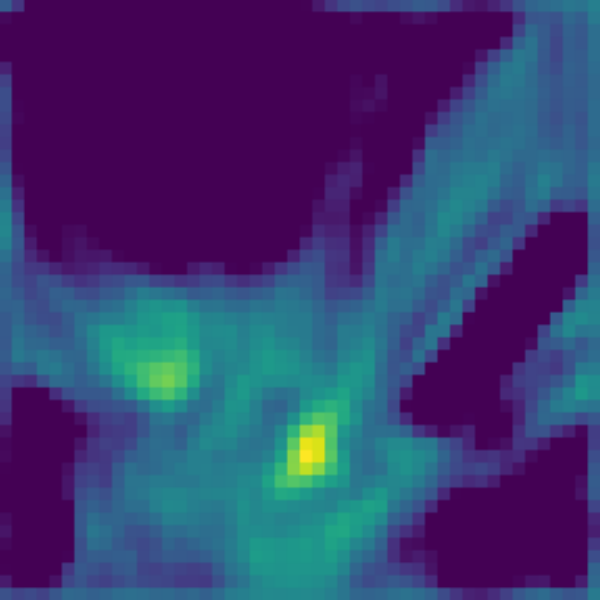} 
    \\

  \includegraphics[width=5.3cm,height=2cm]{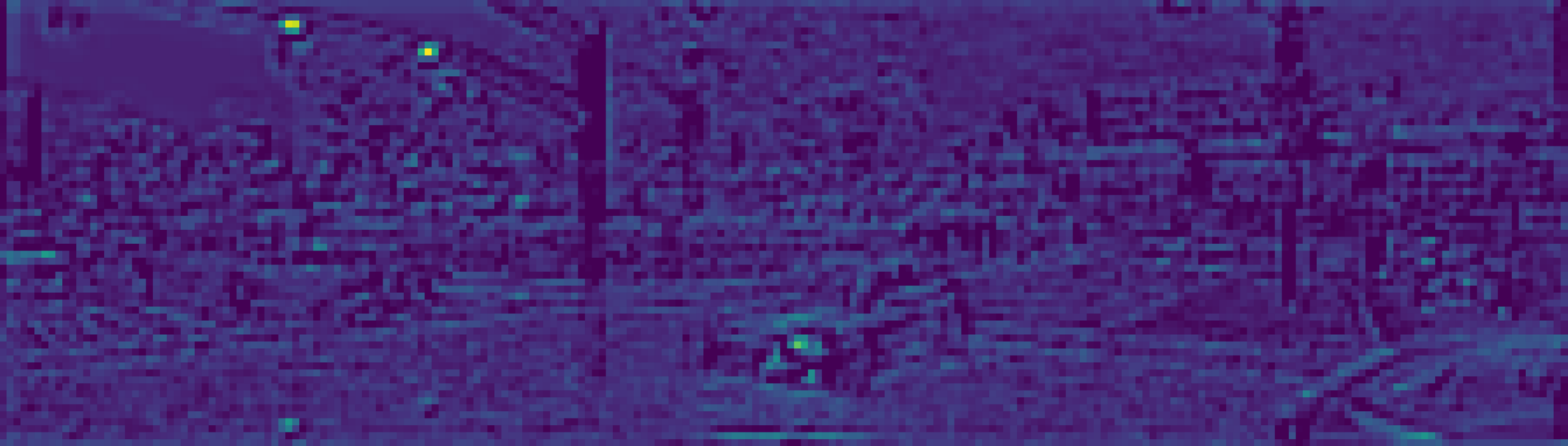} && \includegraphics[width=2cm]{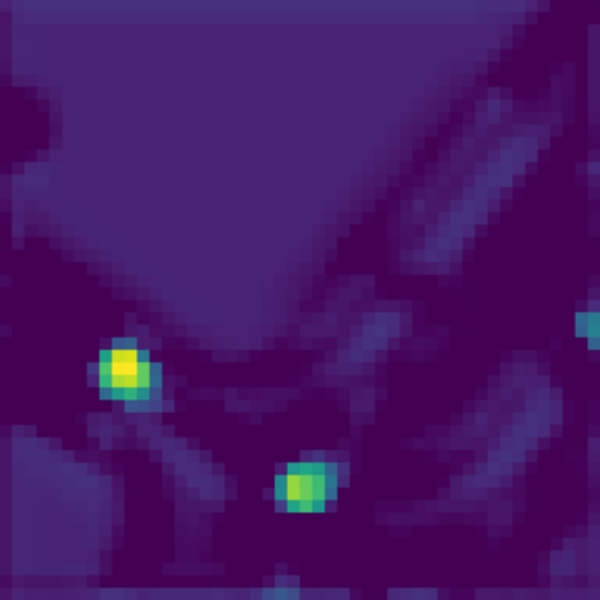} & 
  \includegraphics[width=2cm]{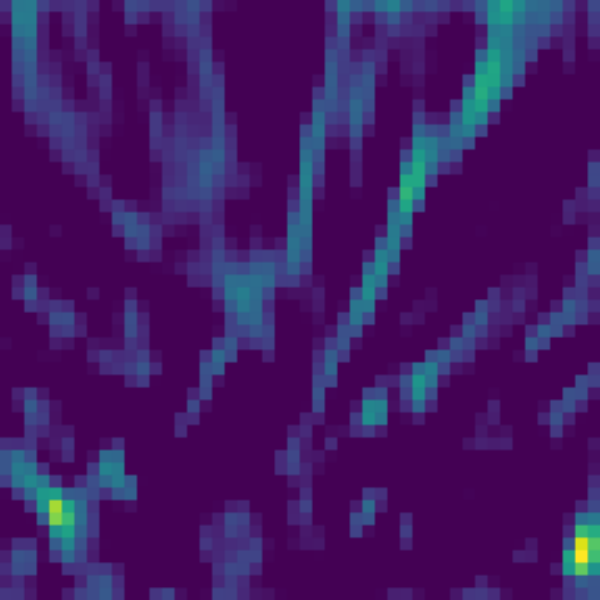} & 
  \includegraphics[width=2cm]{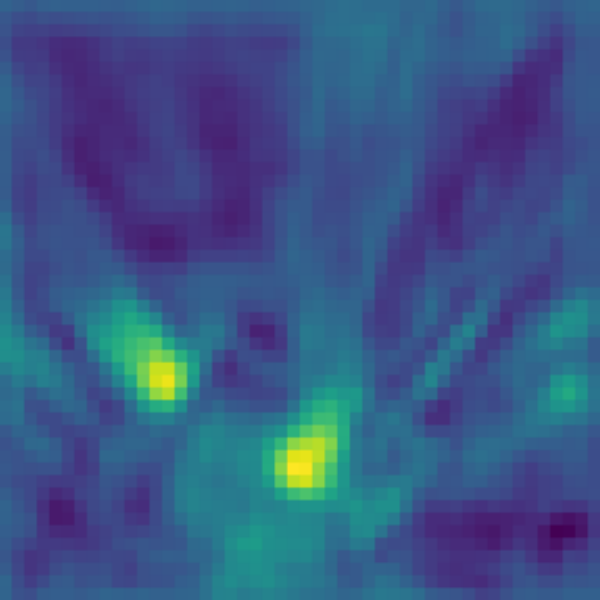} 
    \\

    \midrule
    
    \includegraphics[width=5.3cm,height=2cm]{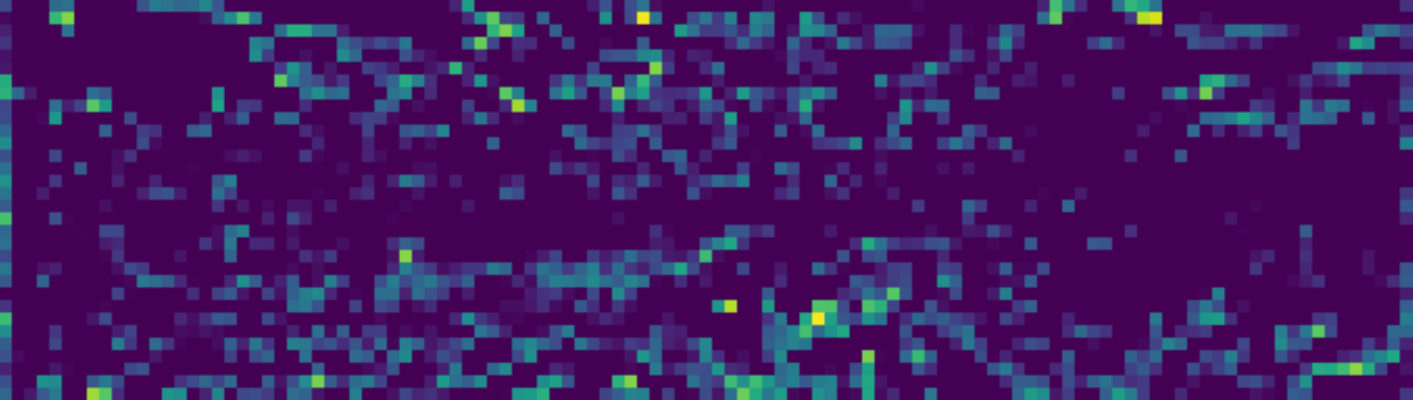} && \includegraphics[width=2cm]{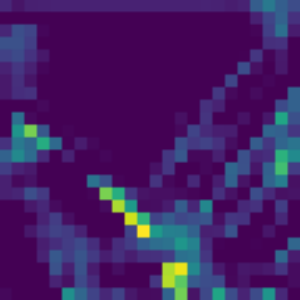} & 
  \includegraphics[width=2cm]{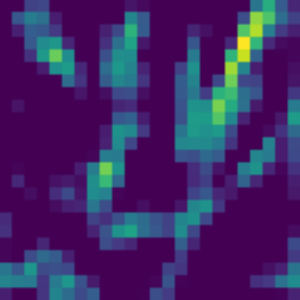} & 
  \includegraphics[width=2cm]{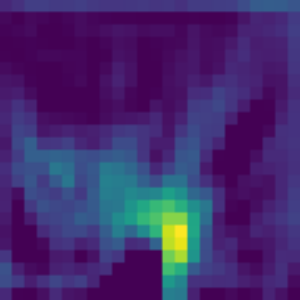} 
    \\

  \includegraphics[width=5.3cm,height=2cm]{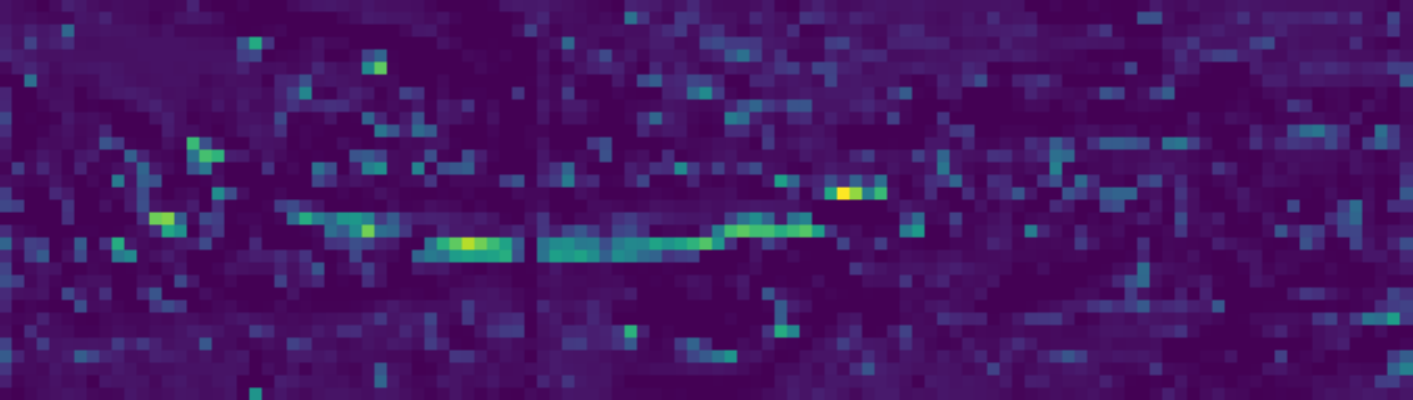} && \includegraphics[width=2cm]{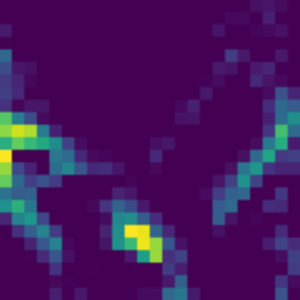} & 
  \includegraphics[width=2cm]{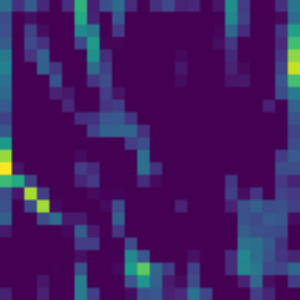} & 
  \includegraphics[width=2cm]{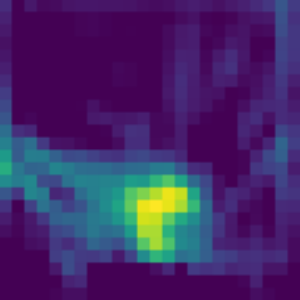} 
    \\

\end{tabular}
    \caption{\textbf{Visualization of Model Input and Features.} Analysis of a scenario where the agent is making an unprotected left turn on a hard-rain night conditions.}
    \label{fig:feature_j}
\end{figure*}

\begin{figure*}
    \centering
    \setlength{\tabcolsep}{2pt}
    \begin{tabular}{ccccc}
    \multicolumn{4}{c}{\textbf{Image}}&
    \multicolumn{1}{c}{\textbf{BEV}}

    \\
    \multicolumn{4}{c}{\includegraphics[height=2cm]{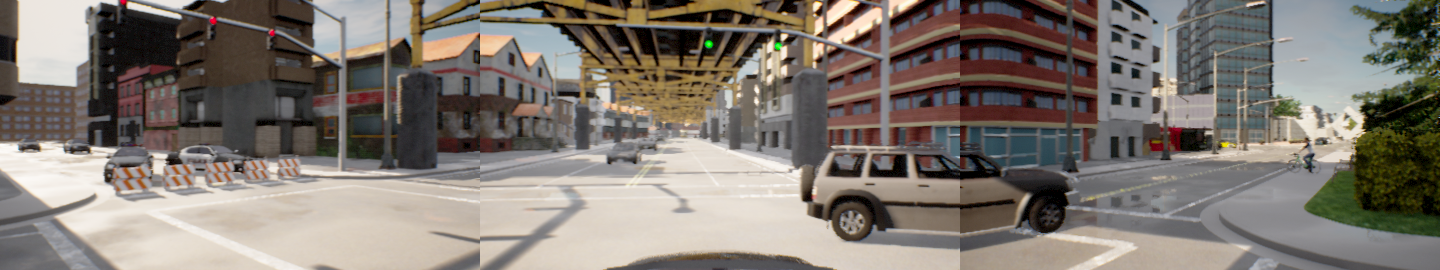}}&
    \multicolumn{1}{c}{\includegraphics[height=2cm]{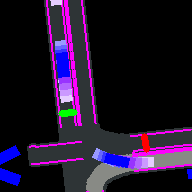}}
 
    \\
      \textbf{No Alignment Module} & & \textbf{Privileged} & \textbf{Output Distillation} & \textbf{CaT} 
    \\  
  \includegraphics[width=5.3cm,height=2cm]{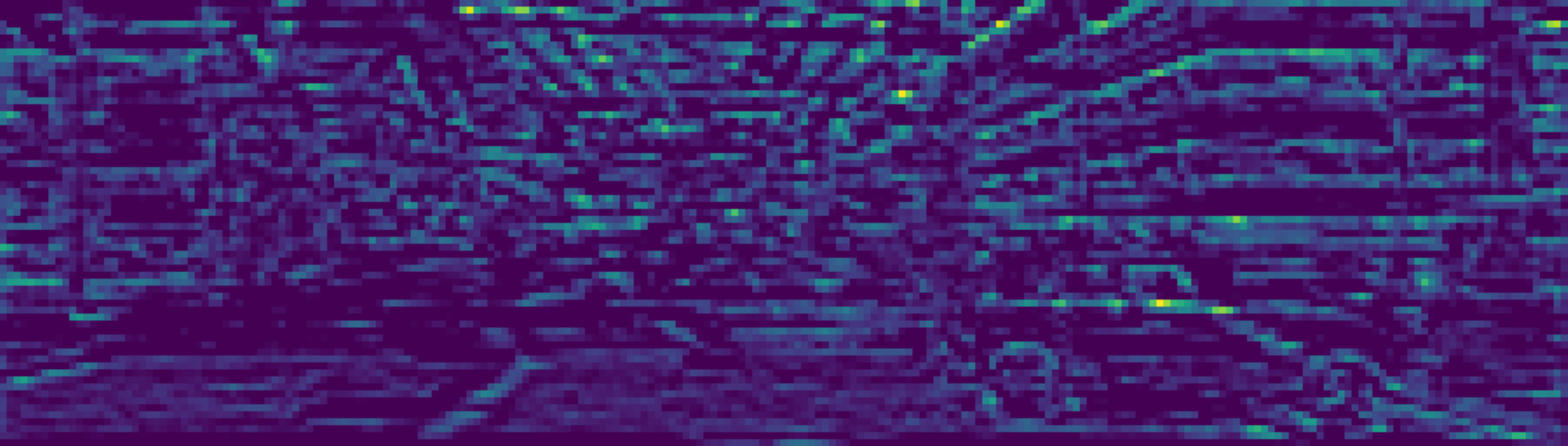} && \includegraphics[width=2cm]{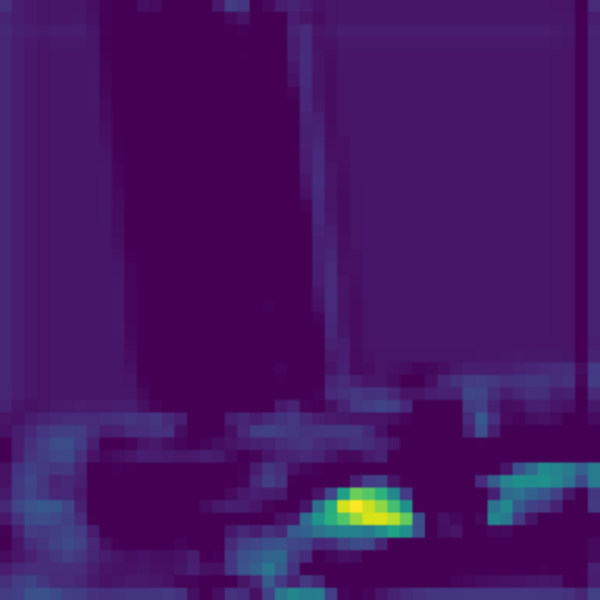} & 
  \includegraphics[width=2cm]{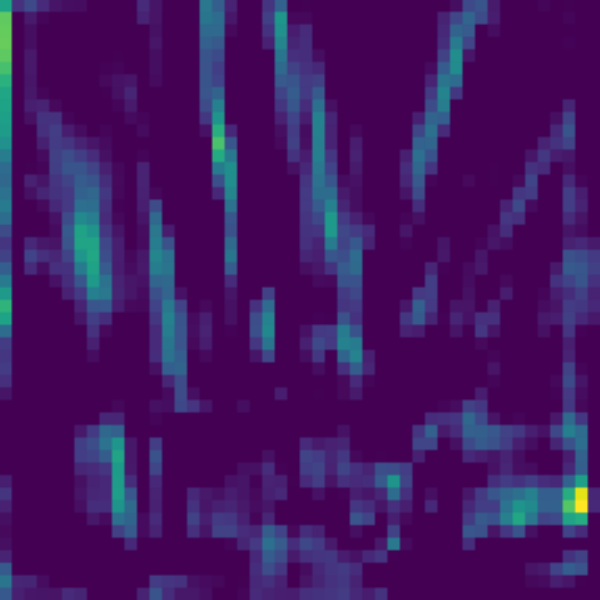} & 
  \includegraphics[width=2cm]{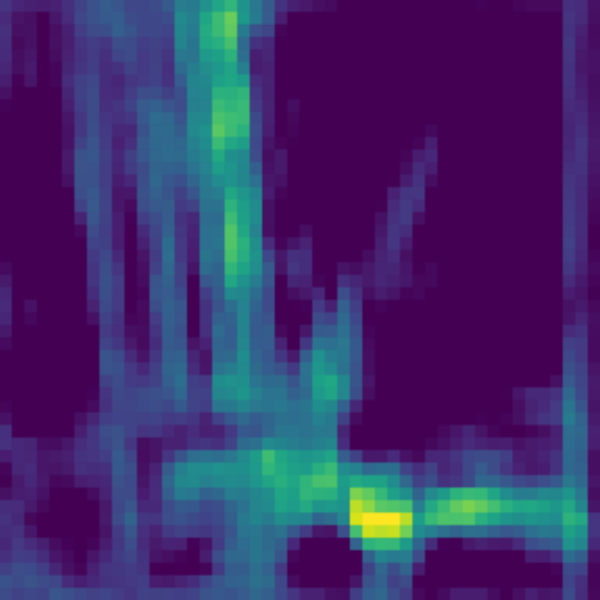} 
    \\

  \includegraphics[width=5.3cm,height=2cm]{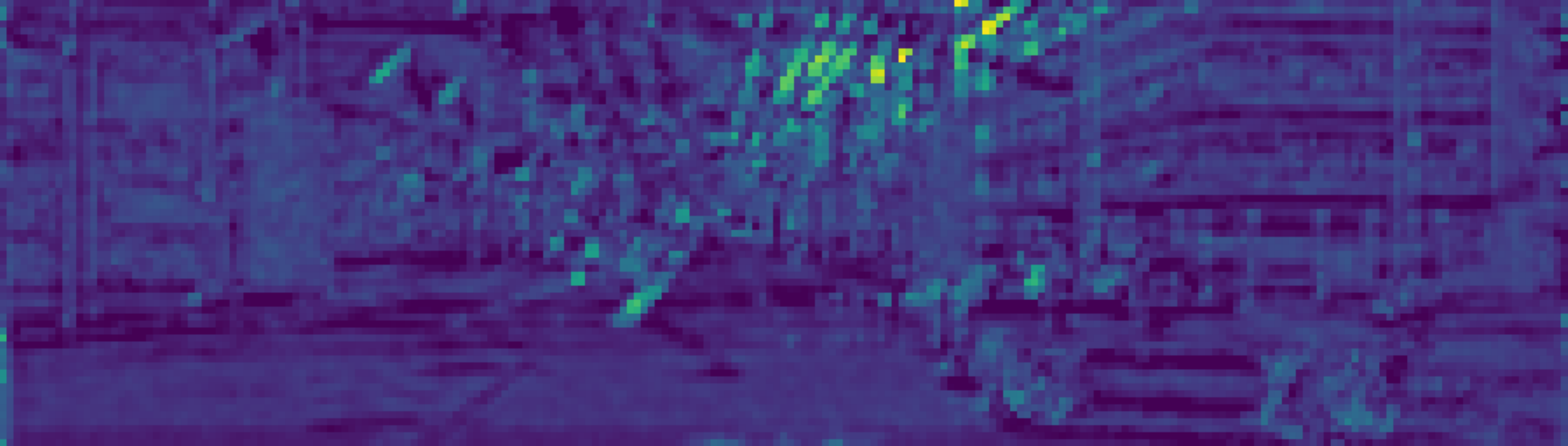} && \includegraphics[width=2cm]{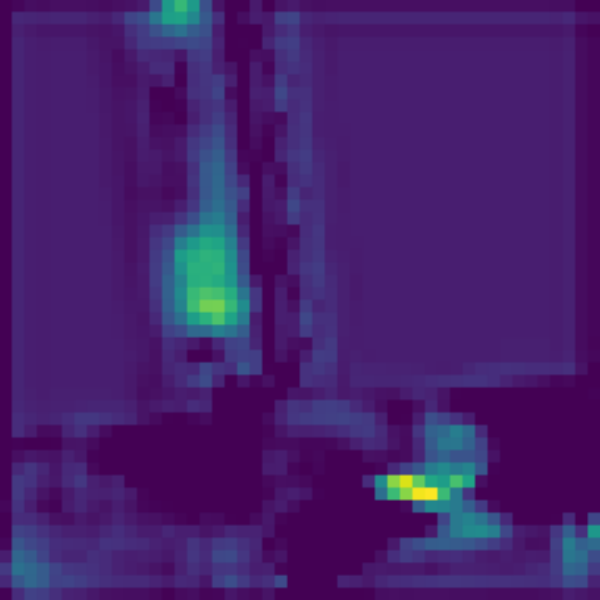} & 
  \includegraphics[width=2cm]{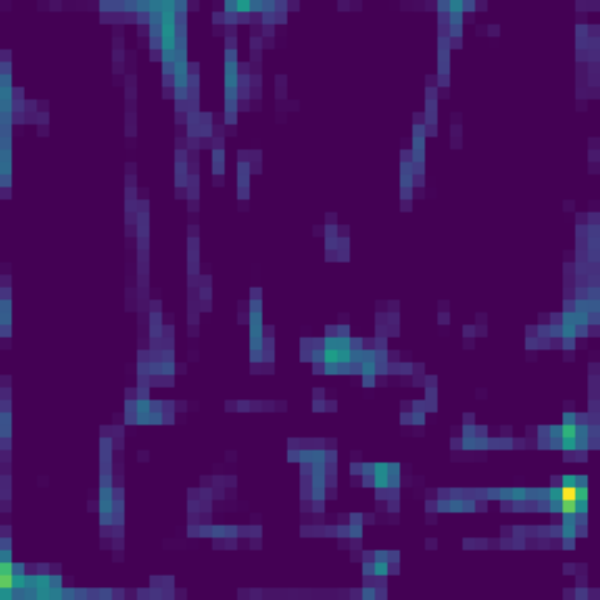} & 
  \includegraphics[width=2cm]{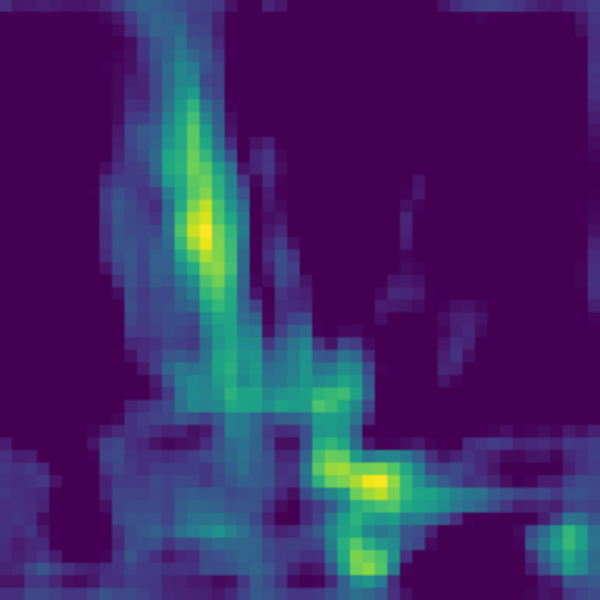} 
    \\
    
    \midrule

    \includegraphics[width=5.3cm,height=2cm]{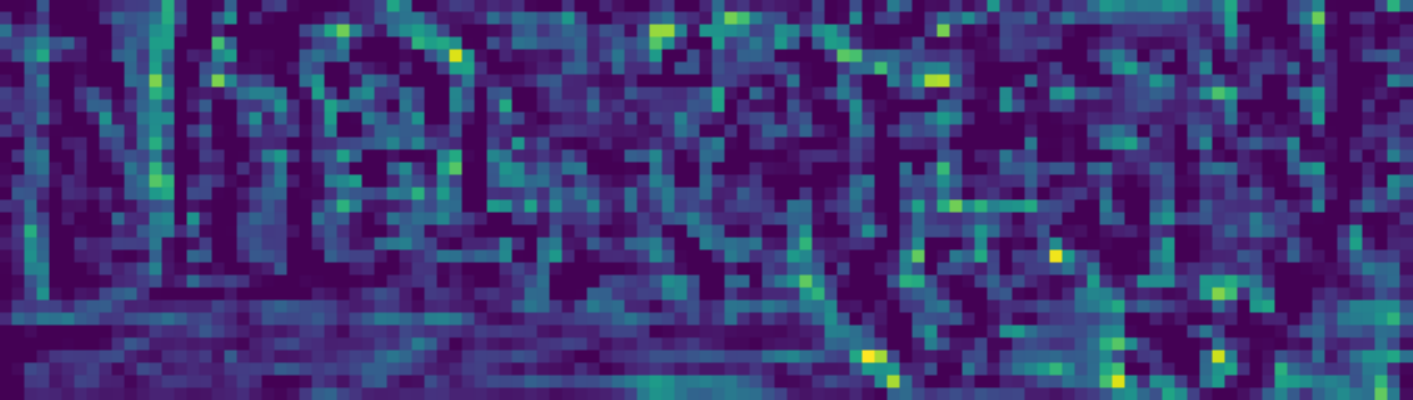} && \includegraphics[width=2cm]{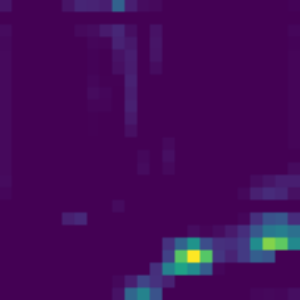} & 
  \includegraphics[width=2cm]{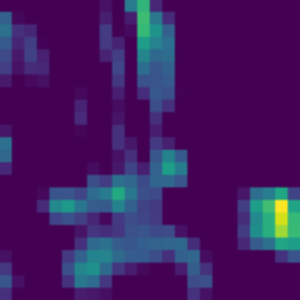} & 
  \includegraphics[width=2cm]{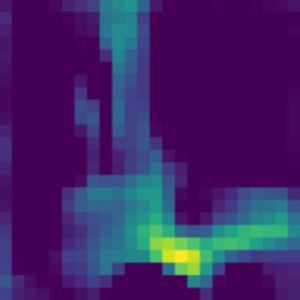} 
    \\

  \includegraphics[width=5.3cm,height=2cm]{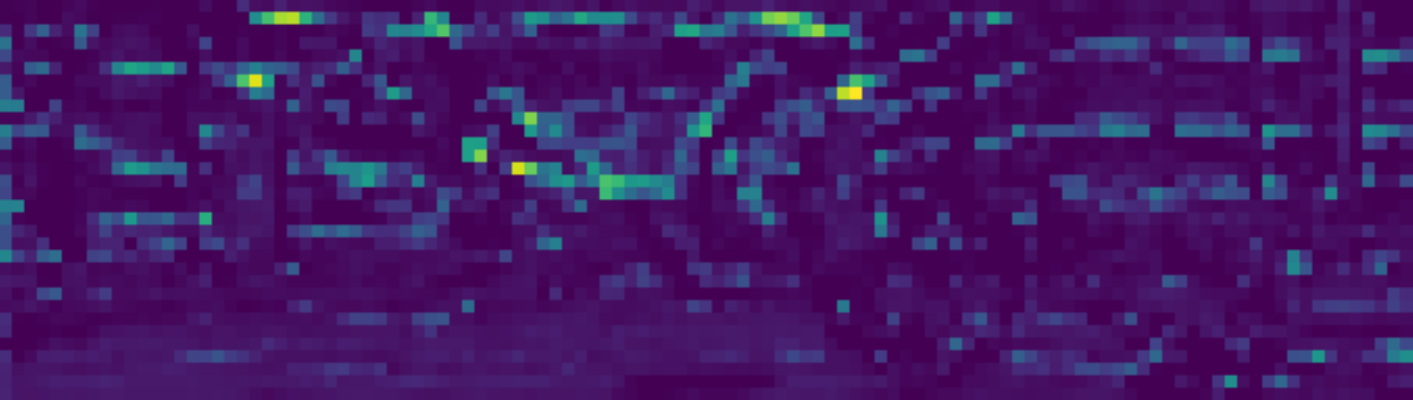} && \includegraphics[width=2cm]{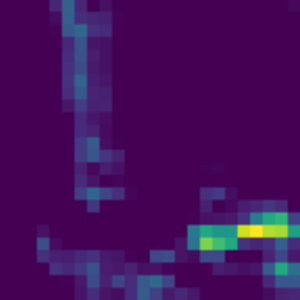} & 
  \includegraphics[width=2cm]{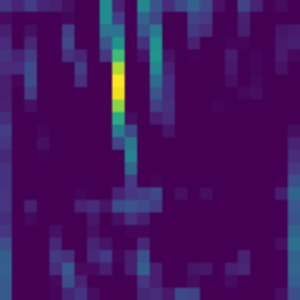} & 
  \includegraphics[width=2cm]{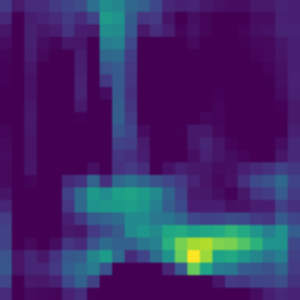} 
    \\

\end{tabular}
    \caption{\textbf{Visualization of Model Input and Features.} Analysis of a scenario where the agent is turning right at an intersection, yielding to crossing traffic.}
    \label{fig:feature_k}
\end{figure*}

\begin{figure*}
    \centering
    \setlength{\tabcolsep}{2pt}
    \begin{tabular}{ccccc}
    \multicolumn{4}{c}{\textbf{Image}}&
    \multicolumn{1}{c}{\textbf{BEV}}

    \\
    \multicolumn{4}{c}{\includegraphics[height=2cm]{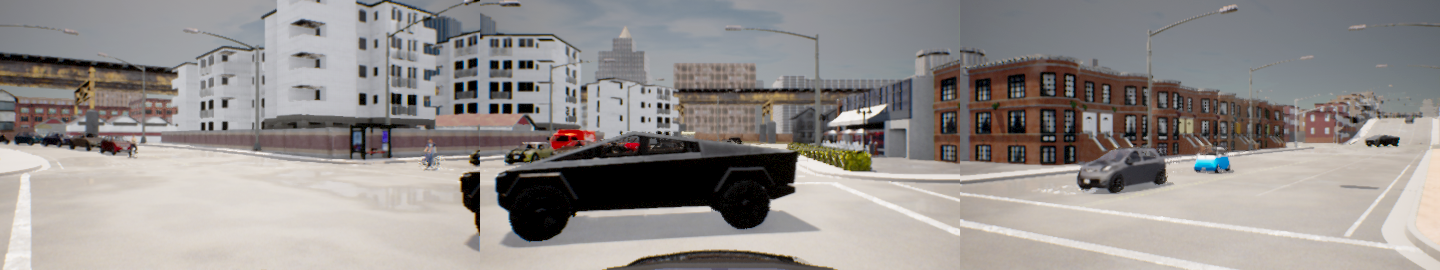}}&
    \multicolumn{1}{c}{\includegraphics[height=2cm]{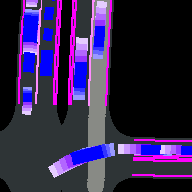}}
 
    \\
      \textbf{No Alignment Module} & & \textbf{Privileged} & \textbf{Output Distillation} & \textbf{CaT} 
    \\  
  \includegraphics[width=5.3cm,height=2cm]{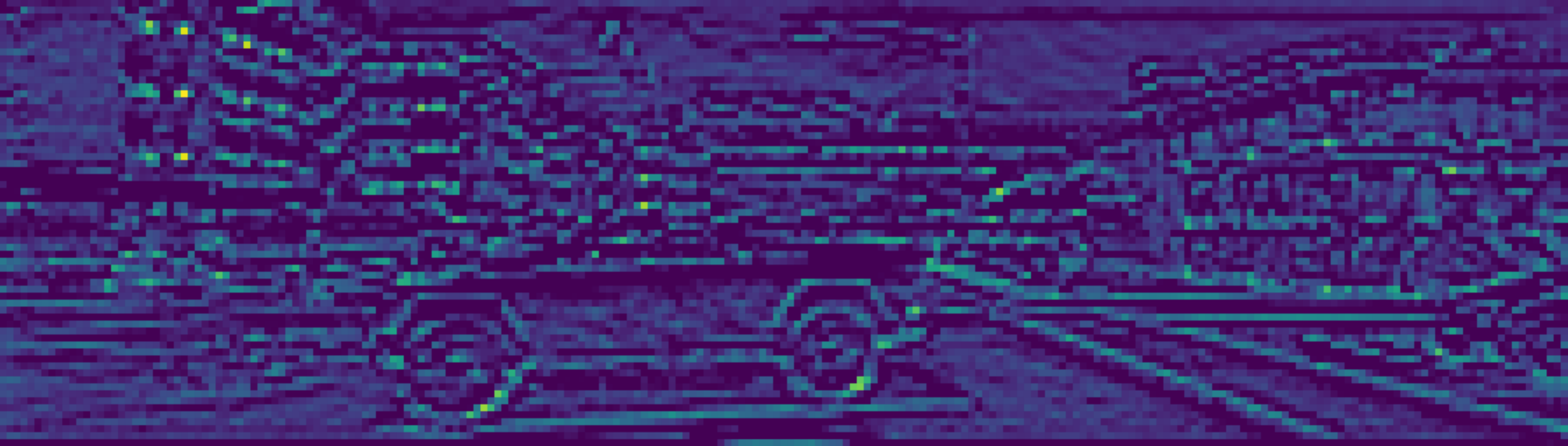} && \includegraphics[width=2cm]{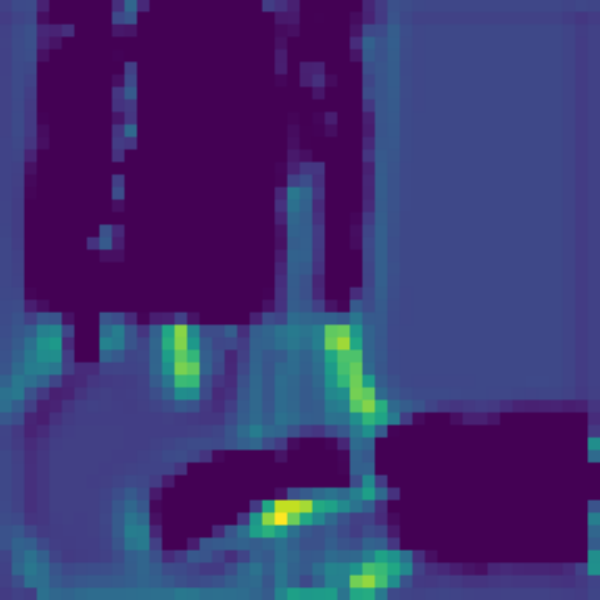} & 
  \includegraphics[width=2cm]{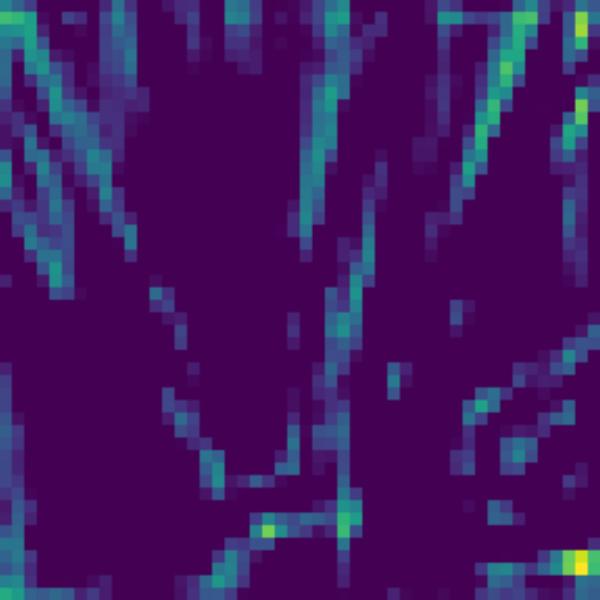} & 
  \includegraphics[width=2cm]{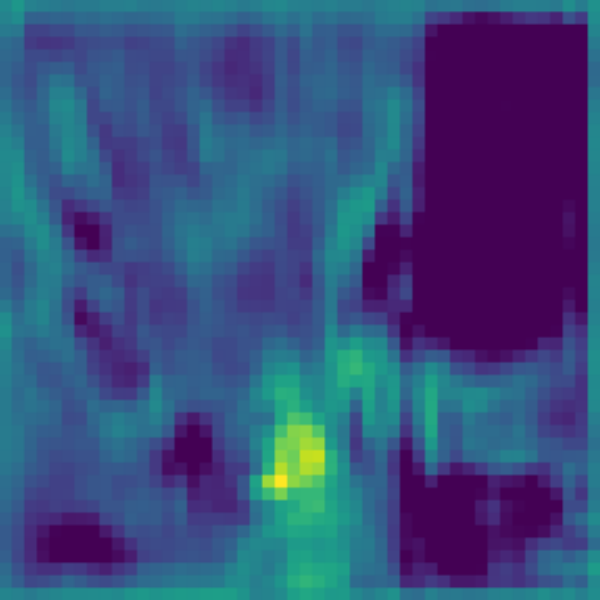} 
    \\

  \includegraphics[width=5.3cm,height=2cm]{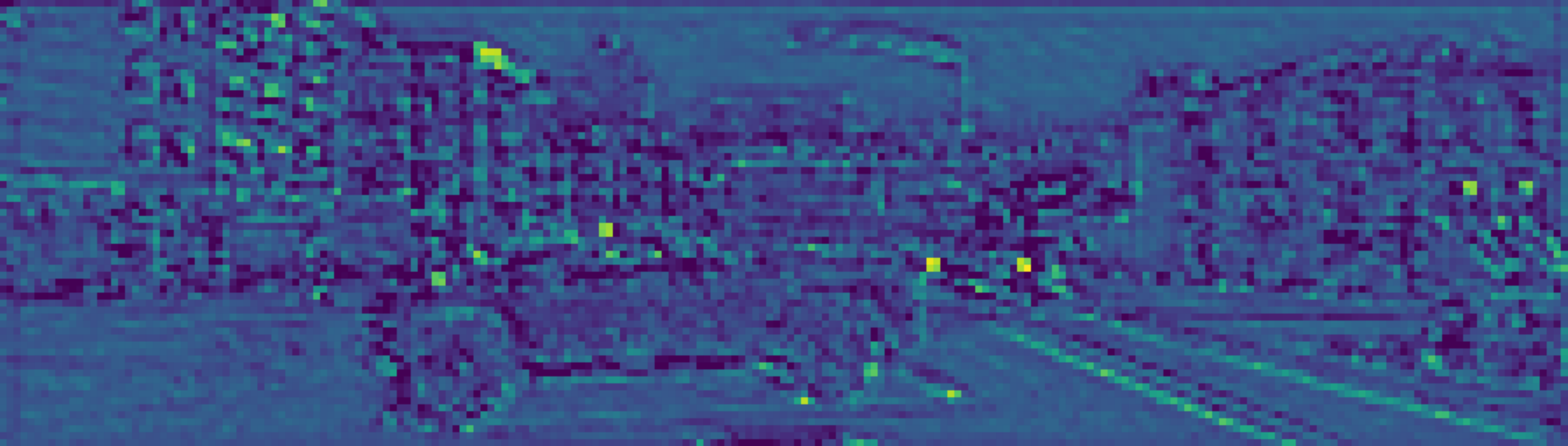} && \includegraphics[width=2cm]{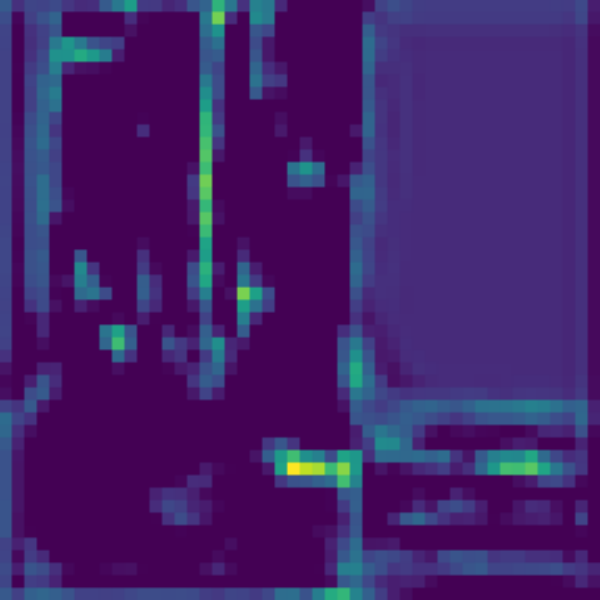} & 
  \includegraphics[width=2cm]{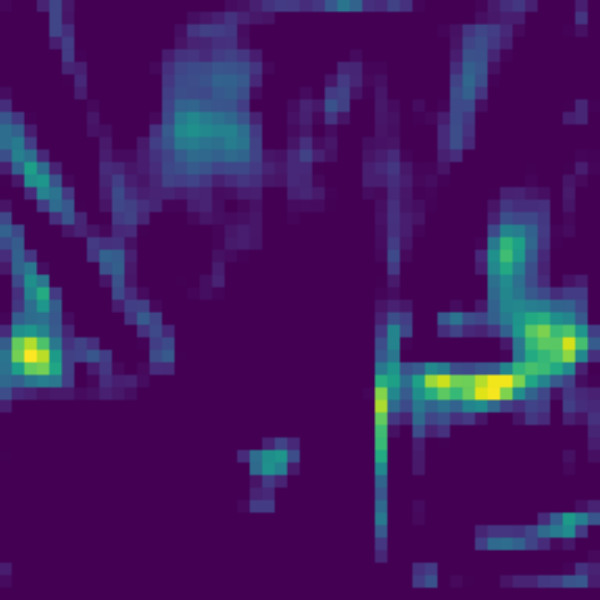} & 
  \includegraphics[width=2cm]{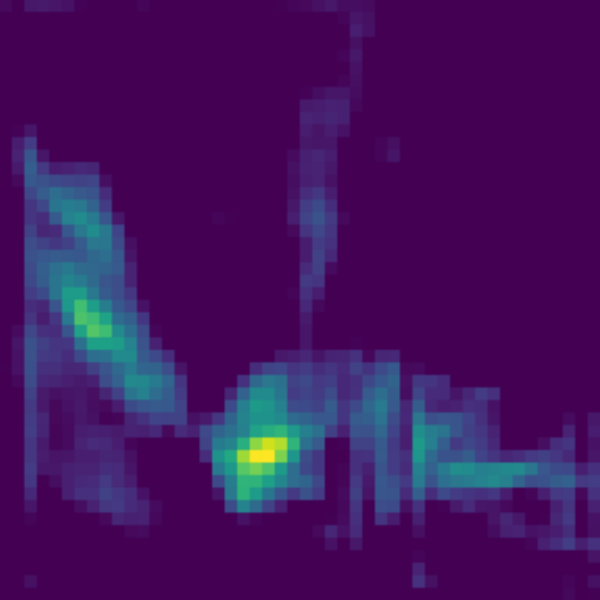} 
    \\
    
    \midrule
    
    \includegraphics[width=5.3cm,height=2cm]{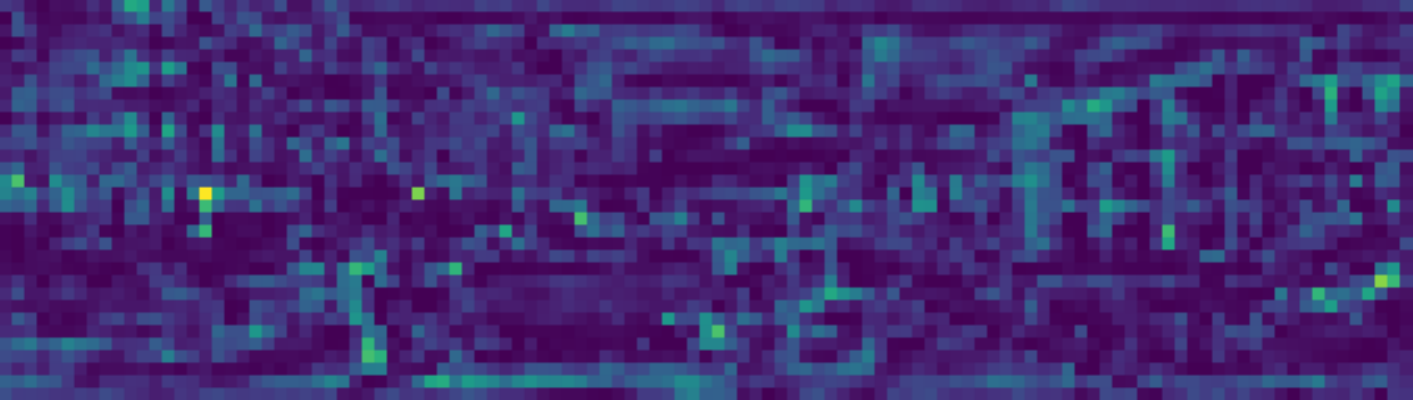} && \includegraphics[width=2cm]{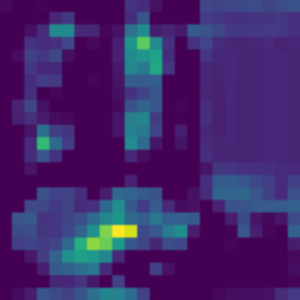} & 
  \includegraphics[width=2cm]{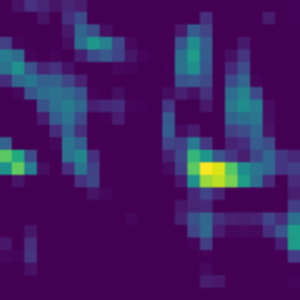} & 
  \includegraphics[width=2cm]{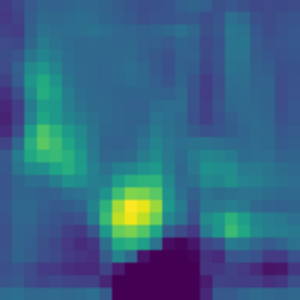} 
    \\

  \includegraphics[width=5.3cm,height=2cm]{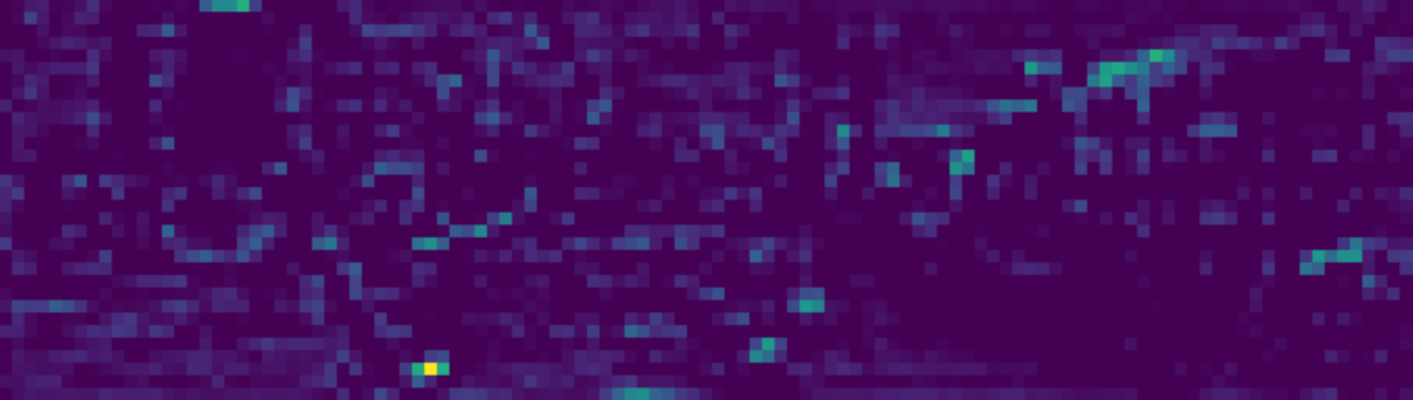} && \includegraphics[width=2cm]{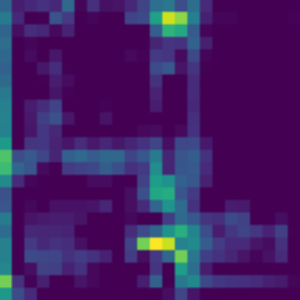} & 
  \includegraphics[width=2cm]{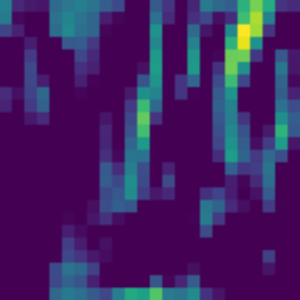} & 
  \includegraphics[width=2cm]{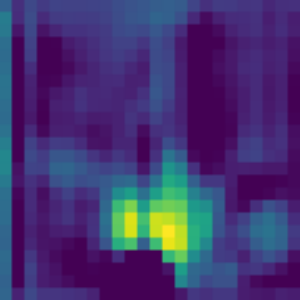} 
    \\

    \end{tabular}
    \caption{\textbf{Visualization of Model Input and Features.} Analysis of a scenario where the agent is negotiating with traffic at a no-signal intersection.}
    \label{fig:feature_l}
\end{figure*}

\begin{figure*}
    \centering
    \setlength{\tabcolsep}{2pt}
    \begin{tabular}{ccccc}
    \multicolumn{4}{c}{\textbf{Image}}&
    \multicolumn{1}{c}{\textbf{BEV}}

    \\
    \multicolumn{4}{c}{\includegraphics[height=2cm]{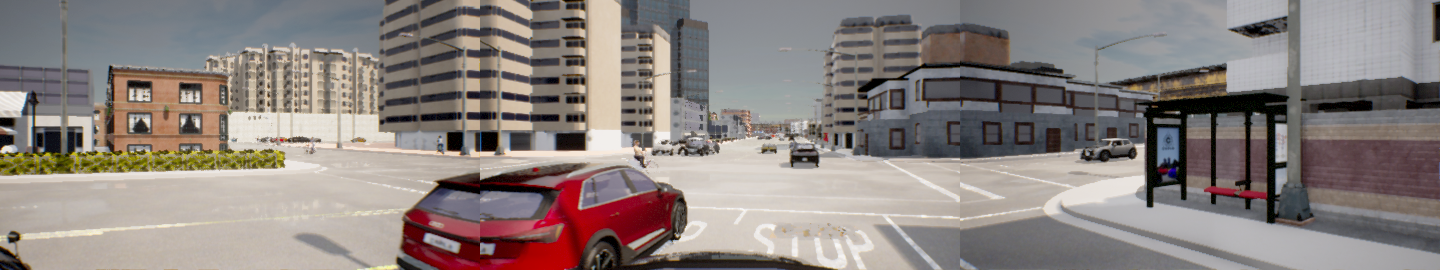}}&
    \multicolumn{1}{c}{\includegraphics[height=2cm]{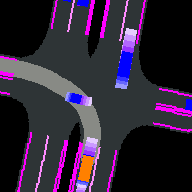}}
 
    \\
      \textbf{No Alignment Module} & & \textbf{Privileged} & \textbf{Output Distillation} & \textbf{CaT} 
    \\  
  \includegraphics[width=5.3cm,height=2cm]{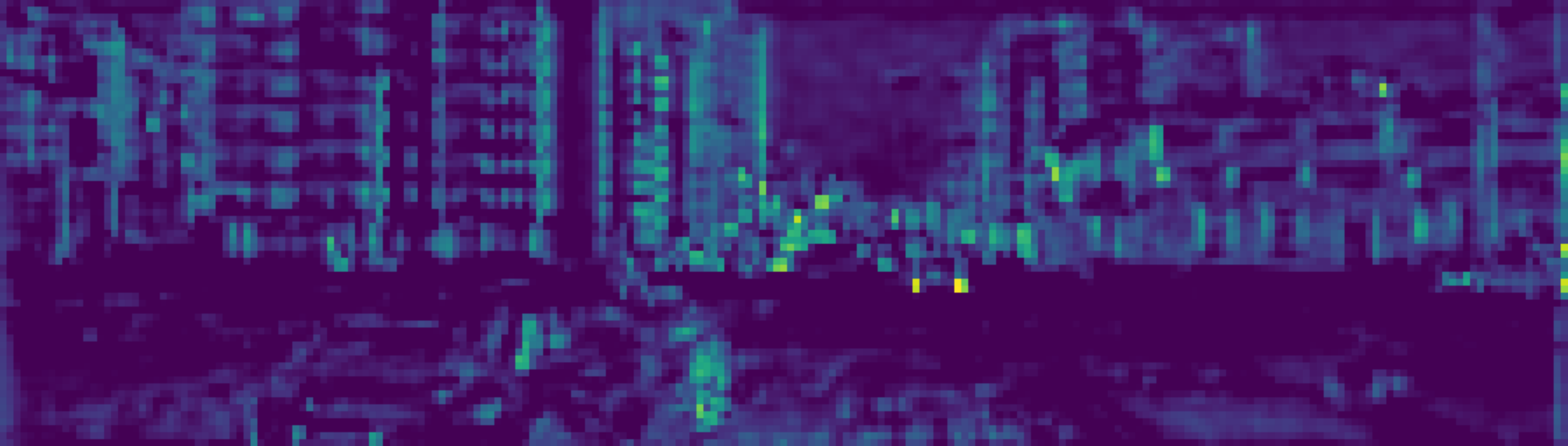} && \includegraphics[width=2cm]{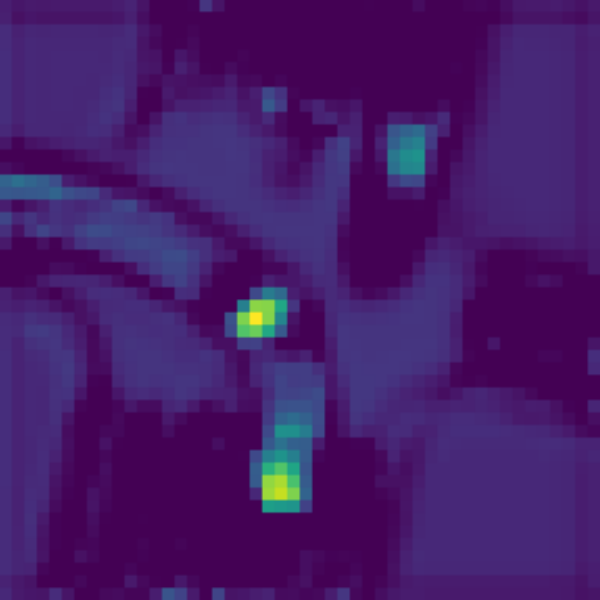} & 
  \includegraphics[width=2cm]{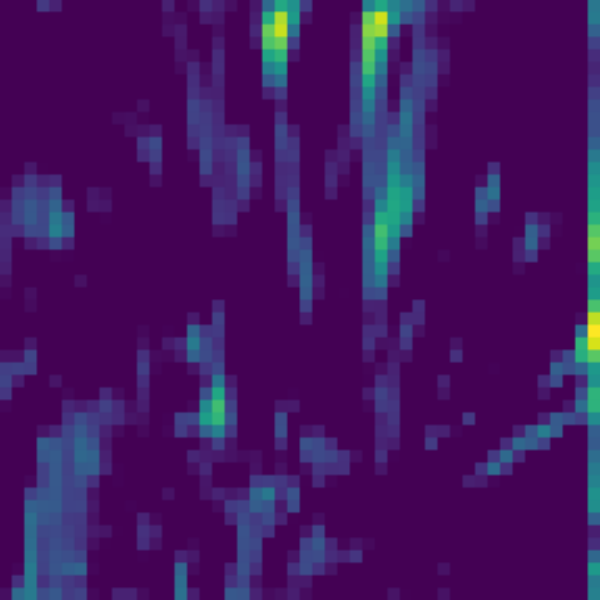} & 
  \includegraphics[width=2cm]{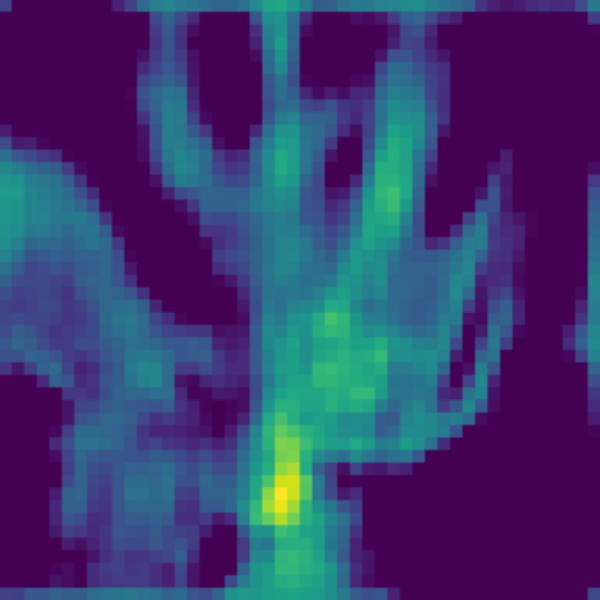} 
    \\

  \includegraphics[width=5.3cm,height=2cm]{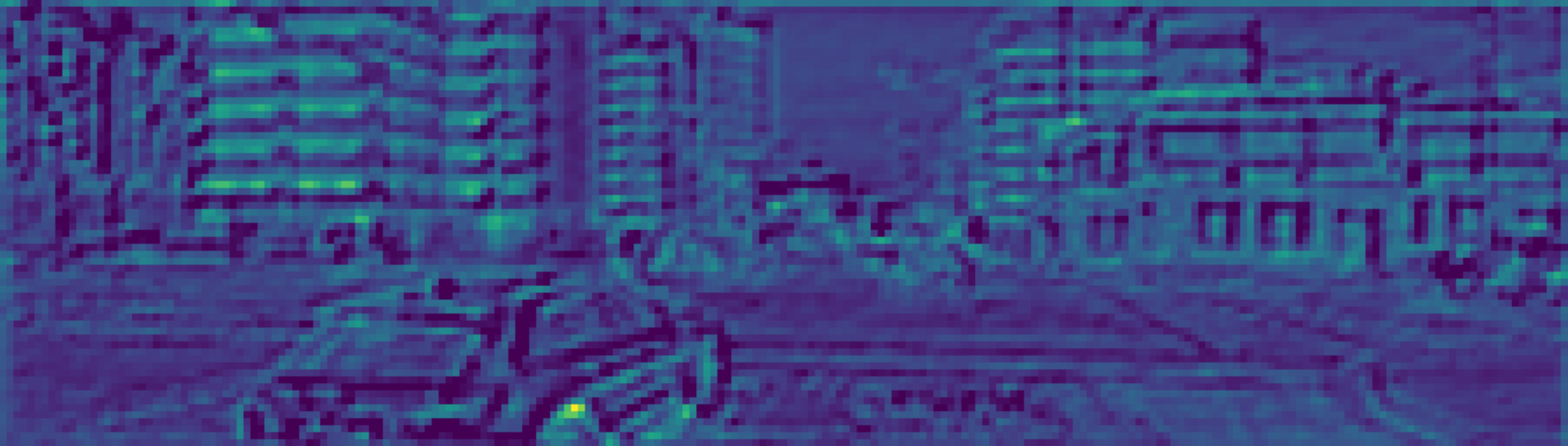} && \includegraphics[width=2cm]{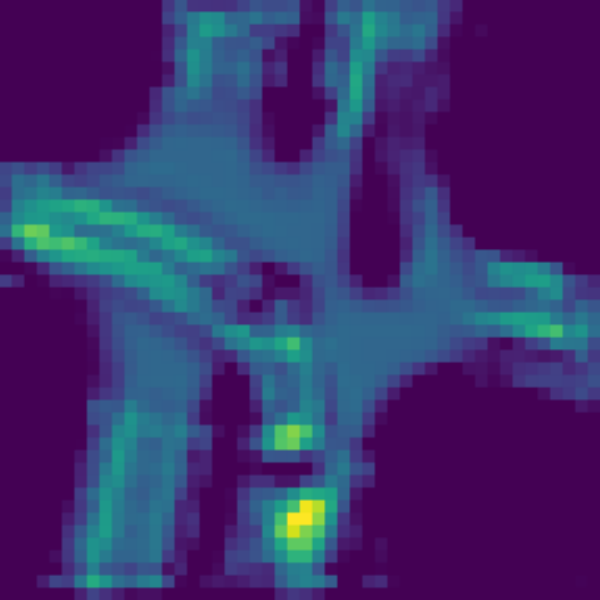} & 
  \includegraphics[width=2cm]{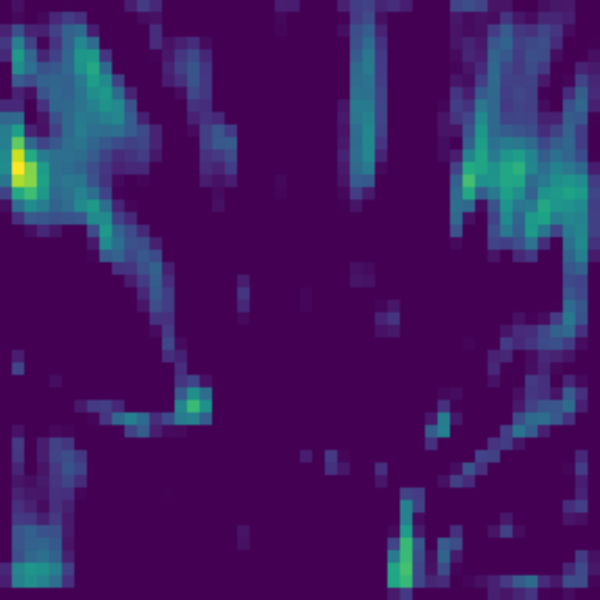} & 
  \includegraphics[width=2cm]{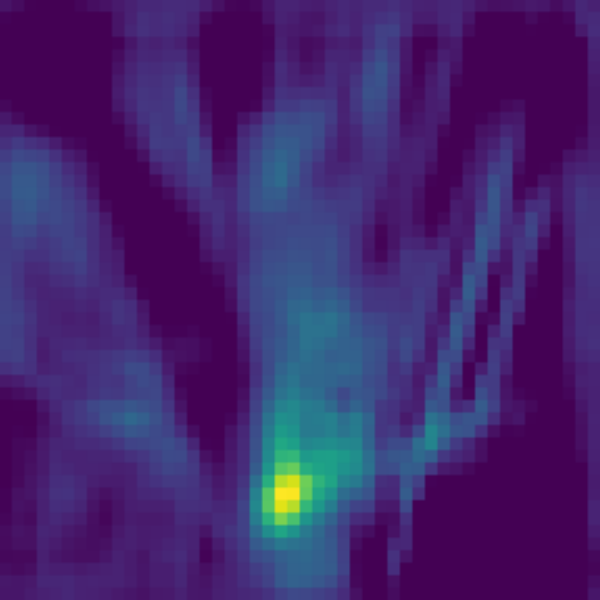} 
    \\
    \midrule
    
    \includegraphics[width=5.3cm,height=2cm]{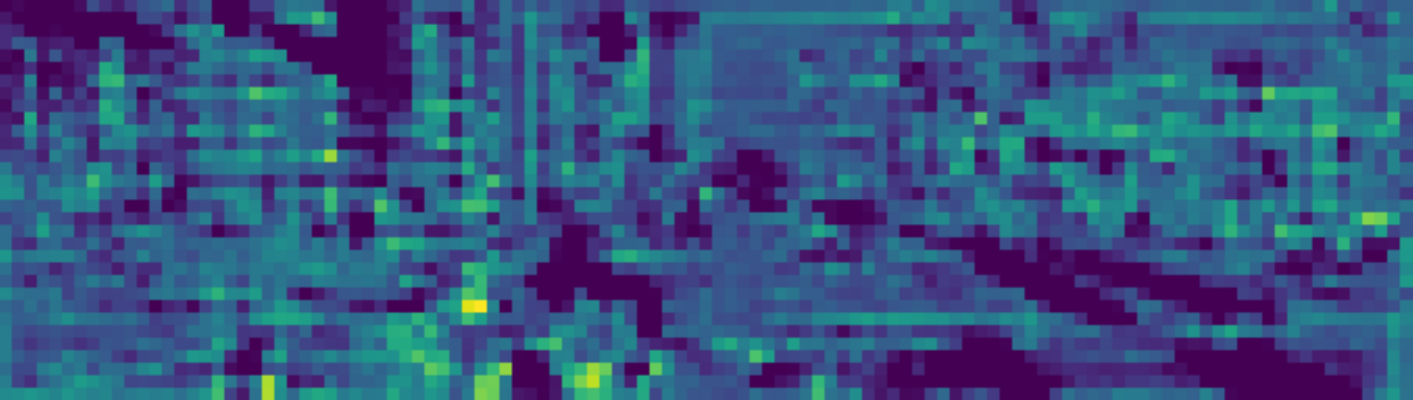} && \includegraphics[width=2cm]{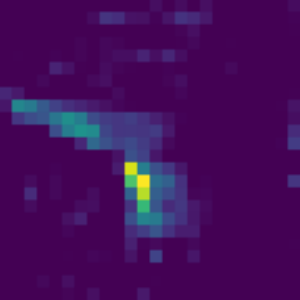} & 
  \includegraphics[width=2cm]{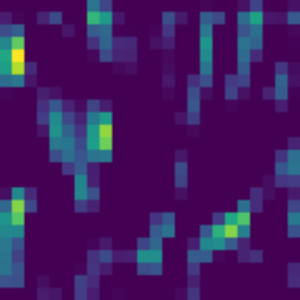} & 
  \includegraphics[width=2cm]{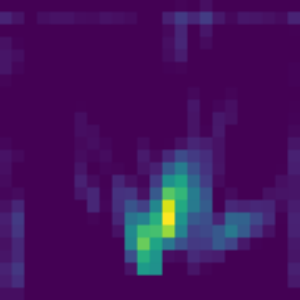} 
    \\

  \includegraphics[width=5.3cm,height=2cm]{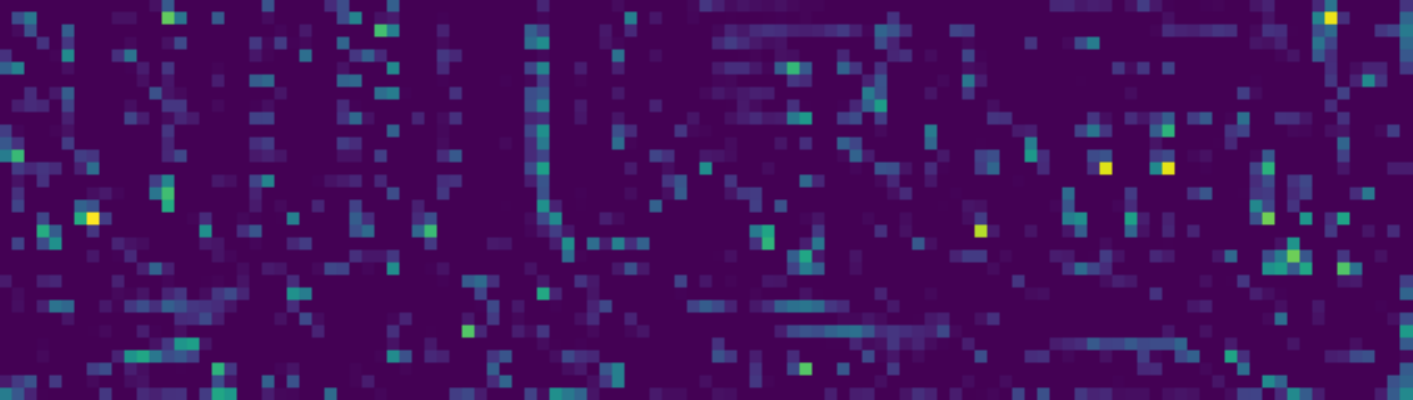} && \includegraphics[width=2cm]{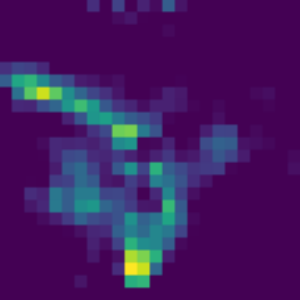} & 
  \includegraphics[width=2cm]{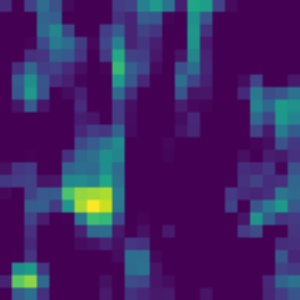} & 
  \includegraphics[width=2cm]{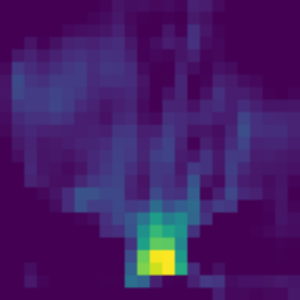} 
    \\

\end{tabular}
   \caption{\textbf{Visualization of Model Input and Features.} Analysis of a scenario where the agent is attempting to perform a left lane change.}
    \label{fig:feature_n}
\end{figure*}

\begin{figure*}
    \centering
    \setlength{\tabcolsep}{2pt}
    \begin{tabular}{ccccc}
    \multicolumn{4}{c}{\textbf{Image}}&
    \multicolumn{1}{c}{\textbf{BEV}}
    
    \\
    \multicolumn{4}{c}{\includegraphics[height=2cm]{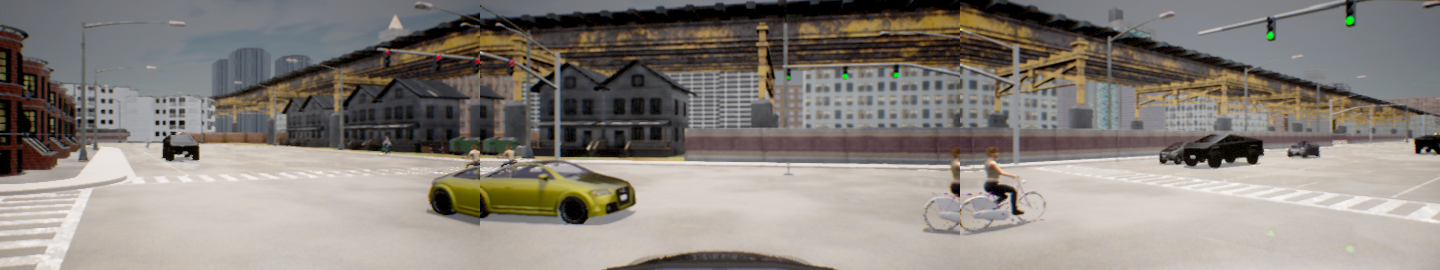}}&
    \multicolumn{1}{c}{\includegraphics[height=2cm]{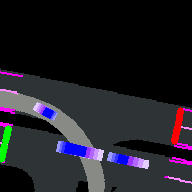}}
 
    \\
      \textbf{No Alignment Module} & & \textbf{Privileged} & \textbf{Output Distillation} & \textbf{CaT} 
    \\  
  \includegraphics[width=5.3cm,height=2cm]{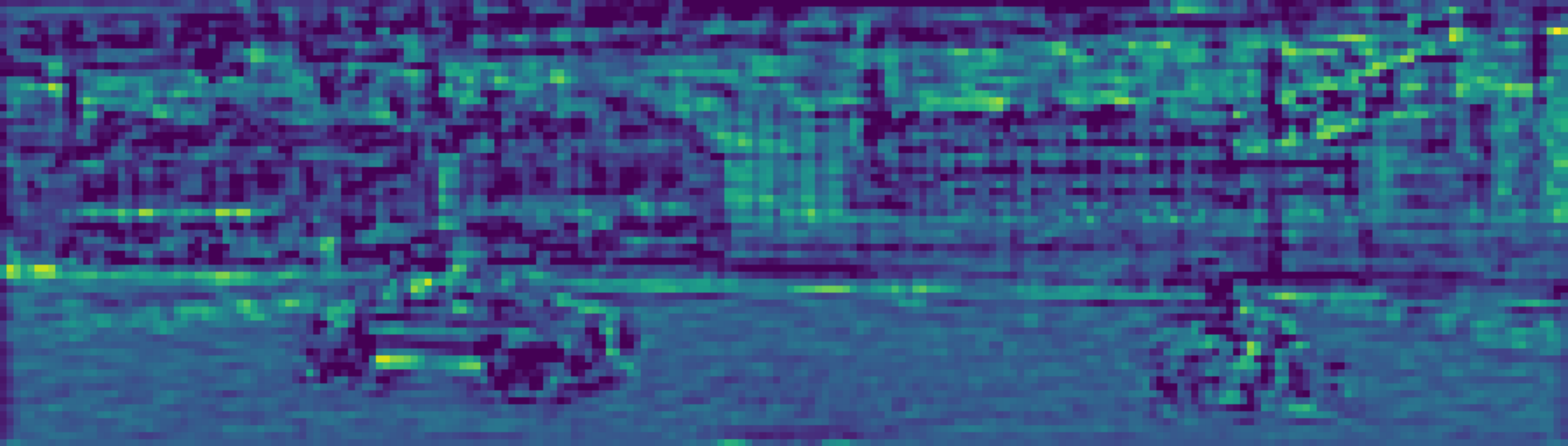} && \includegraphics[width=2cm]{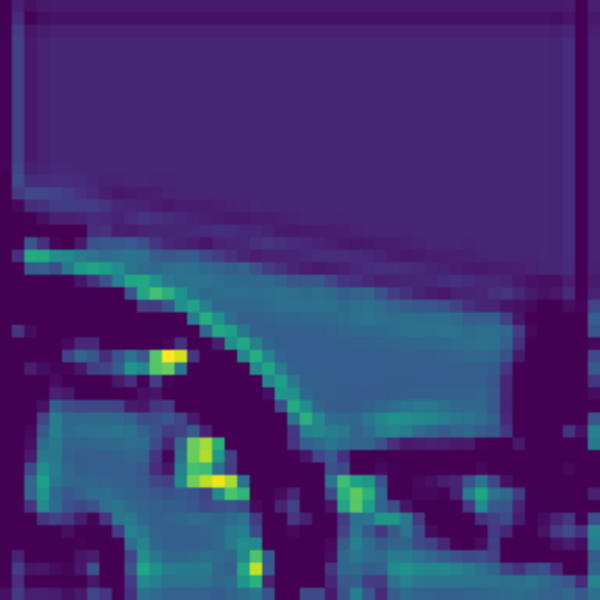} & 
  \includegraphics[width=2cm]{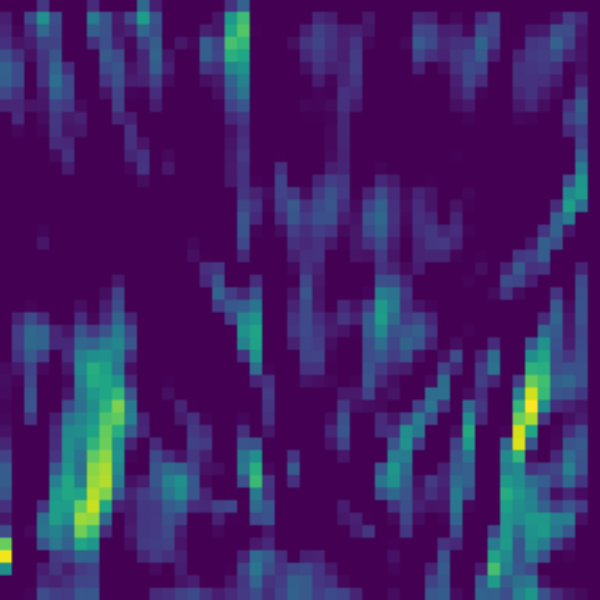} & 
  \includegraphics[width=2cm]{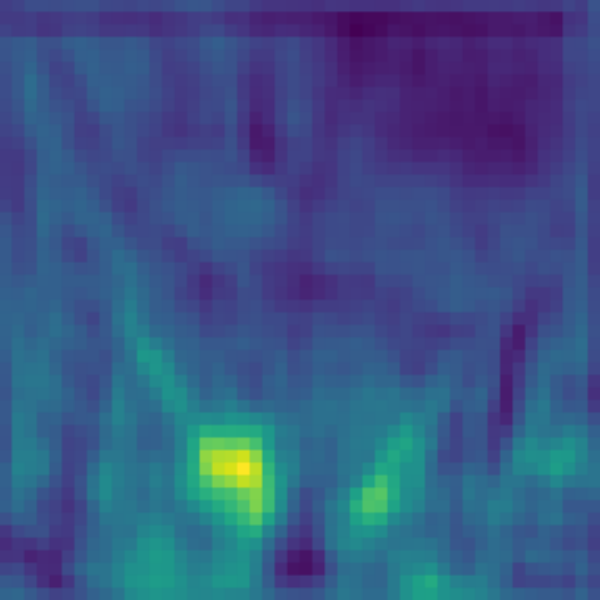} 
    \\

  \includegraphics[width=5.3cm,height=2cm]{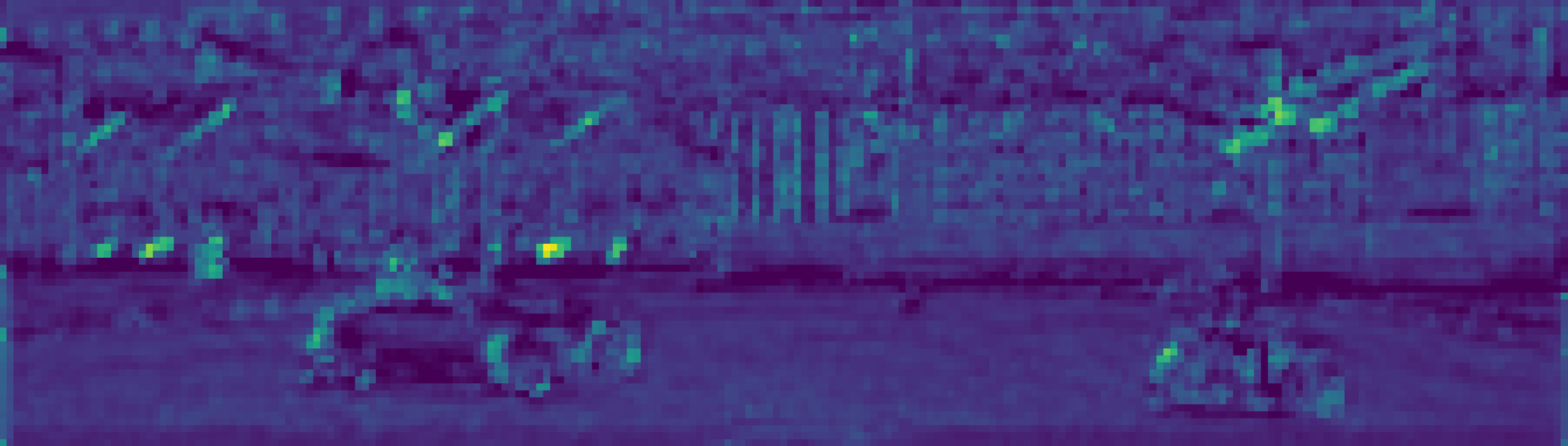} && \includegraphics[width=2cm]{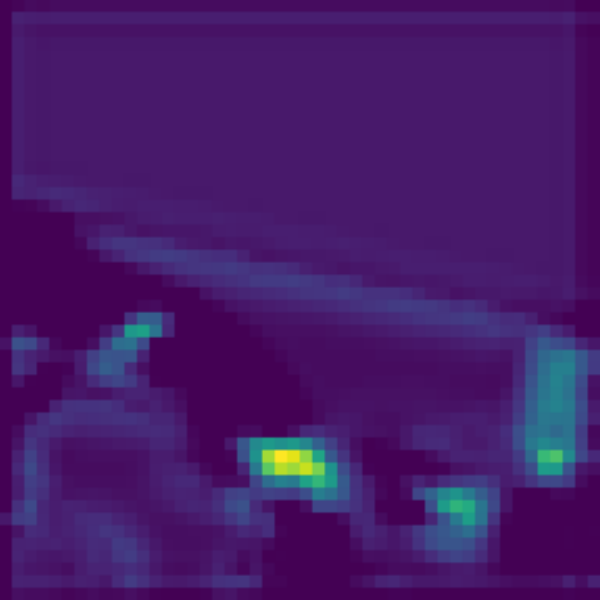} & 
  \includegraphics[width=2cm]{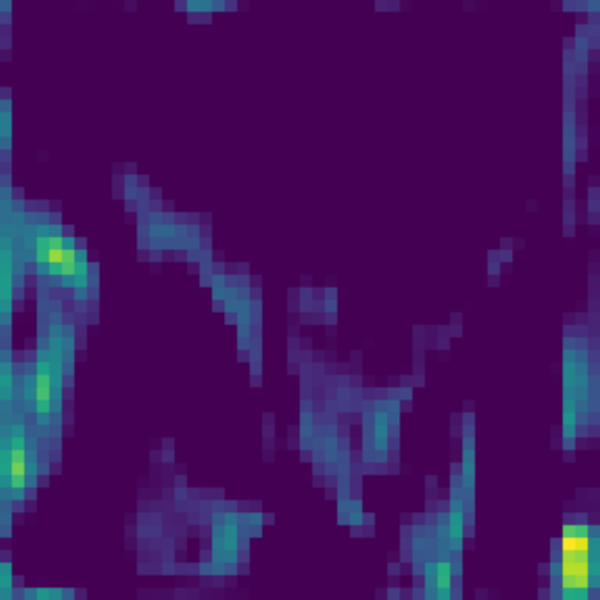} & 
  \includegraphics[width=2cm]{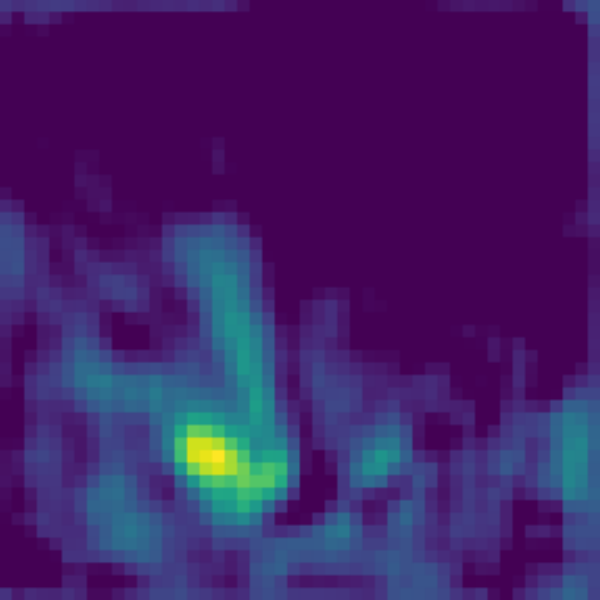} 
    \\
    \midrule
    
     \includegraphics[width=5.3cm,height=2cm]{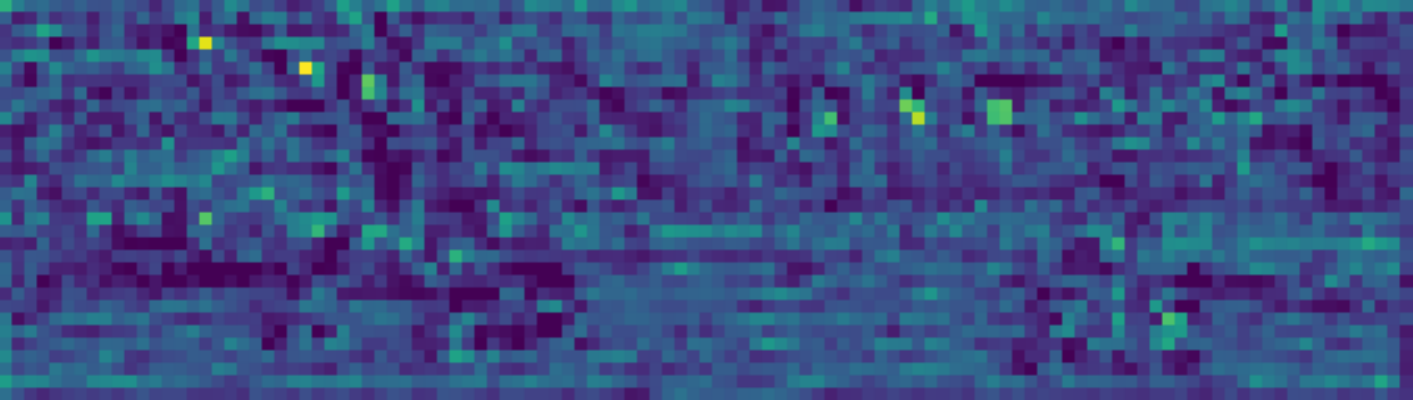} && \includegraphics[width=2cm]{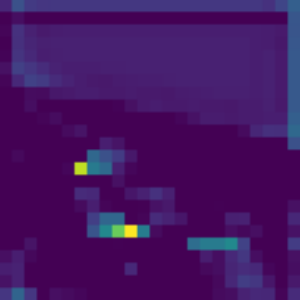} & 
  \includegraphics[width=2cm]{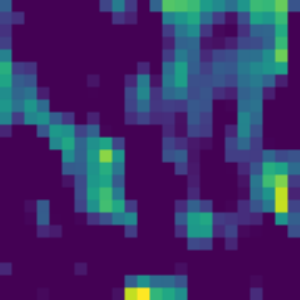} & 
  \includegraphics[width=2cm]{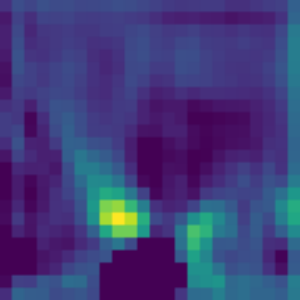} 
    \\

  \includegraphics[width=5.3cm,height=2cm]{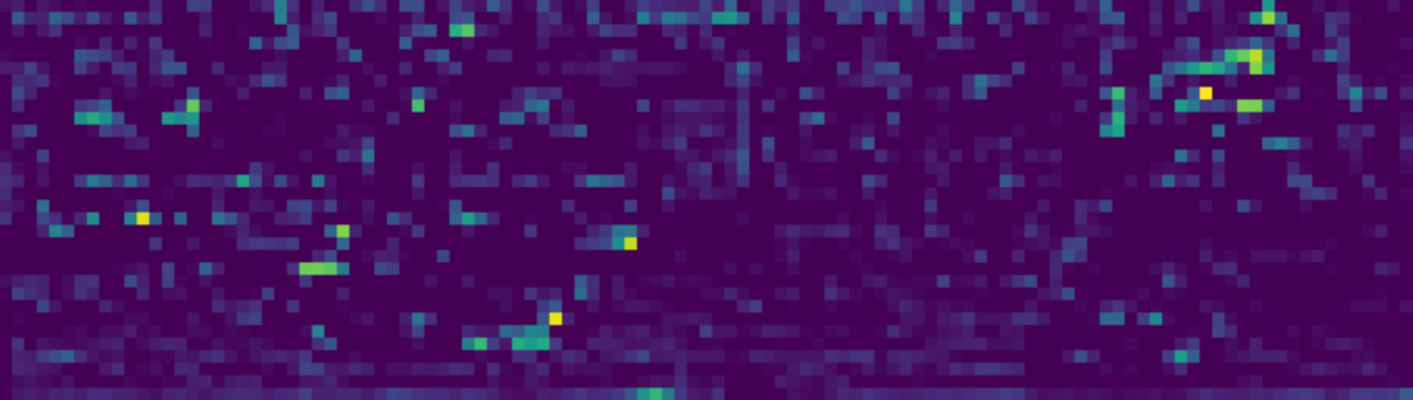} && \includegraphics[width=2cm]{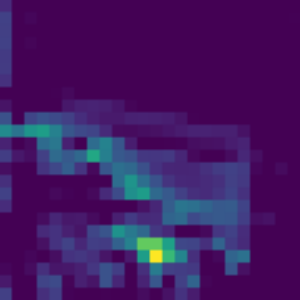} & 
  \includegraphics[width=2cm]{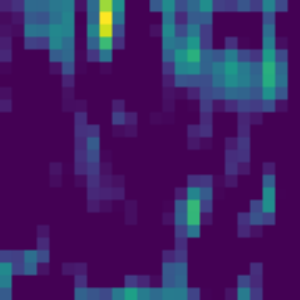} & 
  \includegraphics[width=2cm]{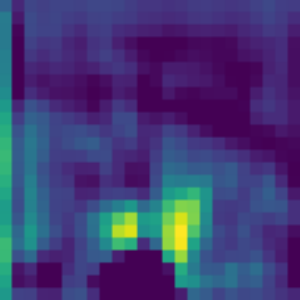} 
    \\

\end{tabular}
    \caption{\textbf{Visualization of Model Input and Features.} Analysis of a scenario where the agent is negotiating with traffic while turning left at an intersection.}
    \label{fig:feature_o}
\end{figure*}

\begin{figure*}
    \centering
    \setlength{\tabcolsep}{2pt}
    \begin{tabular}{ccccc}
    \multicolumn{4}{c}{\textbf{Image}}&
    \multicolumn{1}{c}{\textbf{BEV}}

\\
    \multicolumn{4}{c}{\includegraphics[height=2cm]{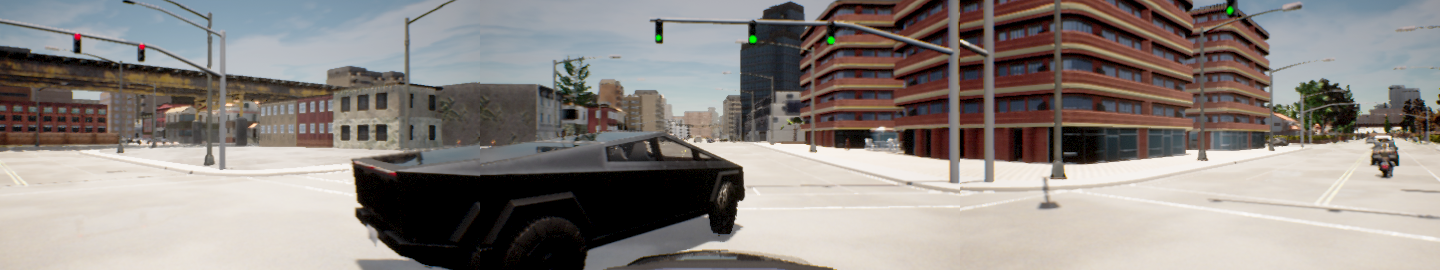}}&
    \multicolumn{1}{c}{\includegraphics[height=2cm]{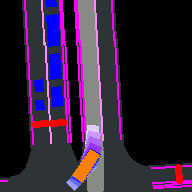}}
 
    \\
      \textbf{No Alignment Module} & & \textbf{Privileged} & \textbf{Output Distillation} & \textbf{CaT} 
    \\  
  \includegraphics[width=5.3cm,height=2cm]{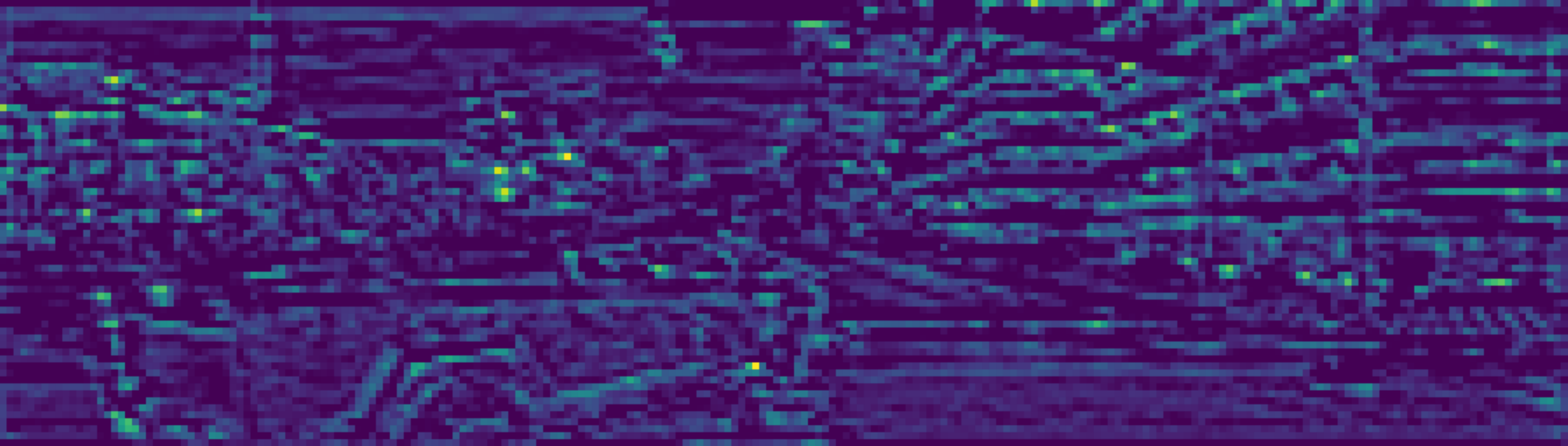} && \includegraphics[width=2cm]{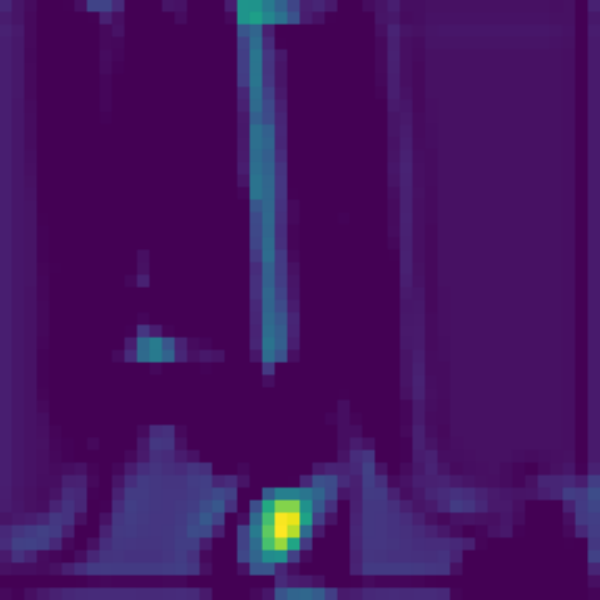} & 
  \includegraphics[width=2cm]{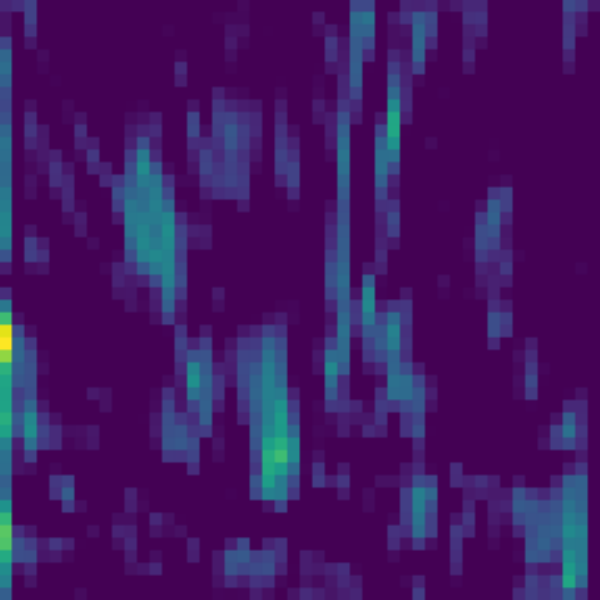} & 
  \includegraphics[width=2cm]{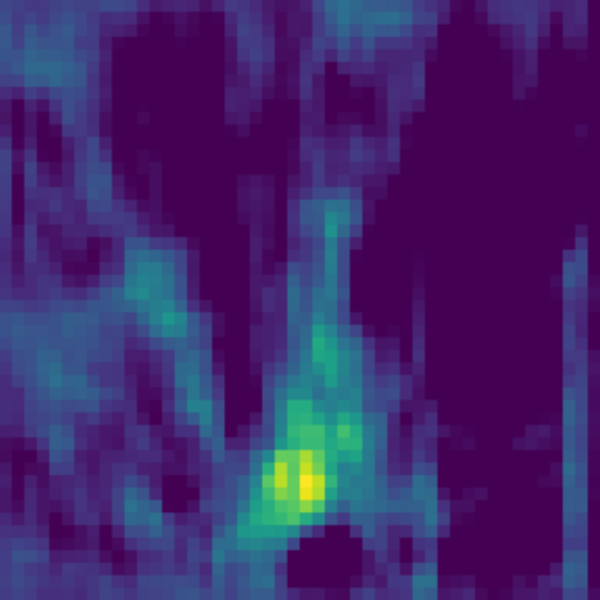} 
    \\

  \includegraphics[width=5.3cm,height=2cm]{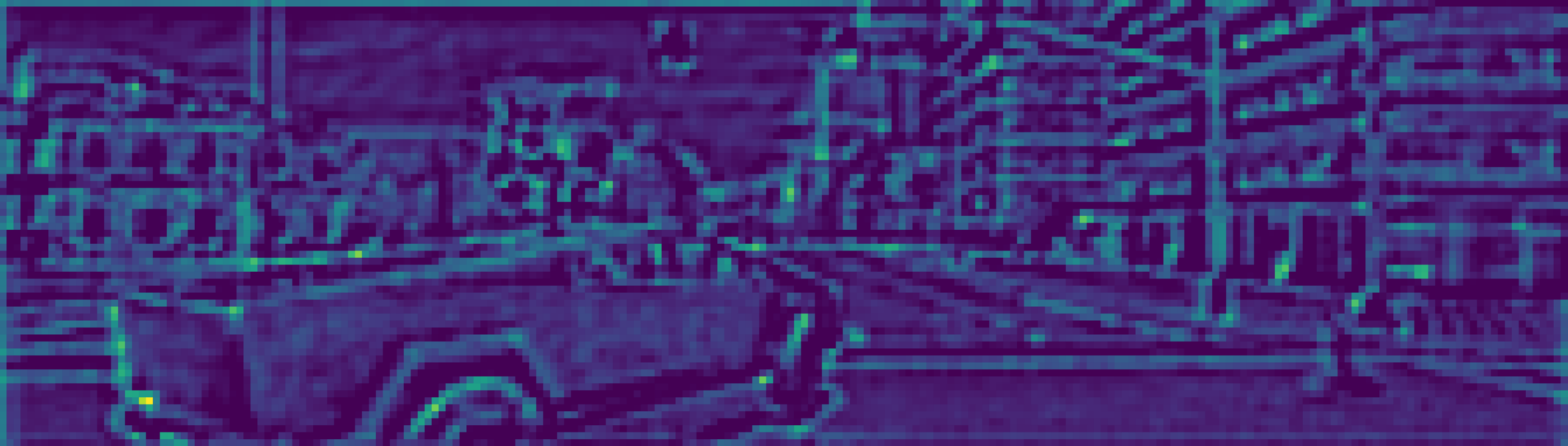} && \includegraphics[width=2cm]{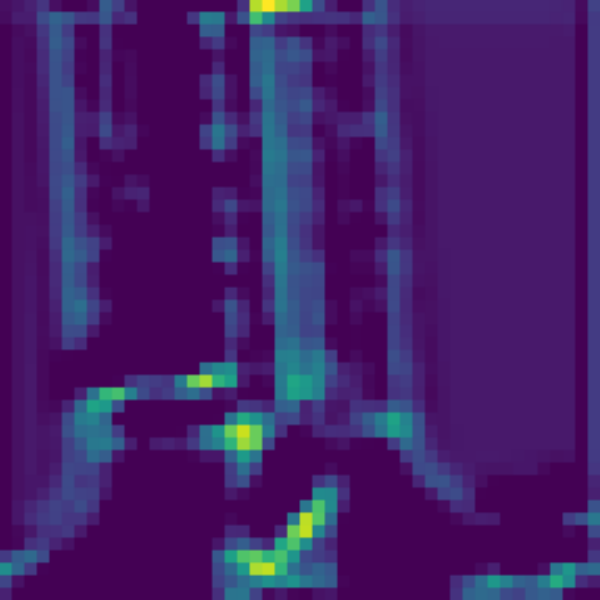} & 
  \includegraphics[width=2cm]{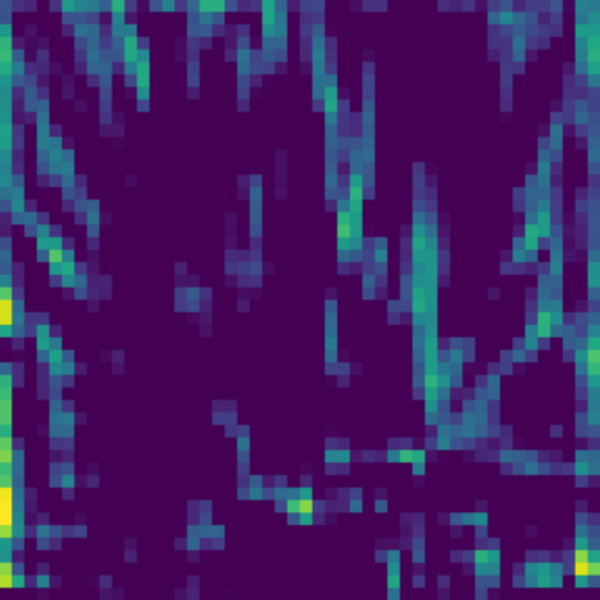} & 
  \includegraphics[width=2cm]{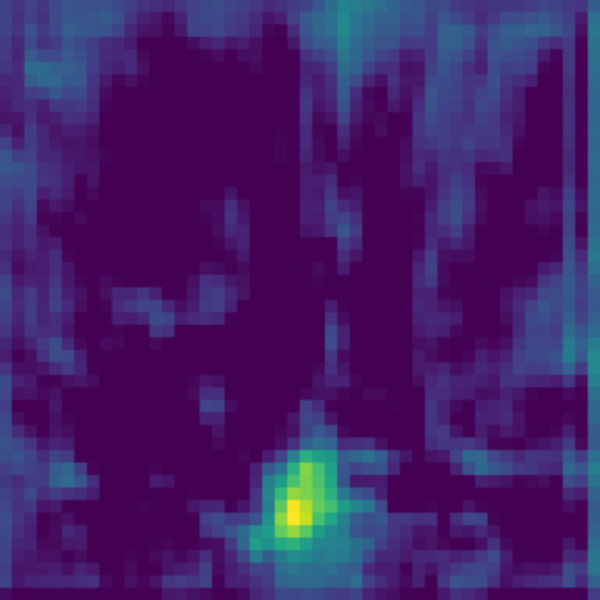} 

    \\
    
    \midrule
    
    \includegraphics[width=5.3cm,height=2cm]{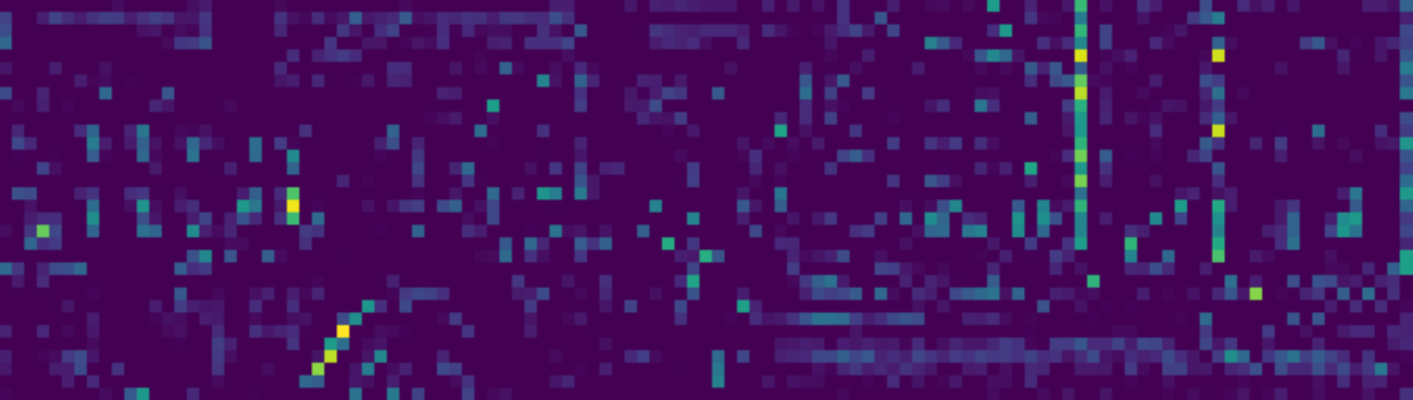} && \includegraphics[width=2cm]{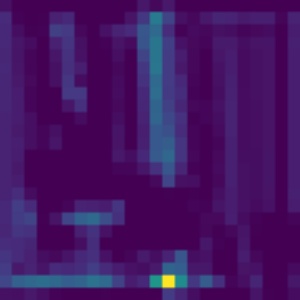} & 
  \includegraphics[width=2cm]{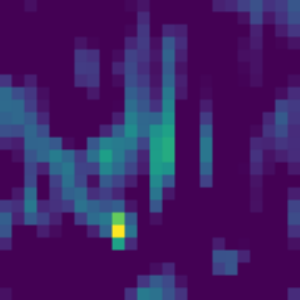} & 
  \includegraphics[width=2cm]{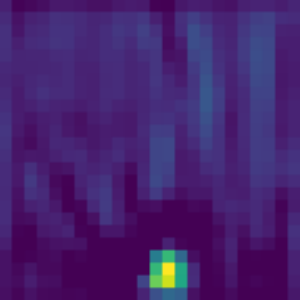} 
    \\

  \includegraphics[width=5.3cm,height=2cm]{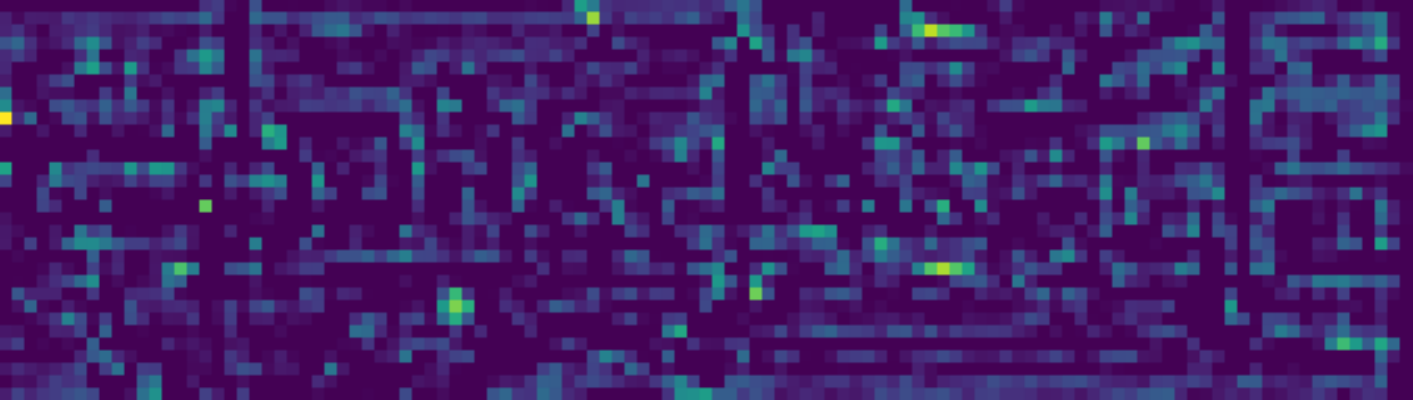} && \includegraphics[width=2cm]{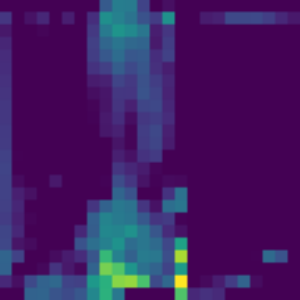} & 
  \includegraphics[width=2cm]{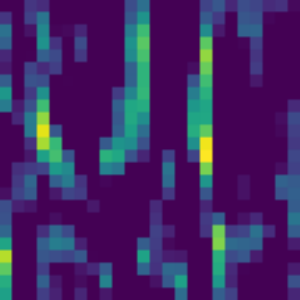} & 
  \includegraphics[width=2cm]{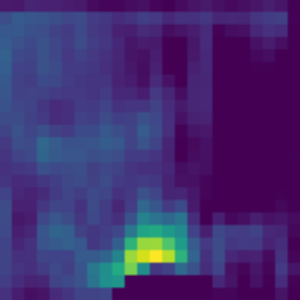} 

    \\

\end{tabular}
    \caption{\textbf{Visualization of Model Input and Features.} Analysis of a scenario the agent is driving forward at an intersection, yielding to a vehicle coming from the left.}
    \label{fig:feature_p}
\end{figure*}

\boldparagraph{Feature Visualization and Additional Qualitative Results} To comprehensively analyze the benefits of the proposed CaT approach, we provide ample qualitative analysis under various scenarios in CARLA in Figure \ref{fig:feature_a}-\ref{fig:feature_p}. For each example, we show the three RGB views, the ground truth BEV observations, as well as internal features for diagnosing four different models. We visualize the higher resolution first and second distillation feature maps (\ie, $\cF_1$ and $\cF_2$) for the privileged, output distillation, and proposed CaT sensorimotor agent. In general, we observe how the internal features of CaT demonstrate a more effective representation of scene geometry and objects across diverse scenarios. Moreover, we compare the results with a traditional baseline RGB model (based on Wu~\etal~\cite{wu2022trajectory}). While the baseline RGB model is trained to predict BEV waypoints, we demonstrate the final convolutional feature maps to generally exhibit image-level reasoning, in contrast to the proposed CaT agent. Moreover, RGB-based agents may also use a homography transform over the predicted waypoints at the very end of the architecture~\cite{lbc}. However, in the case of CaT, we find that instilling BEV-based reasoning early on in the network significantly improves driving performance. Overall, this benefits driving in cases where traditional sensorimotor agents tend to fail, \ie, around dense and dynamic scenarios.

\end{document}